\documentclass[]{article}

 \PassOptionsToPackage{round}{natbib}
 \usepackage[final]{mysty}

\usepackage[utf8]{inputenc} 
\usepackage[T1]{fontenc}    
\usepackage[usenames,dvipsnames,svgnames,table]{xcolor}
\usepackage[colorlinks=true,linkcolor=red,citecolor=gray]{hyperref}       
\usepackage{url}            
\usepackage{booktabs}       
\usepackage{amsfonts}       
\usepackage{nicefrac}       
\usepackage{microtype}      

\title{Kernel functions based on triplet comparisons}

\author{
  Matthäus Kleindessner\thanks{Work done while being a PhD student at the University of Tübingen.} \\
  Department of Computer Science\\
  Rutgers University\\
  Piscataway, NJ 08854 \\
\texttt{mk1572@cs.rutgers.edu} \\
  \And
  Ulrike von Luxburg \\
  Department of Computer Science\\
   University of Tübingen \\
  Max Planck Institute for Intelligent Systems, Tübingen\\
  \texttt{luxburg@informatik.uni-tuebingen.de}
}

\usepackage{amsmath}
\usepackage{amssymb}
\usepackage{bbm}
\usepackage{bm}
\usepackage{graphicx}
\usepackage{overpic}
\usepackage{color}
\usepackage{array}
\usepackage[rightcaption]{sidecap}    
\sidecaptionvpos{figure}{c}

\newcommand{\dataset}{\mathcal{D}}
\newcommand{\ktau}{Kendall's~$\boldsymbol\tau$}
\newcommand{\R}{\mathbb{R}}
\newcommand{\N}{\mathbb{N}}
\newcommand{\indi}{\mathbbm{1}}

\begin{document}

\maketitle

\begin{abstract}
Given only information in the form  of similarity triplets ``Object A is more similar to object B than to object~C'' about a data set,
we propose two ways of defining a kernel function on the data set. 
While previous approaches construct a low-dimensional Euclidean embedding of the data set that reflects the given similarity triplets, we aim at defining 
kernel functions 
that correspond to high-dimensional embeddings. 
These kernel functions can subsequently be used to apply 
any kernel method to 
the data set. 
\end{abstract}

\section{Introduction}\label{sec_intro}
Assessing similarity between objects is an inherent part of many machine learning problems,
be it in an unsupervised task like clustering, in which similar objects should be grouped
together, or in  classification, where many algorithms are based on the
assumption that similar inputs 
should 
produce similar outputs. 
In a typical machine learning setting one 
assumes to be given a data set~$\dataset$ of objects together with a dissimilarity function~$d$ 
(or,
 equivalently, 
a similarity function~$s$) 
quantifying how ``close'' objects are to each other. In recent years, however, a 
new
branch  of  the  machine  learning  literature  has  emerged  that  relaxes  this  scenario (see
the next paragraph and
 Section~\ref{section_related_work} for references). Instead of being able to evaluate
$d$
itself, we only get to see a collection of similarity triplets of the form ``Object~$A$ is more similar to object~$B$ than to object~$C$'', 
which claims 
that $d(A,B)<d(A,C)$. 
The main motivation for 
this relaxation 
comes from human-based computation:  It is widely accepted that humans are 
better and more reliable at providing 
similarity triplets, 
which means
assessing similarity on a relative scale, 
than at providing 
similarity estimates on an absolute scale 
(``The similarity between objects $A$ and $B$ is 0.8''). 
This 
can be seen as a special case of the general observation that humans are 
better at
comparing 
two 
stimuli than at identifying a single one \citep{stewart2005}. 
 For this reason, whenever one is lacking a meaningful dissimilarity function that can be evaluated automatically and has to incorporate human expertise into the machine learning process, collecting similarity triplets (e.g., via crowdsourcing) 
may
 be an appropriate means.  

Given a data set $\dataset$ and similarity triplets for its objects,
it is not immediately clear how to solve machine learning problems on
$\dataset$. A general approach is to construct an ordinal embedding of
$\dataset$, that is to map objects to a Euclidean space of a small
dimension such that the given 
triplets are preserved
as well as possible \citep{AgarwalEtal07,TamuzEtal2011,stoch_trip_embed,
terada14,
Ukkonen_multiview,Heim2015,amid2017,jamieson_finite}. 
Once such an ordinal embedding has been constructed, one can 
solve a problem on $\dataset$ by solving it on the 
embedding. Only recently, algorithms have been proposed~for~solving~various
specific
problems directly 
without constructing an ordinal embedding as an intermediate step 
\citep{crowdmedian,kleindessner16}.   
With this paper we provide another generic means for solving machine learning problems based on similarity triplets that is different from the ordinal embedding approach. We define two data-dependent kernel functions
on~$\dataset$, 
corresponding to high-dimensional embeddings
of $\dataset$, 
that can subsequently be used 
by 
any kernel method. 
Our proposed kernel functions measure similarity between two objects in $\dataset$ by comparing to which extent the two objects 
give rise to resembling similarity triplets. The 
intuition 
is that this 
quantifies
the relative difference in the locations of the two objects in $\dataset$. 
Experiments on both artificial and real data show that this is indeed the case and that the similarity scores defined by our kernel functions are meaningful. 
Our approach is appealingly simple, and other than ordinal embedding algorithms our kernel functions are  deterministic and 
parameter-free. We observe them to run 
significantly
 faster than well-known embedding algorithms and to be ideally suited for a landmark~design.

\paragraph{Setup}
Let $\mathcal{X}$ be an arbitrary set and $d:\mathcal{X}\times\mathcal{X}\rightarrow \R^+_0$ be a symmetric dissimilarity function on $\mathcal{X}$: a higher value of $d$ means that two elements of $\mathcal{X}$
are more dissimilar 
to each other. 
The terms dissimilarity and distance are used synonymously.
To simplify presentation, we assume that for all triples of distinct objects $A,B,C\in\mathcal{X}$ either $d(A,B)<d(A,C)$ or $d(A,B)>d(A,C)$ is true. 
Note that we do not 
require 
$d$ to be a metric.
We formally define a similarity triplet as binary answer to a dissimilarity comparison 
\begin{align}\label{parwise_comp}
d(A,B)\stackrel{?}{<}d(A,C). 
\end{align}
We refer to 
$A$ as the anchor object.
A similarity triplet 
can
 be incorrect, 
meaning that it 
claims a positive answer to the comparison \eqref{parwise_comp} although in fact 
the negative answer is true.
In the following, we deal with a finite data set $\dataset=\{x_1,\ldots,x_n\}\subseteq \mathcal{X}$ and collections of similarity triplets 
that are encoded as follows: an ordered triple of distinct objects
$(x_i,x_j,x_k)$ 
means 
$d(x_i,x_j)<d(x_i,x_k)$. A collection of similarity triplets is the only information that we are given about $\dataset$. 
Note that such a collection  does not necessarily provide an answer to every possible dissimilarity comparison \eqref{parwise_comp}.

\section{Our kernel functions}\label{sec_kernel_descr}

Assume we are given a collection $\mathcal{S}$ of similarity triplets for the objects of $\dataset$. 
Similarity triplets in $\mathcal{S}$ 
can
be incorrect, 
but
for the moment 
assume that 
 contradicting triples
 $(x_i,x_j,x_k)$ and $(x_i,x_k,x_j)$ cannot be present in $\mathcal{S}$ at the same time. We will discuss how to deal with the general case below.

\paragraph{Kernel function $\bm{k_1}$} 
Our first kernel function is based on the following idea: We fix two objects $x_a$ and $x_b$. In order to compute a similarity score between $x_a$ and $x_b$ we would like to rank all 
objects in $\dataset$ with respect to their distance from $x_a$ and also rank them with respect to their distance from $x_b$, and take a similarity score between these two rankings as similarity score between $x_a$ and $x_b$. One possibility to measure similarity between rankings is given by the 
famous
Kendall tau correlation
coefficient \citep{kendalltau}, which is also known as \ktau: for two rankings of $n$ items, \ktau~between the two rankings is 
the fraction of concordant pairs of items minus the fraction of
discordant pairs of items. Here, 
a pair of two items $i_1$ and $i_2$ is concordant if  $i_1 \prec i_2$ or $i_1 \succ i_2$ 
according to
 both rankings, and discordant 
if it satisfies $i_1 \prec i_2$ according to one and $i_1 \succ i_2$ 
according to the other ranking. Formally, a ranking is represented by a permutation $\sigma: \{1,\ldots,n\}\rightarrow \{1,\ldots,n\}$ such that $\sigma(i)\neq\sigma(j)$, $i\neq j$, and $\sigma(i)=m$ means that item~$i$ is ranked at the \mbox{$m$-th} position. Given two rankings $\sigma_1$ and $\sigma_2$, the number of concordant pairs equals
\begin{align*}
f_c(\sigma_1,\sigma_2)=\sum_{i<j}&[\indi\{\sigma_1(i)<\sigma_1(j)\}\indi\{\sigma_2(i)<\sigma_2(j)\}+\indi\{\sigma_1(i)>\sigma_1(j)\}\indi\{\sigma_2(i)>\sigma_2(j)\}],
\end{align*}
the number of discordant pairs equals
\begin{align*}
f_d(\sigma_1,\sigma_2)=\sum_{i<j}&[\indi\{\sigma_1(i)<\sigma_1(j)\}\indi\{\sigma_2(i)>\sigma_2(j)\}+\indi\{\sigma_1(i)>\sigma_1(j)\}\indi\{\sigma_2(i)<\sigma_2(j)\}],
\end{align*}
and \ktau~between $\sigma_1$ and $\sigma_2$ is given by
$\boldsymbol\tau(\sigma_1,\sigma_2)=\left[f_c(\sigma_1,\sigma_2)-f_d(\sigma_1,\sigma_2)\right]/\tbinom{n}{2}$.
\begin{figure}[t]
\vspace{2.6mm}
\centering
\includegraphics[scale=0.16]{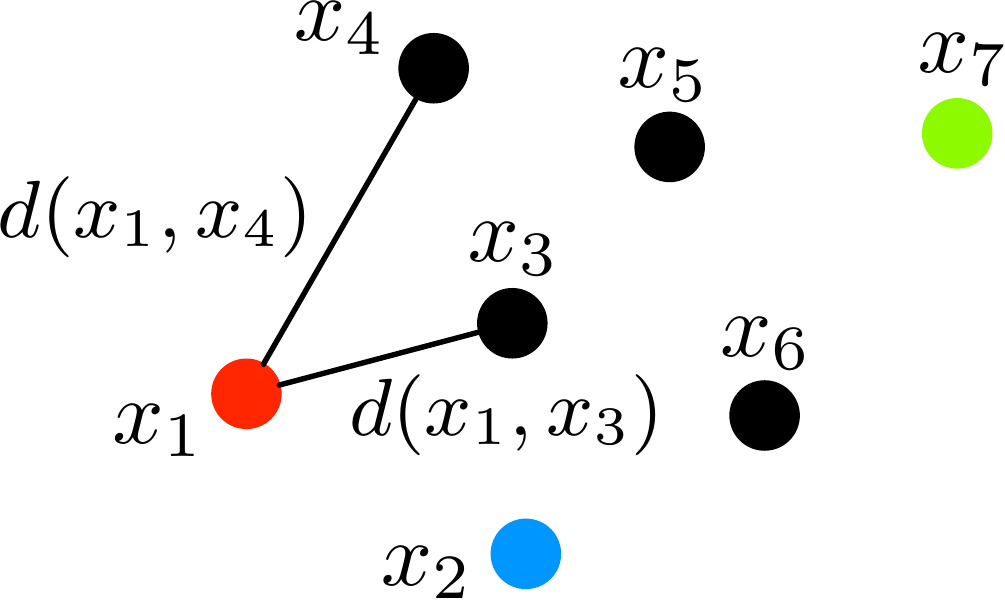}
\hspace{4cm}
\includegraphics[scale=0.16]{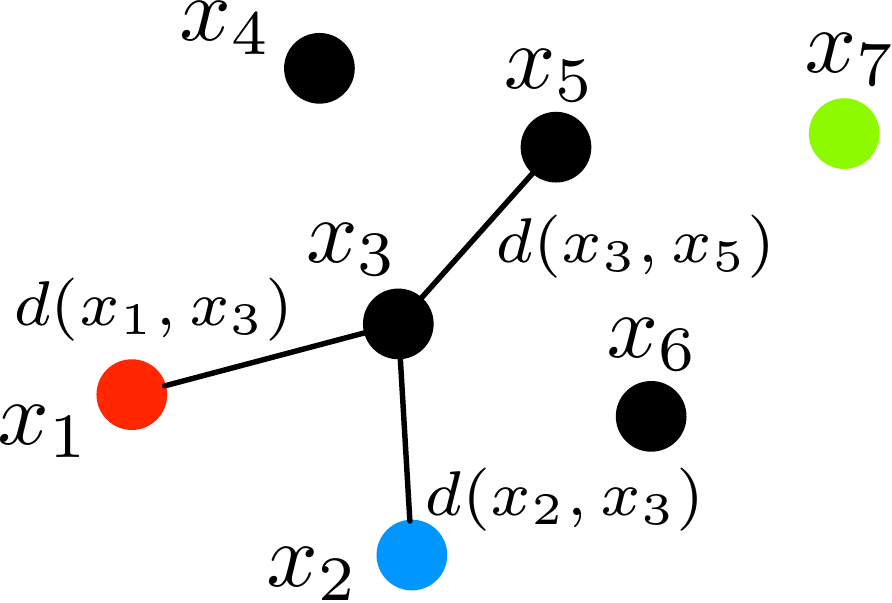}
\caption{Illustrations of the ideas behind 
$k_1$ (left) and $k_2$ (right). \textbf{For $\bm{k_1}$:} 
In order to compute a similarity score between $x_1$ (in red) and $x_2$ (in blue) we would like to rank all objects with respect to their distance from $x_1$ and also with respect to their distance from $x_2$ and compute \ktau~between the two rankings. In this example, the objects would rank as $x_1 \prec x_3 \prec x_2 \prec x_4 \prec x_5 \prec x_6 \prec x_7$ and $x_2 \prec x_3 \prec x_6 \prec x_1 \prec x_5 \prec x_4 \prec x_7$, respectively. \ktau~between  these two rankings
is~$1/3$,
and 
this would be the 
similarity 
score between $x_1$ and $x_2$. For comparison, 
the 
score between $x_1$ and $x_7$ (in green) would be 
$-5/7$,
 and between $x_2$ and $x_7$ it would be
 $-3/7$.
\textbf{For~$\bm{k_2}$:}  
In order to compute a similarity score between $x_1$ 
and $x_2$ 
we would like to check for every pair of objects $(x_i,x_j)$ whether the distance comparisons $d(x_i,x_1)\stackrel{\mbox{\tiny ?}}{\scriptstyle <}d(x_i,x_j)$ and $d(x_i,x_2)\stackrel{\mbox{\tiny ?}}{\scriptstyle <}d(x_i,x_j)$ yield the same result or not. 
Here, we have 32 pairs for which they yield the same result 
and 17 pairs for which they do not. 
We would assign $7^{-2}\cdot(32-17)=15/49$ as similarity score between $x_1$ and $x_2$. 
The 
score between $x_1$ and $x_7$ 
would be 3/49, and between $x_2$ and $x_7$ it would be 1/49.
}\label{kernel_ideas}
\end{figure}

By measuring similarity between the two rankings of objects (one with respect to their distance from $x_a$ and one with respect to their distance from $x_b$) with \ktau~we would  compute a similarity score between $x_a$ and $x_b$. This idea is illustrated 
with an example in Figure \ref{kernel_ideas} (left). 
It has been established  recently that \ktau~is actually a kernel function on the set of total rankings \citep{jiao_kendall}. 
Hence, 
by measuring similarity on $\dataset$ in 
the described 
 way we would even end up with a kernel function on~$\dataset$ since the following holds: for any mapping $h:\dataset\rightarrow\mathcal{Z}$ and kernel function $k:\mathcal{Z}\times \mathcal{Z}\rightarrow \R$, $k\circ (h,h):\dataset \times \dataset \rightarrow \R$ is a kernel function. 

In our situation, the problem is that in most cases $\mathcal{S}$ will contain only a small fraction of all possible similarity triplets and also that some of the triplets in $\mathcal{S}$ might be incorrect, so that there is no way of ranking all objects with respect to their distance from any fixed object based on the 
similarity 
triplets in $\mathcal{S}$.
To adapt the procedure, we consider a feature map that corresponds to the kernel function just described. By a feature map corresponding to a kernel function $k:\dataset\times \dataset \rightarrow \R$ we mean a mapping $\Phi:\dataset\rightarrow \R^m$ for some $m\in\N$ such that 
$k(x_i,x_j)=\langle \Phi(x_i),\Phi(x_j) \rangle =\Phi(x_i)^T \cdot\Phi(x_j)$. 
It is easy to see from 
the above formulas 
\citep[also compare with][]{jiao_kendall} that 
a feature map corresponding to the described kernel function 
 is given by $\Phi_{k_\tau}: \dataset\rightarrow\R^{\binom{n}{2}}$ 
with
\begin{align*}
\Phi_{k_\tau}(x_a)=\frac{1}{\sqrt{\binom{n}{2}}}\cdot\bigg(\indi\{d(x_a,x_i)<d(x_a,x_j)\}-\indi\{d(x_a,x_i)>d(x_a,x_j)\}\bigg)_{1\leq i<j\leq n}.
\end{align*} 
In our situation, where we are only given $\mathcal{S}$ and cannot evaluate $\Phi_{k_\tau}$ in most cases, we 
have to replace $\Phi_{k_\tau}$~by an approximation: up to a normalizing factor, we replace an entry in $\Phi_{k_\tau}(x_a)$ by zero if we cannot evaluate it based on the 
triplets in $\mathcal{S}$. More precisely, we consider the feature map $\Phi_{k_1}: \dataset\rightarrow\R^{\binom{n}{2}}$ given by 
$\Phi_{k_1}(x_a)= \left([\Phi_{k_1}(x_a)]_{i,j}\right)_{1\leq i<j\leq n}$ with
\begin{align}\label{definition_K1}
\left[\Phi_{k_1}(x_a)\right]_{i,j}=\frac{1}{\sqrt{|\{(x_i,x_j,x_k)\in\mathcal{S}: x_i=x_a\}|}}\cdot\bigg(\indi\{(x_a,x_i,x_j)\in\mathcal{S}\}-\indi\{(x_a,x_j,x_i)\in\mathcal{S}\}\bigg)
\end{align}
and define our first 
proposed 
kernel function 
$k_1:\dataset\times\dataset\rightarrow\R$ by 
\begin{align}\label{scalarp_K1}
k_1(x_i,x_j)=
\Phi_{k_1}(x_i)^T \cdot\Phi_{k_1}(x_j). 
\end{align}
Note that the scaling factor in the definition of $\Phi_{k_1}$, 
ensuring that the feature embedding 
lies on the unit sphere, is crucial whenever the number of similarity triplets in which an object appears as anchor object is not approximately constant over the different objects. 
For ease of exposition we have assumed that every object in $\dataset$ appears at least once as an anchor object in a similarity triplet in $\mathcal{S}$. In the unlikely case that $x_a$ does not appear at least once as an anchor object, meaning that we do not have any information for ranking the objects in $\dataset$ with respect to their distance from $x_a$ at all, we simply set $\Phi_{k_1}(x_a)$ to zero (which is consistent with \eqref{definition_K1} under the convention ``0/0=0'').

\paragraph{Kernel function $\bm{k_2}$} 
Our second kernel function is based on a similar idea.  
Now we do not consider $x_a$ and $x_b$ as anchor objects when measuring their similarity, but 
compare whether they rank similarly with respect to their distances from the various other objects. 
Concretely,  we would like to count the number of pairs of objects~$(x_i,x_j)$ for which the 
comparisons
\begin{align}\label{fragenpaar}
d(x_i,x_a)\stackrel{?}{<}d(x_i,x_j)
\quad\text{and}\quad 
d(x_i,x_b)\stackrel{?}{<}d(x_i,x_j) 
\end{align}
yield the same result and subtract the number of pairs for which these comparisons yield different results. 
See the right-hand side of 
Figure \ref{kernel_ideas}
 for an illustration of this idea.
Adapted to our situation
of being only given $\mathcal{S}$ 
it corresponds to 
considering the feature map $\Phi_{k_2}: \dataset\rightarrow\R^{n^2}$ given by
\begin{align*}
&\Phi_{k_2}(x_a)=\frac{1}{\sqrt{|\{(x_i,x_j,x_k)\in\mathcal{S}: x_j=x_a \vee x_k=x_a\}|}}\cdot\\
&~~~~~~~~~~~~~~~~~~~~~~~~~~~~~~~~~~~~~~~~~~~~~~~~~~~\bigg(\indi\{(x_i,x_a,x_j)\in\mathcal{S}\}-\indi\{(x_i,x_j,x_a)\in\mathcal{S}\}\bigg)_{1\leq i,j\leq n}
\end{align*}
and defining our second 
proposed 
kernel function
$k_2:\dataset\times\dataset\rightarrow\R$ by 
\begin{align}\label{scalarp_K2}
k_2(x_i,x_j)=
\Phi_{k_2}(x_i)^T \cdot\Phi_{k_2}(x_j). 
\end{align}
Again, the scaling factor in the definition of $\Phi_{k_2}$ 
is crucial whenever there are objects appearing in more similarity triplets than others and we apply the convention ``0/0=0''.

\paragraph{Contradicting similarity triplets}
If $\mathcal{S}$ contains contradicting triples $(x_i,x_j,x_k)$ and $(x_i,x_k,x_j)$ and there might be triples 
being present repeatedly, one can 
alter the definition of $\Phi_{k_1}$ or $\Phi_{k_2}$ as follows: if $\#\{(x_a,x_i,x_j)\in\mathcal{S}\}$ denotes the number of how often the triple $(x_a,x_i,x_j)$ appears in $\mathcal{S}$, set 
$\Phi_{k_1}(x_a)=
{\widetilde{\Phi}_{k_1}(x_a)}/{\big\|\widetilde{\Phi}_{k_1}(x_a)\big\|}$ where $\widetilde{\Phi}_{k_1}(x_a)$ equals
\begin{align*}
\bigg(\frac{\#\{(x_a,x_i,x_j)\in\mathcal{S}\}-\#\{(x_a,x_j,x_i)\in\mathcal{S}\}}{\#\{(x_a,x_i,x_j)\in\mathcal{S}\}+\#\{(x_a,x_j,x_i)\in\mathcal{S}\}}\bigg)_{1\leq i<j\leq n}.
\end{align*}
The definition of $\Phi_{k_2}$ can be revised in an analogous way. In doing so, we incorporate a simple estimate of the likelihood of a triple being correct.

\subsection{Reducing diagonal dominance}
If the number $|\mathcal{S}|$ of given similarity triplets is small, our kernel functions suffer from a problem that is shared by many other kernel functions defined on complex data: 
$\Phi_{k_1}$ and $\Phi_{k_2}$ map the objects in $\dataset$ to 
sparse vectors, that is almost all of their entries are zero. As a consequence, two different feature vectors $\Phi_{k_i}(x_a)$ and $\Phi_{k_i}(x_b)$ appear to be almost orthogonal and the similarity score $k_i(x_a,x_b)$ is much smaller than the self-similarity scores $k_i(x_a,x_a)$ or $k_i(x_b,x_b)$. This phenomenon, usually referred to as diagonal dominance of the kernel function, has been observed to pose difficulties for the kernel methods using the kernel function, and several ways have been proposed for 
dealing
 with it \citep{schoelk_diagdom,Greene_diagdom}. In 
 all
 our experiments we deal with diagonal dominance in the following simple way: Let $k$ denote a kernel function and $K$ the kernel matrix on $\dataset$, that is $K=(k(x_i,x_j))_{i,j=1}^n$, which would be the input to a kernel method. Then we replace $K$ by $K-\lambda_{\textrm{min}}I$ where  
$I\in\R^{n\times n}$ 
denotes the 
identity
matrix and $\lambda_{\textrm{min}}$ is the smallest eigenvalue of $K$.

\begin{SCfigure}[50][t]
\includegraphics[scale=0.33]{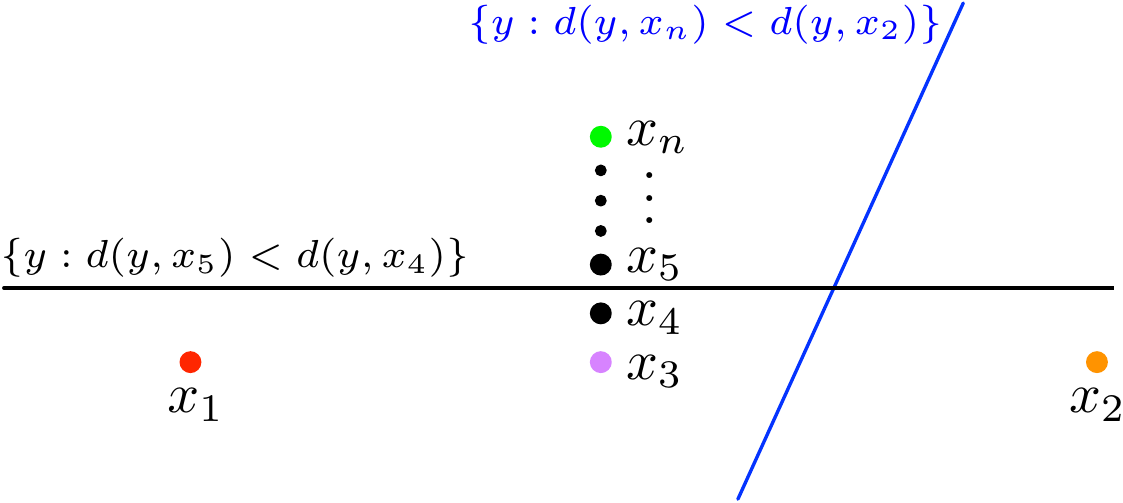}
\caption{$k_1$ measures similarity between two objects 
by counting
in how many of the halfspaces that are obtained from distance comparisons the two objects reside at the same time. The outcome does not only depend on the distance between the two objects, but also on their location within the data set: although $x_1$ and $x_2$ are located far apart, 
$k_1$ considers them to be very similar. See the running text for details.}\label{counterex}
\end{SCfigure}

\subsection{Geometric intuition}\label{sec_geom_intuition}
Intuitively, our kernel functions measure similarity between $x_a$ and $x_b$ by quantifying  
to which extent $x_a$ and $x_b$ can be expected to be located in the same region of $\dataset$:
Think of $\dataset$ as 
a subset of $\R^m$ and $d$ being the Euclidean metric. A similarity triplet $d(x_a,x_i)<d(x_a,x_j)$ then tells us that $x_a$ resides in the halfspace defined by the hyperplane that is perpendicular to the line segment connecting $x_i$ and $x_j$ and goes through 
the segment's
 midpoint. If there is also a similarity triplet $d(x_b,x_i)<d(x_b,x_j)$, $x_a$ and $x_b$ thus are located in the same halfspace (assuming the correctness of the similarity triplets) and this is reflected by a higher value of 
$k_1(x_a,x_b)$.
Similarly, a similarity triplet $d(x_i,x_a)<d(x_i,x_j)$ tells us that $x_a$ is located in a ball with radius $d(x_i,x_j)$ centered at $x_i$, and the value of 
$k_2(x_a,x_b)$
 is higher if there is a similarity triplet $d(x_i,x_b)<d(x_i,x_j)$ telling us that $x_b$ is located in 
this 
 ball too and 
it
  is smaller if there is a 
  triplet $d(x_i,x_j)<d(x_i,x_b)$ telling us that $x_b$ is not located in 
 this 
 ball. 

 Note that 
the similarity scores between $x_a$ and $x_b$ defined by 
$k_1$ or $k_2$ 
do not only depend on 
$d(x_a,x_b)$, 
but rather on the locations of $x_a$ and $x_b$ within $\dataset$ and 
on how the points in $\dataset$ are spread in the space since this affects how the various hyperplanes or balls are related to each other. Consider the example illustrated in Figure \ref{counterex}: Let $d(x_3,x_n)=1$ implying that $d(x_i,x_{i+1})=\Theta(1/n)$, $3\leq i<n$, and $d(x_1,x_2)>d(x_2,x_n)>d(x_1,x_n)>d(x_2,x_3)>d(x_1,x_3)>1$ be arbitrarily large. Although $x_1$ and $x_2$ are located at the maximum distance to each other, they satisfy  $d(x_1,x_i)<d(x_1,x_j)$ and $d(x_2,x_i)<d(x_2,x_j)$ for all $3\leq i<j\leq n$, and hence both $x_1$ and $x_2$ are jointly located in all the halfspaces obtained from these distance comparisons. 
We end up with 
$k_1(x_1,x_2)\rightarrow 1$, $n\rightarrow \infty$, assuming $k_1$ is computed based on all possible similarity triplets, all of which are correct.
 The distance between $x_3$ and $x_n$ is much smaller, but there are many points in  between them and the hyperplanes obtained from the distance comparisons with these points separate $x_3$ and $x_n$.  We end up with 
$k_1(x_3,x_n)\rightarrow -1$, $n\rightarrow \infty$. Depending on the 
task at hand, 
this may be desirable or not.

Let us examine the meaningfulness of our kernel functions by calculating them on  
five  visualizable data sets. 
Each of the first four data sets
 consists of 400 points in $\R^2$ and $d$ equals the Euclidean metric. The fifth data set  consists of 400 vertices of an undirected graph from a stochastic block model and $d$ equals the shortest path distance.
We computed 
$k_1$ and $k_2$ based on 
10\% of all possible similarity triplets (chosen uniformly at random 
from 
all 
triplets).
The results for the first two data sets are shown in Figure~\ref{fig_kernelmatrices}. The results for the remaining data sets are shown in Figure~\ref{fig_kernelmatrices2} in Section \ref{supp_figures} in the supplementary material. 
The first plot of a row shows the data set. The second plot shows the  distance matrix on the data set. Next, we can see the kernel matrices. The last plot of a row shows the similarity scores (encoded by color) based on $k_1$ between one fixed point 
(shown as a black cross) 
and the other points in the data set.
 Clearly, the kernel matrices reflect the block structures of the 
 distance matrices,  
and  
 the similarity scores between a fixed point and the other points tend to decrease as the distances to the fixed point increase. 
A situation like in the example of Figure~\ref{counterex} does not occur.

\setlength{\tabcolsep}{0.026cm}
\newcommand{\cellsi}{2.7cm}
\newcommand{\picsi}{0.1}
\newcommand{\picsiA}{2.3cm}
\newcommand{\cellhi}{0.6cm}
\newcommand{\picsiB}{0.12}
\renewcommand{\arraystretch}{1}
\begin{figure*}[t]
\centering
\begin{tabular}{>{\centering}p{\cellsi} >{\centering}p{\cellsi} 
>{\centering}p{\cellsi} >{\centering}p{\cellsi } >{\centering}p{2.8cm}}
\begin{minipage}[t][\cellhi][c]{\cellsi}
\begin{center}
400 points
\end{center}
\end{minipage} & 
\begin{minipage}[t][\cellhi][c]{\cellsi}
\begin{center}
Distance matrix
\end{center} 
\end{minipage} & 
\begin{minipage}[t][\cellhi][c]{\cellsi}
\begin{center}
$K_1$ 
\end{center}
\end{minipage} & 
\begin{minipage}[t][\cellhi][c]{\cellsi}
\begin{center}
$K_2$ 
\end{center}
\end{minipage} & 
\begin{minipage}[t][\cellhi][c]{2.8cm}
\begin{center}
Similarity scores
\end{center}
\end{minipage}  
\tabularnewline 
\rule{0pt}{20pt}
\hspace{-1.2mm}
\includegraphics[width=\picsiA]{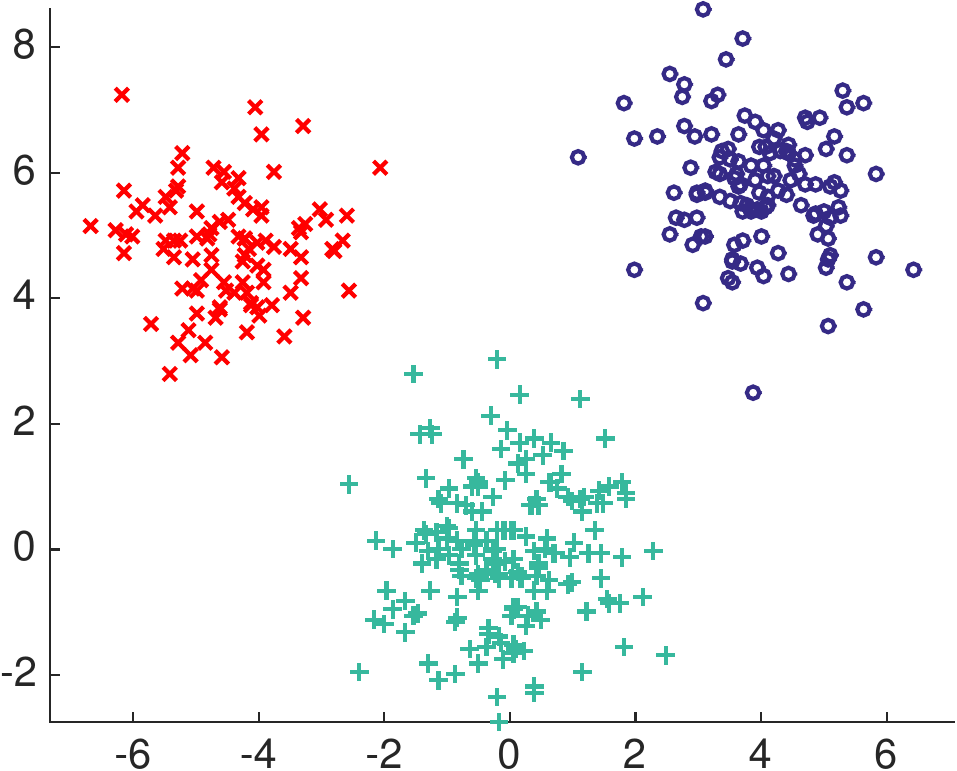} &
\includegraphics[width=\picsiA]{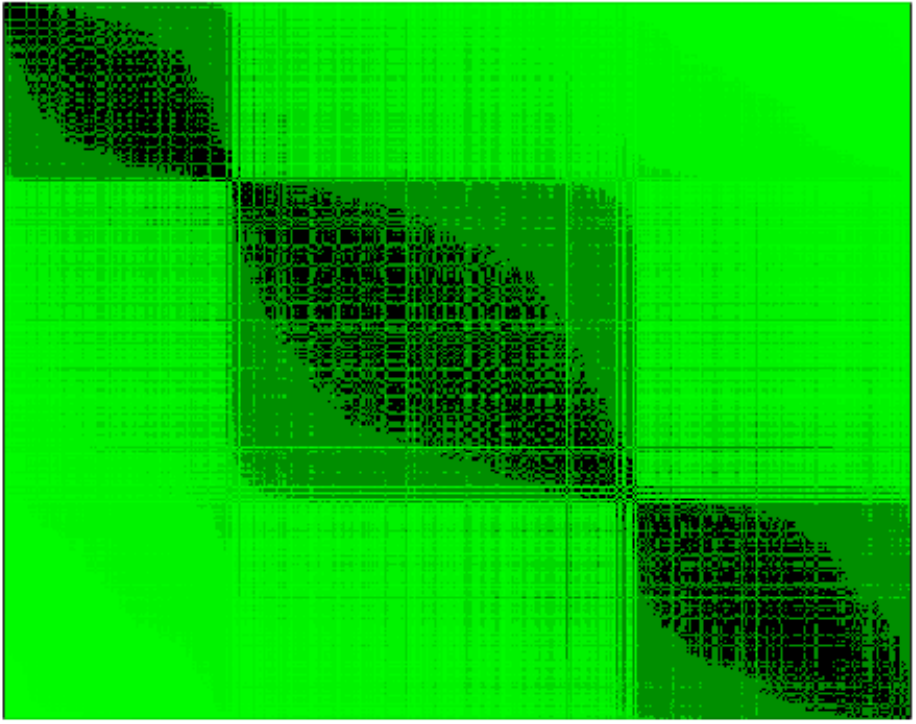} & 
\includegraphics[width=\picsiA]{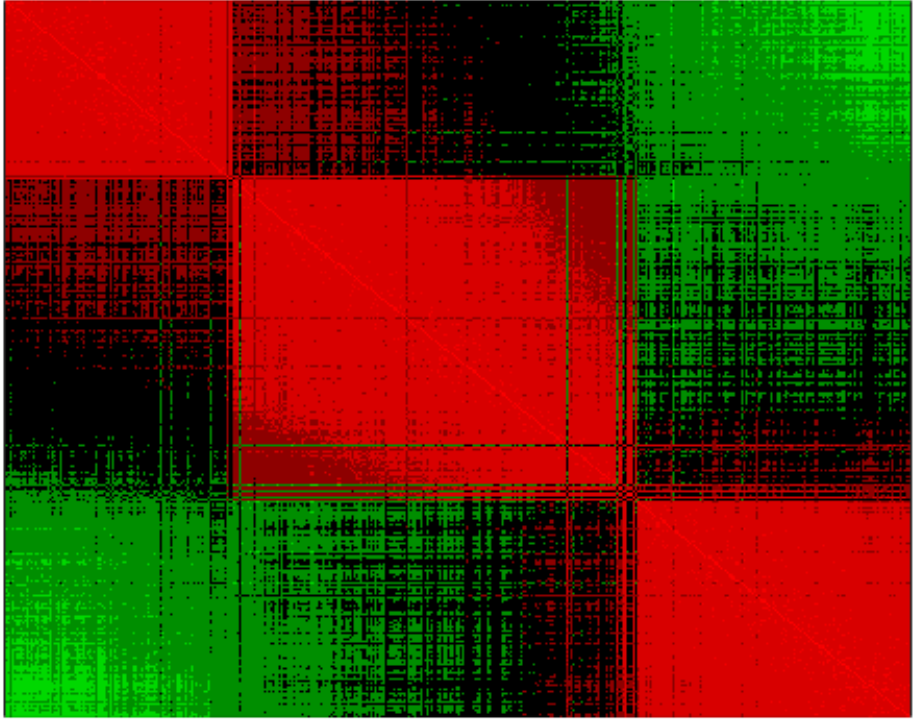} & 
\includegraphics[width=\picsiA]{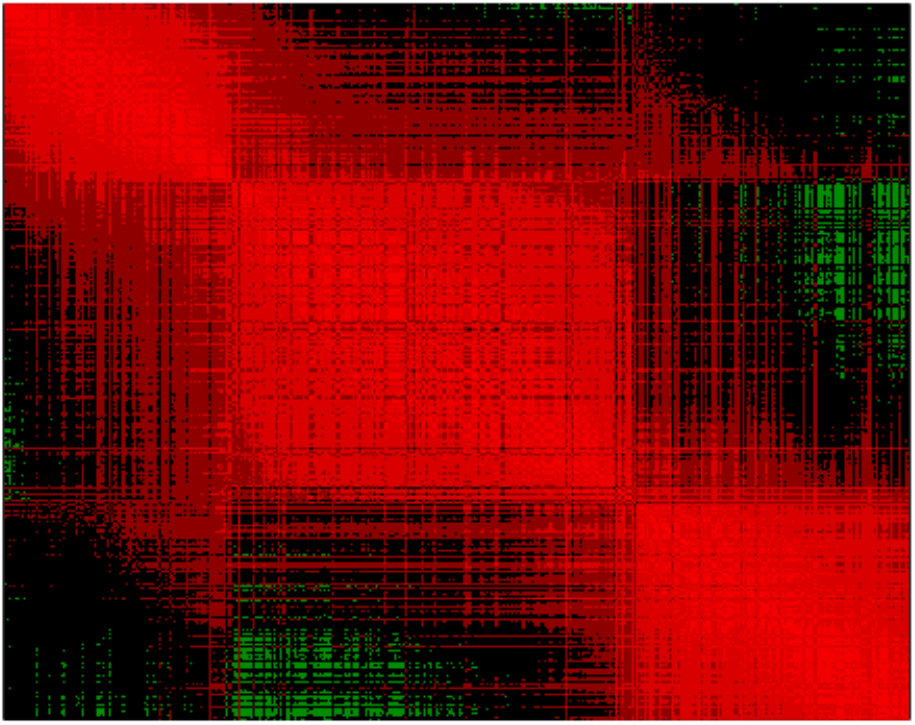} & 
\includegraphics[width=2.7cm]{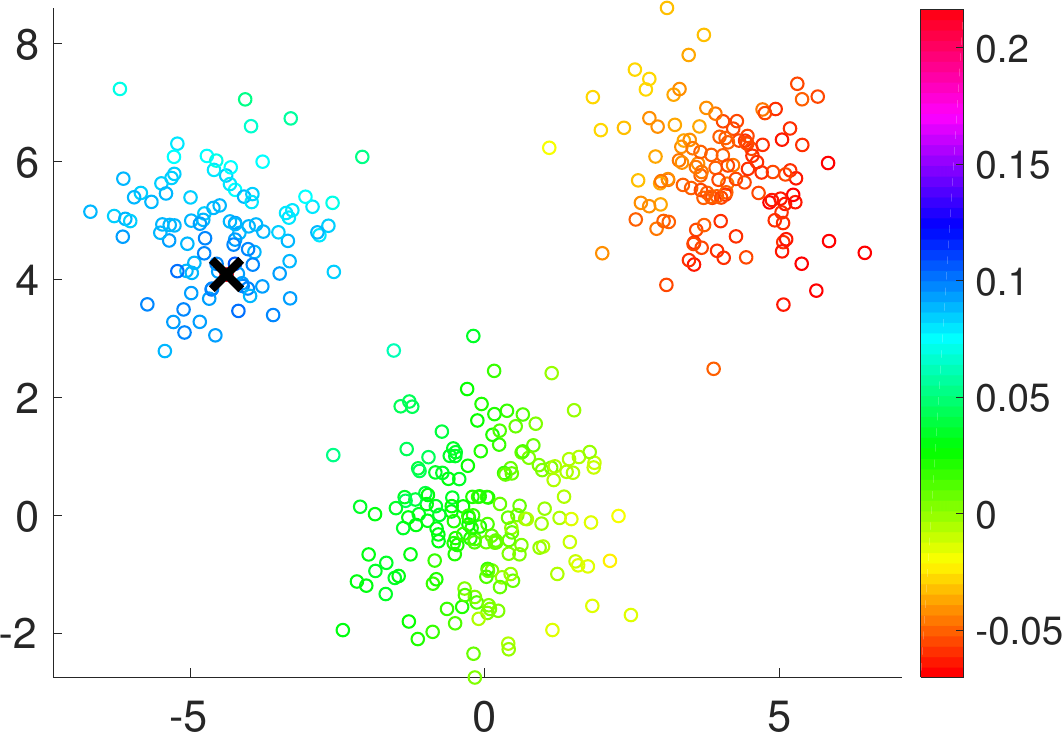}
\tabularnewline
\rule{0pt}{55pt}
\includegraphics[width=\picsiA]{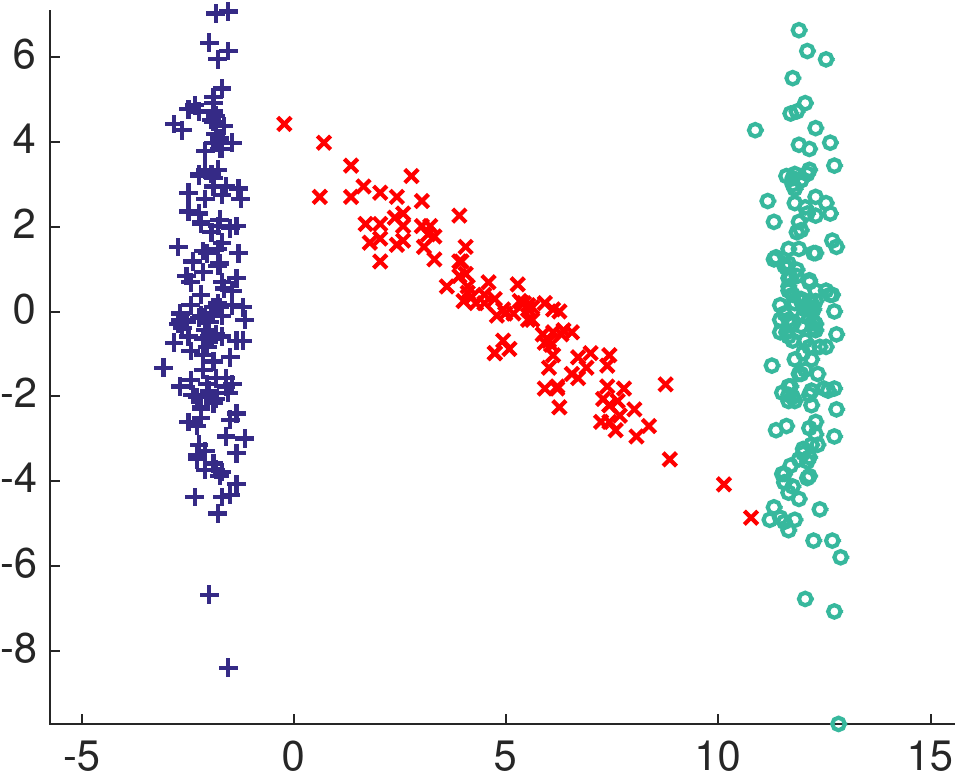} & 
\includegraphics[width=\picsiA]{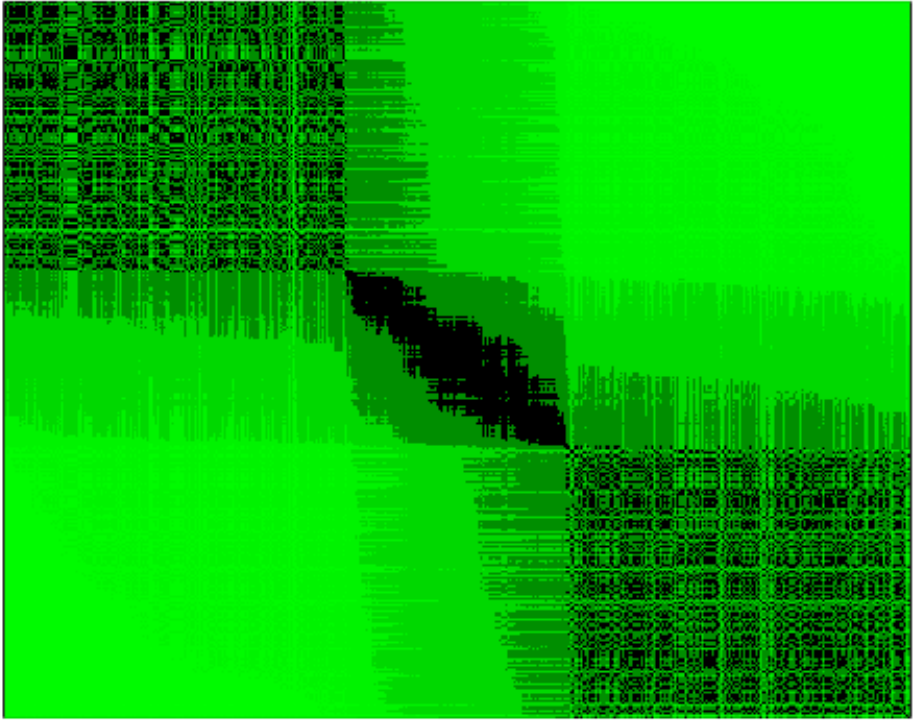} & 
\includegraphics[width=\picsiA]{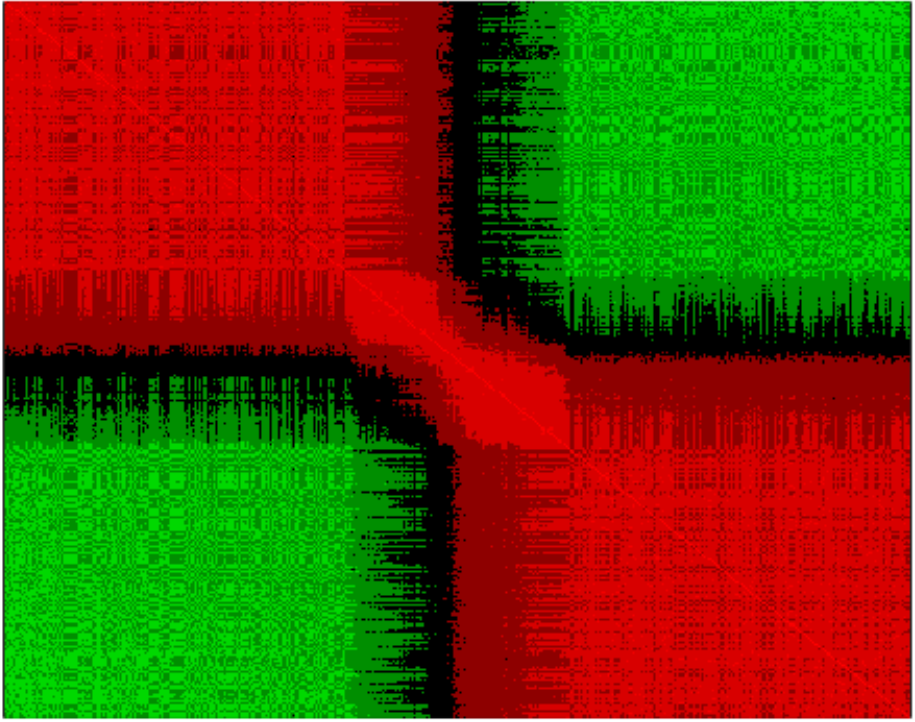} & 
\includegraphics[width=\picsiA]{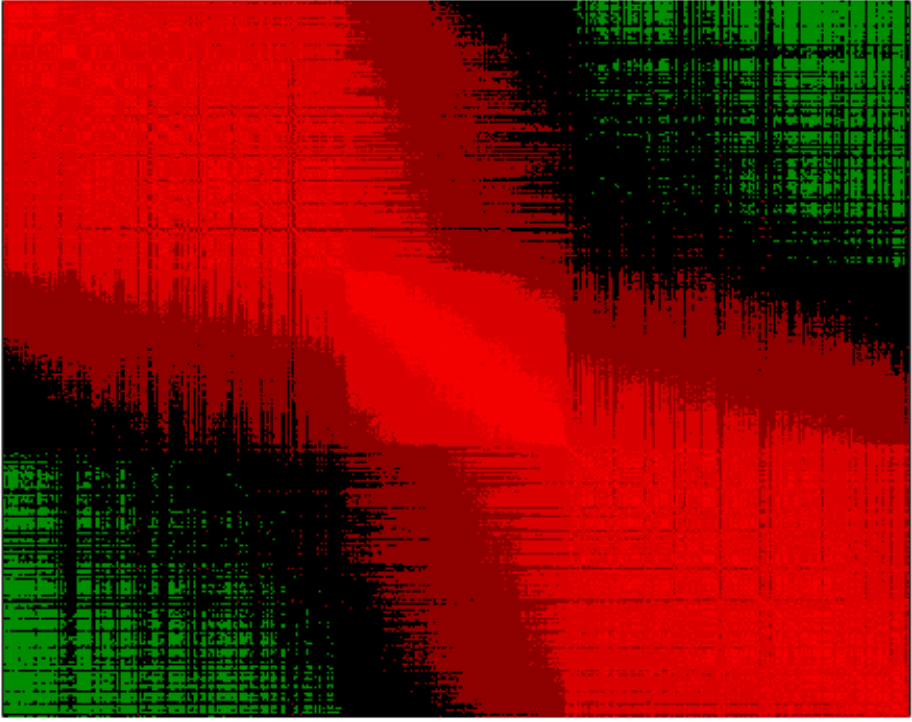} & 
\includegraphics[width=2.6cm]{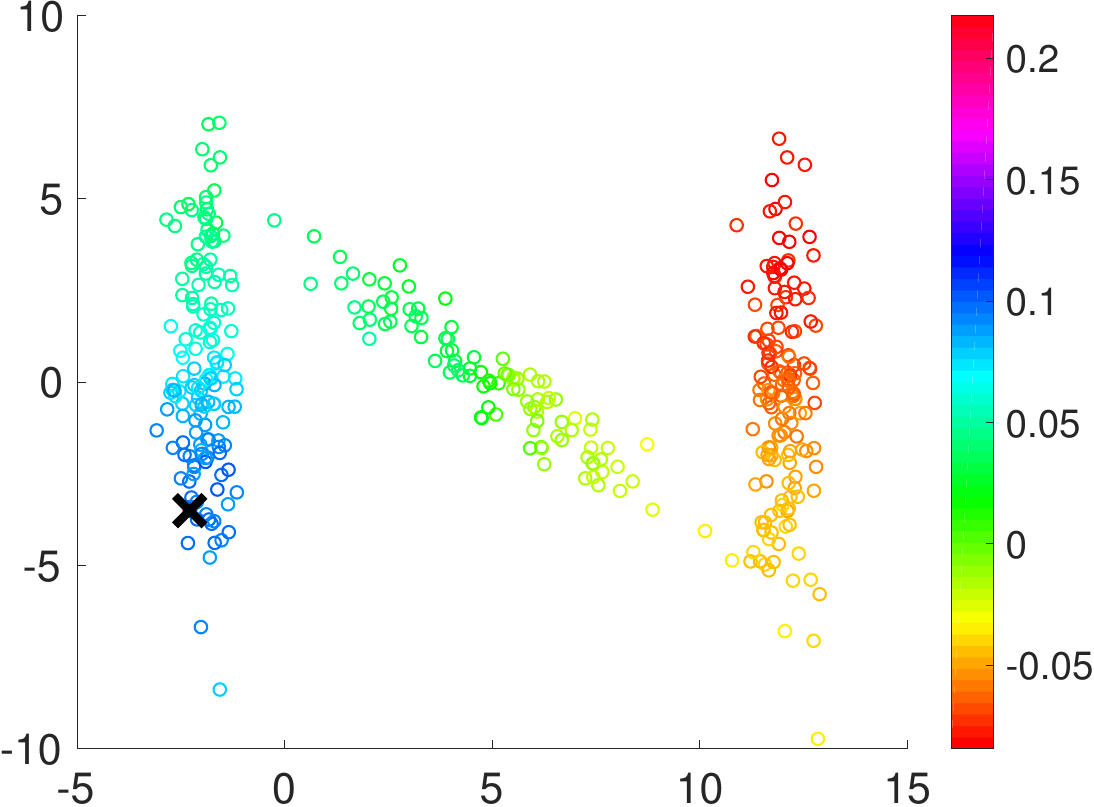}
\end{tabular}
\caption{Kernel matrices 
for two data sets, each  consisting of 400 points,  
based on 10\% of all similarity triplets. 
1st 
plot of a row:
Data points. 
2nd 
plot:
Distance matrix. 
3rd / 4th plot: Kernel 
matrix  
for $k_1$~/~$k_2$. 
6th 
plot:
Similarity scores 
between a fixed point 
and the other points~(for~$k_1$).}\label{fig_kernelmatrices}
\end{figure*}

\subsection{Landmark design}\label{section_landmark}
 Our kernel functions are designed as to extract information from an arbitrary collection $\mathcal{S}$ of similarity triplets. However, by construction, a single 
 triplet is useless, and what matters is the concurrent presence of two triplets: $k_1(x_a,x_b)$ is only affected by pairs of 
 triplets answering $d(x_a,x_i)\stackrel{\mbox{\tiny ?}}{\scriptstyle <}d(x_a,x_j)$ and $d(x_b,x_i)\stackrel{\mbox{\tiny ?}}{\scriptstyle <}d(x_b,x_j)$, 
 while $k_2(x_a,x_b)$ is only affected by pairs of 
 triplets answering~\eqref{fragenpaar}. Hence, when we 
can  
 choose which dissimilarity comparisons 
 of the form
  \eqref{parwise_comp} 
are 
  evaluated for creating $\mathcal{S}$ (e.g., in crowdsourcing), we should 
  aim at maximizing the number of appropriate pairs of 
  triplets. This can 
easily   
  be achieved by means of a landmark design inspired from landmark multidimensional scaling \citep{landmark_mds}: 
  We choose  a small subset of landmark objects 
 $\mathcal{L}\subseteq \dataset$. Then, for $k_1$,   
 only 
 comparisons of the form $d(x_i,x_j)\stackrel{\mbox{\tiny ?}}{\scriptstyle <}d(x_i,x_k)$ 
with $x_i\in\dataset$ and $x_j,x_k\in \mathcal{L}$ are evaluated. For~$k_2$,
only comparisons  of the form 
 $d(x_j,x_i)\stackrel{\mbox{\tiny ?}}{\scriptstyle <}d(x_j,x_k)$ 
 with $x_i\in\dataset$ and $x_j,x_k\in \mathcal{L}$ 
are evaluated. The landmark objects can be chosen either randomly or, if available, based on additional knowledge about $\dataset$ and the task at hand.

\subsection{Computational complexity}

\paragraph{General $\bm{\mathcal{S}}$}
A naive implementation of our kernel functions 
explicitly computes the feature vectors $\Phi_{k_1}(x_i)$ or $\Phi_{k_2}(x_i)$, $i=1,\ldots,n$, and subsequently calculates the kernel matrix $K$ by means of \eqref{scalarp_K1} or~\eqref{scalarp_K2}. In doing so, we store the feature vectors in the feature matrix 
$\Phi_{k_1}(\dataset)=(\Phi_{k_1}(x_i))_{i=1}^n\in\R^{\binom{n}{2}\times n}$
or 
$\Phi_{k_2}(\dataset)=(\Phi_{k_2}(x_i))_{i=1}^n\in\R^{n^2\times n}$. 
Proceeding this way is straightforward and simple, requiring to go through $\mathcal{S}$ only once, but comes with a computational cost of $\mathcal{O}(|\mathcal{S}|+n^4)$ operations. Note that the number of different distance comparisons of the form \eqref{parwise_comp} is $\mathcal{O}(n^3)$ and hence one might expect that $|\mathcal{S}|\in\mathcal{O}(n^3)$ and $\mathcal{O}(|\mathcal{S}|+n^4)=\mathcal{O}(n^4)$. By performing \eqref{scalarp_K1} or \eqref{scalarp_K2} in terms of matrix multiplication $\Phi_{k_1}(\dataset)^T\cdot \Phi_{k_1}(\dataset)$ or $\Phi_{k_2}(\dataset)^T\cdot \Phi_{k_2}(\dataset)$ and 
applying Strassen's algorithm \citep{fast_matrix_multiplication} 
one can 
reduce the number of operations to $\mathcal{O}(|\mathcal{S}|+n^{3.81})$, but still this is infeasible for many data sets. 
Infeasibility for large data sets, however, is even more the case for ordinal embedding algorithms, which are the current state-of-the-art method for solving 
machine learning 
problems based on similarity triplets. All existing ordinal embedding algorithms iteratively solve an optimization problem. For none of these algorithms theoretical bounds for their complexity are available in the literature, but it is 
widely 
known that their running times are prohibitively high \citep{Heim2015,kleindessner16}.

\paragraph{Landmark design}
If we know that $\mathcal{S}$ contains only dissimilarity comparisons involving landmark objects,  
we can adapt the feature matrices such that $\Phi_{k_1}(\dataset)\in\R^{\binom{|\mathcal{L}|}{2}\times n}$
or
$\Phi_{k_2}(\dataset)\in\R^{|\mathcal{L}|^2\times n}$ and reduce the number of operations to $\mathcal{O}(|\mathcal{S}|+\min\{|\mathcal{L}|^2,n\}^{\log_2(7/8)}|\mathcal{L}|^2n^2)$,
which is $\mathcal{O}(|\mathcal{S}|+|\mathcal{L}|^{1.62}n^2)$ if $|\mathcal{L}|^2\leq n$. Note that in this case 
we might 
expect that $|\mathcal{S}|\in\mathcal{O}(|\mathcal{L}|^2n)$.

\vspace{1.65ex} 
In both cases, whenever the number of given similarity triplets~$|\mathcal{S}|$ is small compared to the number of all different distance comparisons under consideration, 
the feature matrix $\Phi_{k_1}(\dataset)$ or $\Phi_{k_2}(\dataset)$ is sparse with only $\mathcal{O}(|\mathcal{S}|)$ non-zero entries and methods for sparse matrix multiplication decrease computational complexity \citep{gustavson1978,kaplan2006}.

\section{Related work}\label{section_related_work}

Similarity triplets 
are a special case of answers to the 
general 
dissimilarity 
comparisons 
$d(A,B)\stackrel{\mbox{\tiny ?}}{\scriptstyle <}d(C,D)$, $A,B,C,D\in\mathcal{X}$.
We refer to any collection of answers to 
these
general
comparisons
as 
ordinal data.
In recent years, ordinal data 
 has become 
 popular in machine learning.  
 Among the work on ordinal data in general \citep[see][for references]{kleindessner14,kleindessner16}, 
similarity triplets have been paid particular attention: 
\citet{JamNow11} deal with the 
question of how many similarity triplets are required for uniquely determining an ordinal embedding of Euclidean data. This work has been carried on and generalized by \citet{jamieson_finite}. Algorithms for constructing an ordinal embedding based on similarity triplets (but not 
on general ordinal data) are proposed in \citet{TamuzEtal2011}, \citet{stoch_trip_embed}, \citet{amid2017},  and \citet{jamieson_finite}. 
\citet{crowdmedian} present a method for medoid estimation based on statements ``Object~$A$ is the outlier within the triple of objects $(A,B,C)$'', which correspond to the two similarity triplets $d(B,C)<d(B,A)$ and $d(C,B)<d(C,A)$. 
\citet{ukkonen_density} use the same kind of statements for density estimation 
and \citet{ukkonen_corrclust} uses them for clustering.  
\citet{wilber2014} examine how 
to minimize time and costs when collecting similarity triplets via crowdsourcing. Producing a number of ordinal embeddings at the same time, each corresponding to a different dissimilarity function based on which a 
comparison~\eqref{parwise_comp} might have been evaluated, is studied in \citet{Ukkonen_multiview}. In \citet{Heim2015}, one of the algorithms by \citet{stoch_trip_embed} is adapted 
from the batch setting
 to an online setting, in which similarity triplets are observed in a sequential way, using stochastic gradient descent. In \citet{kleindessner16}, we propose algorithms for 
medoid estimation, outlier detection, classification, and clustering
based on statements ``Object $A$ is the most central object within 
$(A,B,C)$'', which comprise the 
two 
similarity triplets $d(B,A)<d(B,C)$ and $d(C,A)<d(C,B)$.  
Finally, \citet{siavash_2017} study the problem of efficient nearest neighbor search based on similarity triplets.
There 
is also a number of
 papers that consider similarity triplets as side information to vector data \citep[e.g.,][]{
 SchulzJoachims03,
 McfLan11,Wilber2015}.

\section{Experiments}\label{section_experiments}
We performed 
experiments 
that demonstrate the 
usefulness of our 
kernel functions. We first apply them to three small image data sets for which similarity triplets have been gathered via crowdsourcing. We then study them more systematically and compare them to an ordinal embedding approach in clustering tasks on 
subsets of USPS and MNIST digits 
using
 synthetically generated 
 triplets.

\begin{figure}[t]
\centering  
\includegraphics[width=0.53\linewidth]{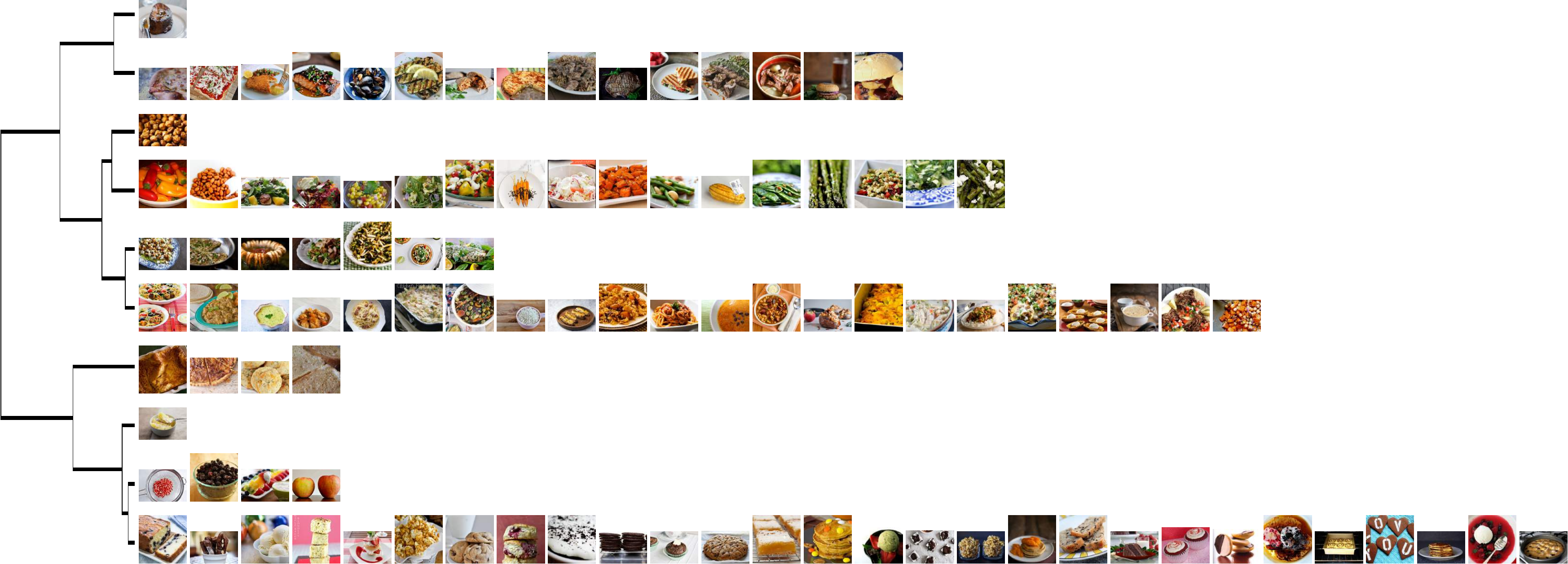}
\hspace{0.9cm}
\includegraphics[width=0.39\linewidth]{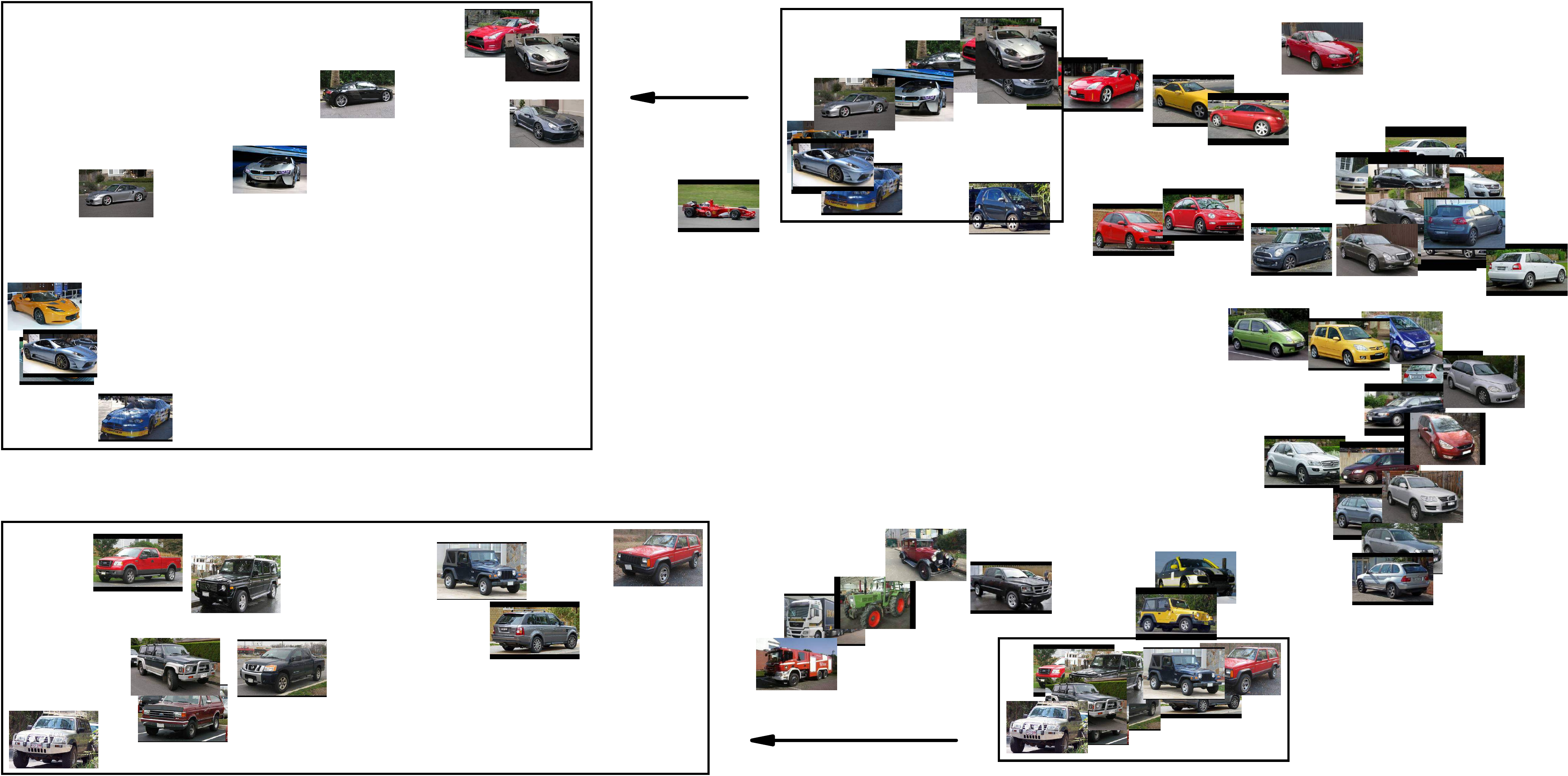}
\caption{Best viewed magnified on screen. Left: Clustering 
of the food data set. Part of the dendrogram obtained from complete-linkage clustering 
using 
$k_1$. Right: Kernel PCA on the car data set 
based on 
the kernel function~$k_2$.}\label{image_food_clustering}
\end{figure}

\subsection{Crowdsourced similarity triplets}\label{real_exp}

In this section we present experiments on real crowdsourcing data that show 
that our kernel functions can capture the structure of a data set. 
Note that 
for the following data sets there is no ground truth available and hence
there is no way other than visual inspection 
for evaluating our results. 

\paragraph{Food data set}
We applied the kernelized version of complete-linkage clustering 
based on our 
kernel function $k_1$ to the food data set introduced in \citet{wilber2014}. 
This data set consists of 100 images\footnote{According to \citeauthor{wilber2014}, 
the data set contains copyrighted material under the educational fair use exemption to the U.S. copyright law.} 
of a wide range of foods and comes with 
190376 (unique) similarity triplets, which 
contain 
 9349  pairs of contradicting triplets.
Figure \ref{image_food_clustering} (left) shows a part of the dendrogram that we obtained. 
Each of the ten clusters depicted there contains pretty homogeneous images. 
For example, the 
fourth row only shows vegetables and salads whereas the ninth row only shows fruits and the last row only shows desserts. 
To give an impression of accelerated running time of our approach compared to an 
ordinal embedding approach: computation of $k_1$ or $k_2$ on this data set took about 0.1 seconds while computing an 
ordinal 
embedding using the GNMDS algorithm \citep{AgarwalEtal07} took 18 seconds 
(embedding dimension equaling two; all computations performed in Matlab---see Section~\ref{syn_exp} for details; the 
embedding 
 is shown in 
Figure \ref{ordemb_food} in Section~\ref{supp_figures} in the supplementary material).

\paragraph{Car data set}
We applied kernel PCA 
\citep{kernel_pca} 
based on our kernel function $k_2$ 
to the car data set, which 
we have introduced in \citet{kleindessner16}. It consists of 60 images of cars.  
For this data set we have collected statements of the kind ``Object $A$ is the most central object within $(A,B,C)$'',
meaning that $d(B,A)<d(B,C)$ and $d(C,A)<d(C,B)$, via crowdsourcing. 
We ended up with 13514 similarity triplets, of which 12502 were unique. 
The projection of the car data set onto the first two kernel principal components can be seen in Figure \ref{image_food_clustering} (right). 
The result looks 
reasonable, with the cars 
arranged in groups of sports cars (top left), ordinary cars (middle right) and  
off-road/sport utility vehicles (bottom left). Also within these groups there is some reasonable structure. 
For example, the race-like sports cars are located near to each other and close to the Formula One car,
and the sport utility vehicles from German manufacturers are placed next to each~other.

\paragraph{Nature data set}
We performed similar experiments on the nature data set introduced in \citet{crowdmedian}. The results are 
presented in Section \ref{sec_supp_nature} in the supplementary material.

We would like to 
discuss 
a question raised by one of the reviewers: in our setup (see Section~\ref{sec_intro}), we assume that similarity triplets are noisy evaluations of dissimilarity comparisons 
\eqref{parwise_comp}, 
where $d$ is some 
fixed 
dissimilarity function. This leads to our (natural)  
way of dealing with contradicting similarity triplets as described in Section \ref{sec_kernel_descr}. In a different setup one could drop the dissimilarity function $d$ and 
consider similarity triplets as  elements of some binary relation on 
$\dataset \times \dataset$ that is not necessarily transitive or antisymmetric. In the latter setup it is not clear whether our way of dealing with contradicting  triplets is the right thing to do. 
However, 
we believe that the experiments of this section 
show that our setup is valid in a wide range of scenarios and our approach works in practice.

\subsection{Synthetically generated triplets}\label{syn_exp}

We studied our kernel functions 
with respect to the number of input similarity triplets that they require in order to produce a valuable solution
in clustering tasks. 
We found that in the scenario of a general collection $\mathcal{S}$ of 
triplets our approach is highly superior compared to an ordinal embedding approach 
in terms of 
running time, but on most data sets it is inferior regarding the required number of 
triplets. The full benefit of our kernel functions emerges in a landmark design. There our approach can compete with an embedding approach in terms of  
the required number of triplets 
and is 
so much faster as to being easily applicable to large data sets to which 
ordinal 
embedding algorithms are not. 
In this section we want to demonstrate this claim.  
We studied $k_1$ and $k_2$ in a landmark design by applying kernel $k$-means clustering \citep{kernel_k_means} to subsets of USPS and MNIST digits, respectively. 
Collections~$\mathcal{S}$ of  
similarity  
triplets were generated as follows: We chose a certain number of landmark objects uniformly at random from all objects of the data set under consideration. Choosing $d$ as the Euclidean metric, we created answers to all possible distance comparisons with the landmark objects 
as explained in Section~\ref{section_landmark}. Answers were incorrect with some probability $0\leq ep\leq 1$ independently of each other. 
From the set of all answers we chose 
triplets in $\mathcal{S}$ uniformly at random without replacement.
We compared our approach to an 
ordinal 
embedding approach with ordinary $k$-means clustering. 
 We tried the GNMDS \citep{AgarwalEtal07}, the CKL \citep{TamuzEtal2011}, and the t-STE \citep{stoch_trip_embed} 
embedding algorithms in the Matlab implementation made available by \citet{stoch_trip_embed}.
In doing so, we set all parameters except the embedding dimension 
to the provided default parameters. The parameter $\mu$ of the CKL algorithm was set to 0.1 since we observed good results with this value. Note that in these unsupervised clustering tasks there is no immediate way of performing cross-validation for choosing parameters. 
We compared to the embedding algorithms in two scenarios: in one case they were provided the same 
triplets as input as our kernel functions, in the other case (denoted by the additional ``rand'' in the plots) they were provided a same number of triplets chosen uniformly at random with replacement from all possible triplets 
(no landmark design) 
and incorrect with the same 
probability 
$ep$. 
For further comparison, we considered ordinary $k$-means applied to the original point set and a random clustering. 
We always provided the correct number of clusters as input, and set the number of replicates in $k$-means and kernel $k$-means to  
five  
and the maximum number of iterations to 100. 
For assessing the quality of a clustering we computed its purity  
\citep[e.g.,][]{manning_intro_inf_ret},
which
measures the accordance with the known ground truth partitioning according to  the digits' values. 
A high purity value indicates a good clustering. 
Note that the limitation for the scale of our experiments only comes from the running time of the embedding algorithms and not from our kernel functions. Still, in terms of the number of data points our experiments are comparable or actually even superior to all the papers on ordinal embedding cited in Section \ref{section_related_work}. In terms of the number of  similarity triplets per data point, we used comparable numbers of triplets.

\newcommand{\mysam}{0.2}
\newcommand{\siko}{3.35cm}  
\newcommand{\sikolop}{2.2cm}
\newcommand{\spo}{0.01cm}
\begin{figure*}[t]
\centering

\includegraphics[width=\siko]{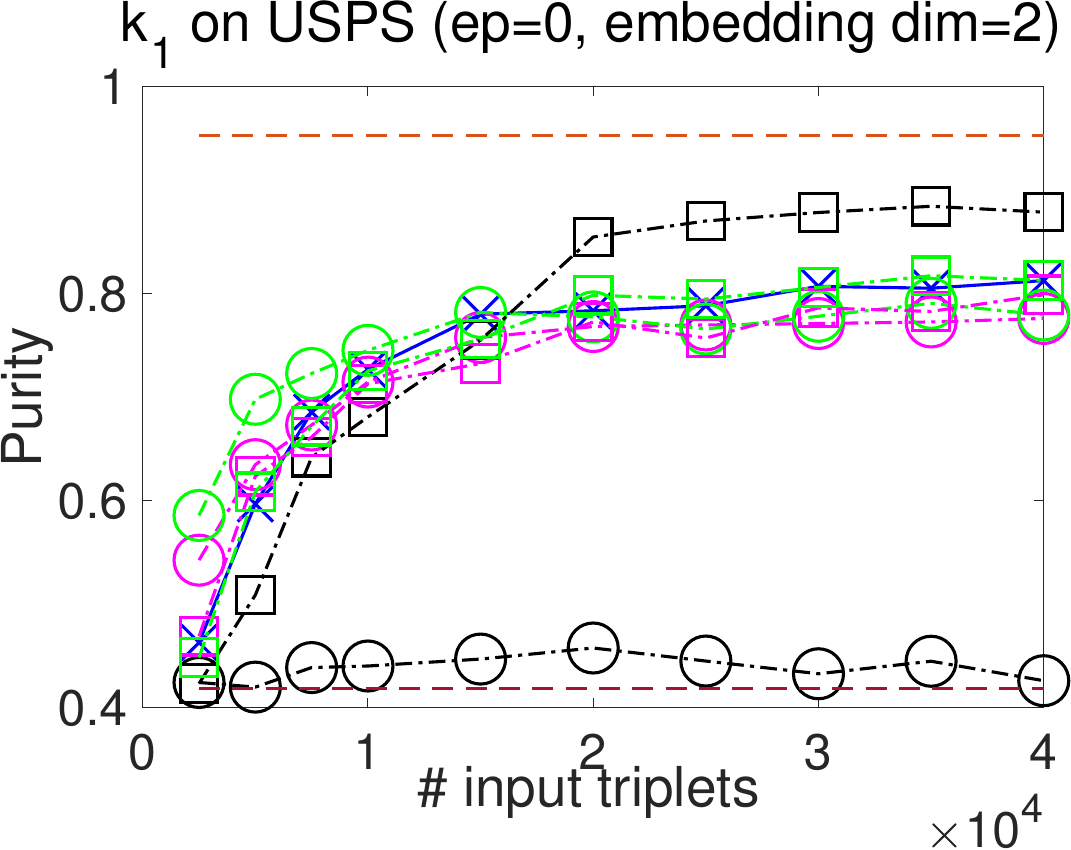} \hspace{\spo}
\includegraphics[width=\siko]{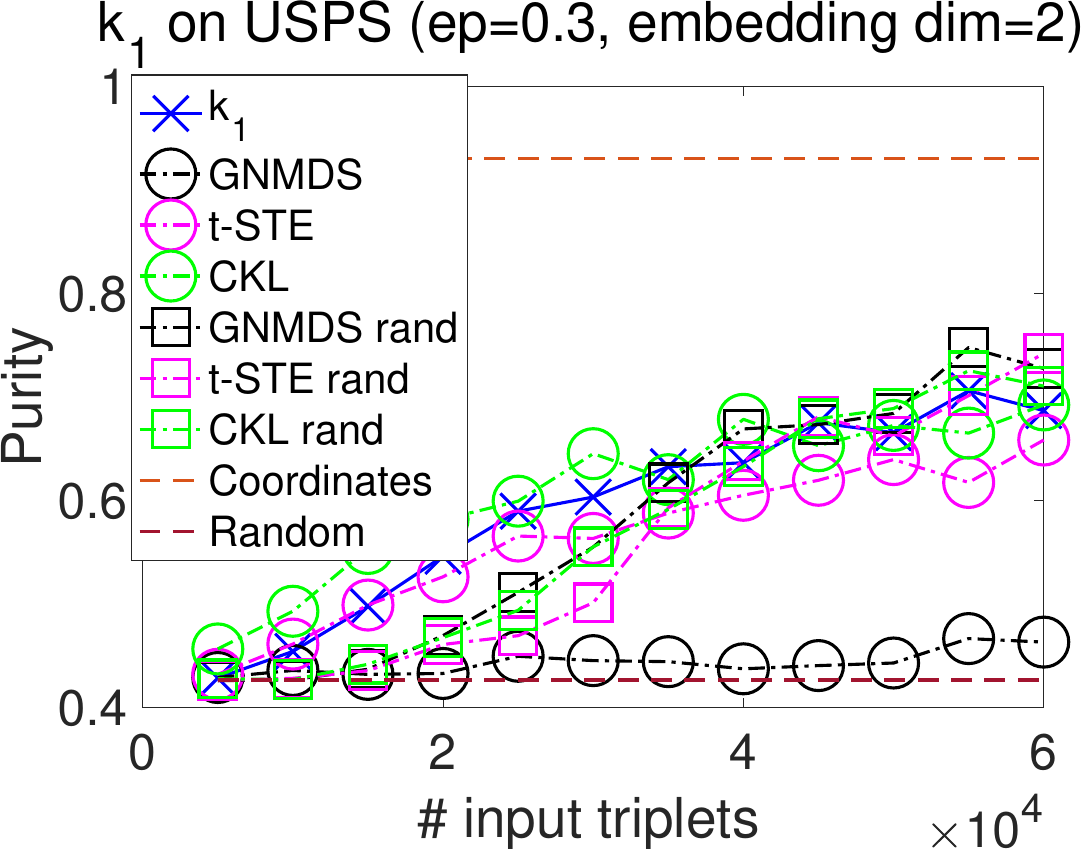} \hspace{\spo}
\includegraphics[width=\siko]{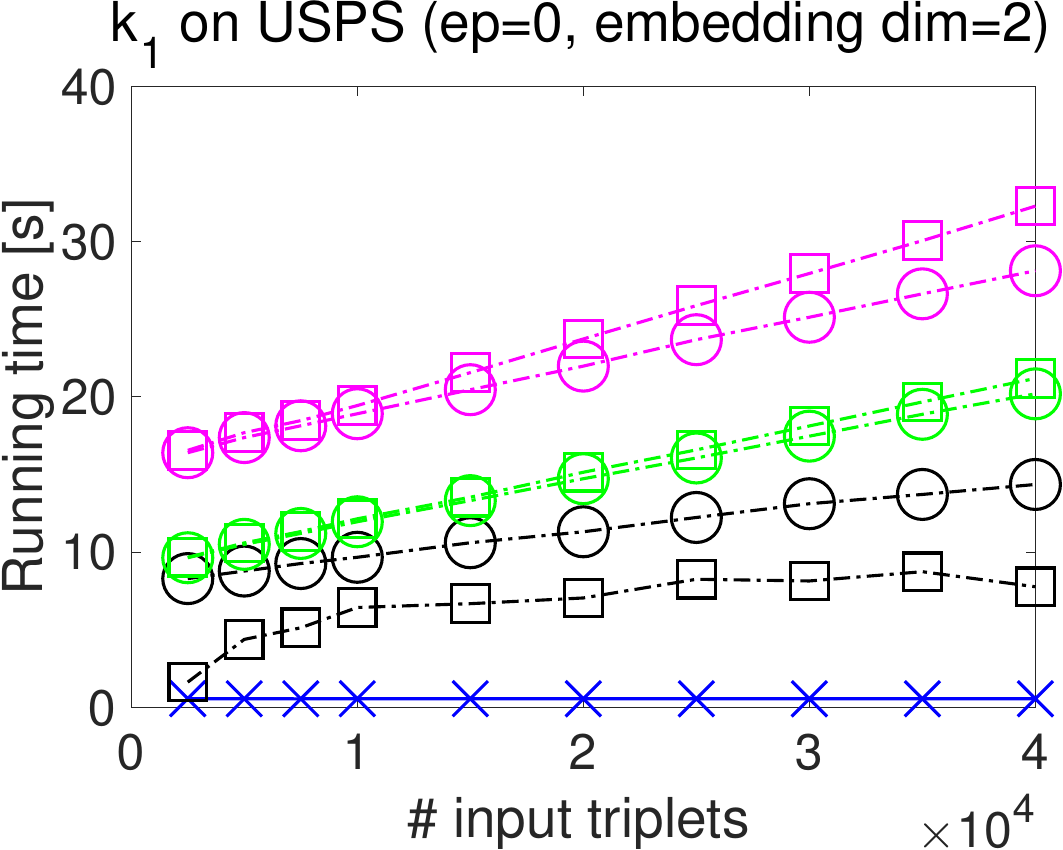}  \hspace{\spo}
\includegraphics[width=\siko]{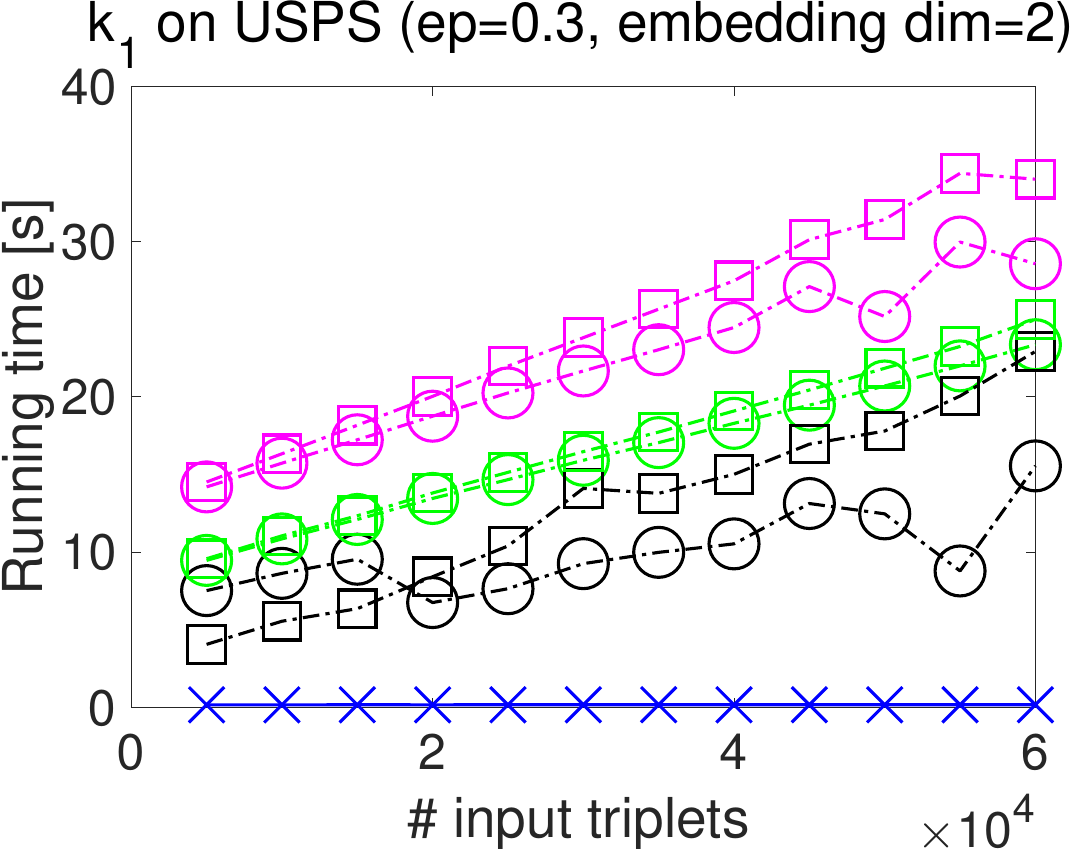}

\vspace{1mm}
\includegraphics[width=\siko]{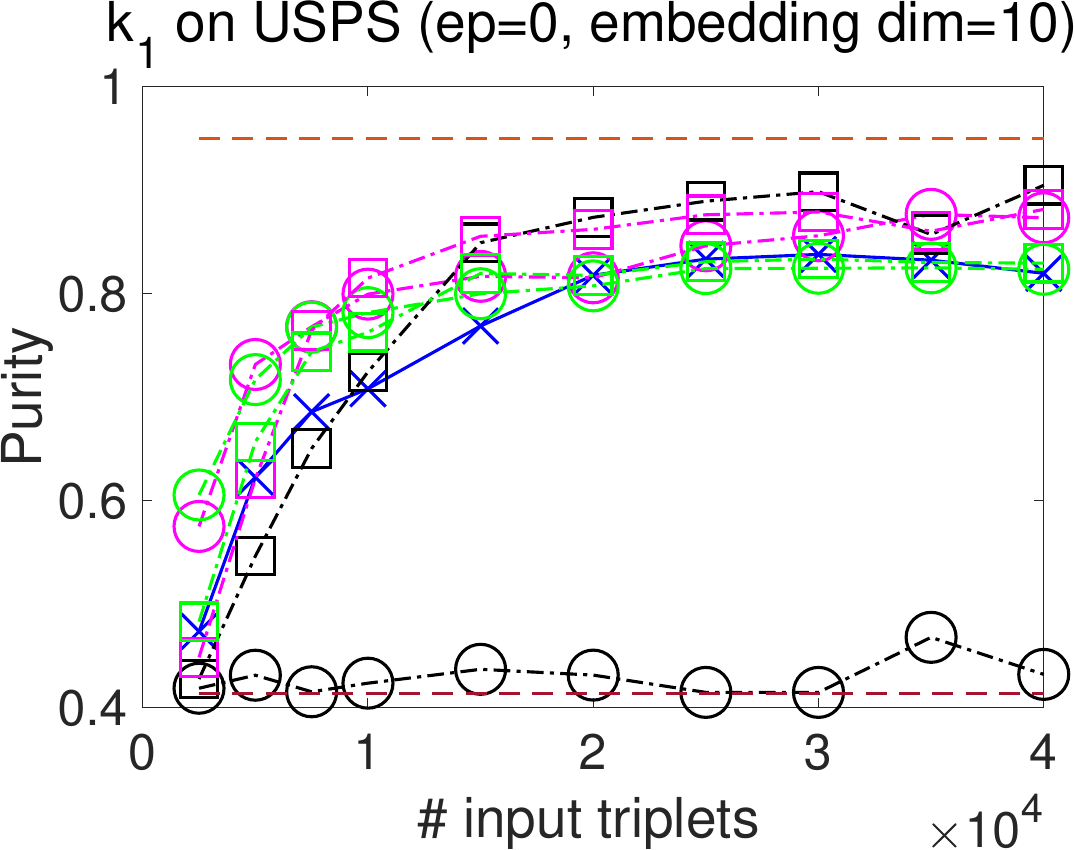} \hspace{\spo}
\includegraphics[width=\siko]{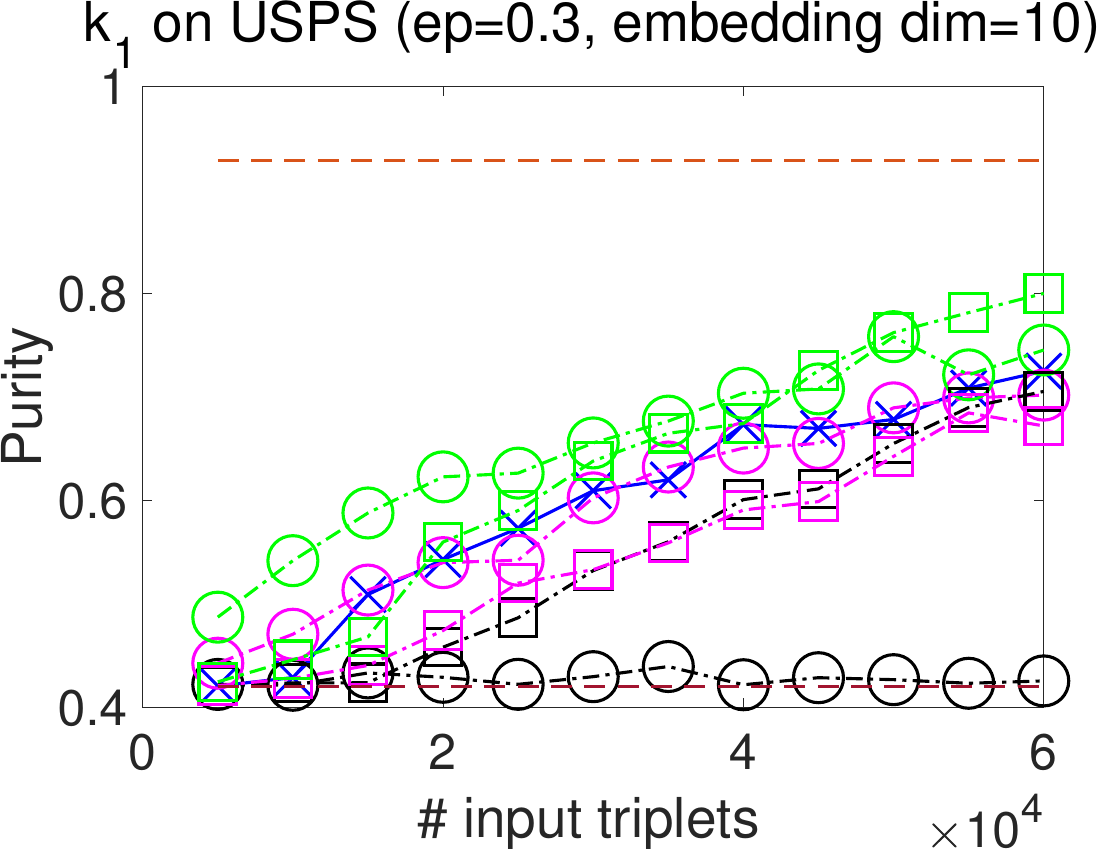} \hspace{\spo}
\includegraphics[width=\siko]{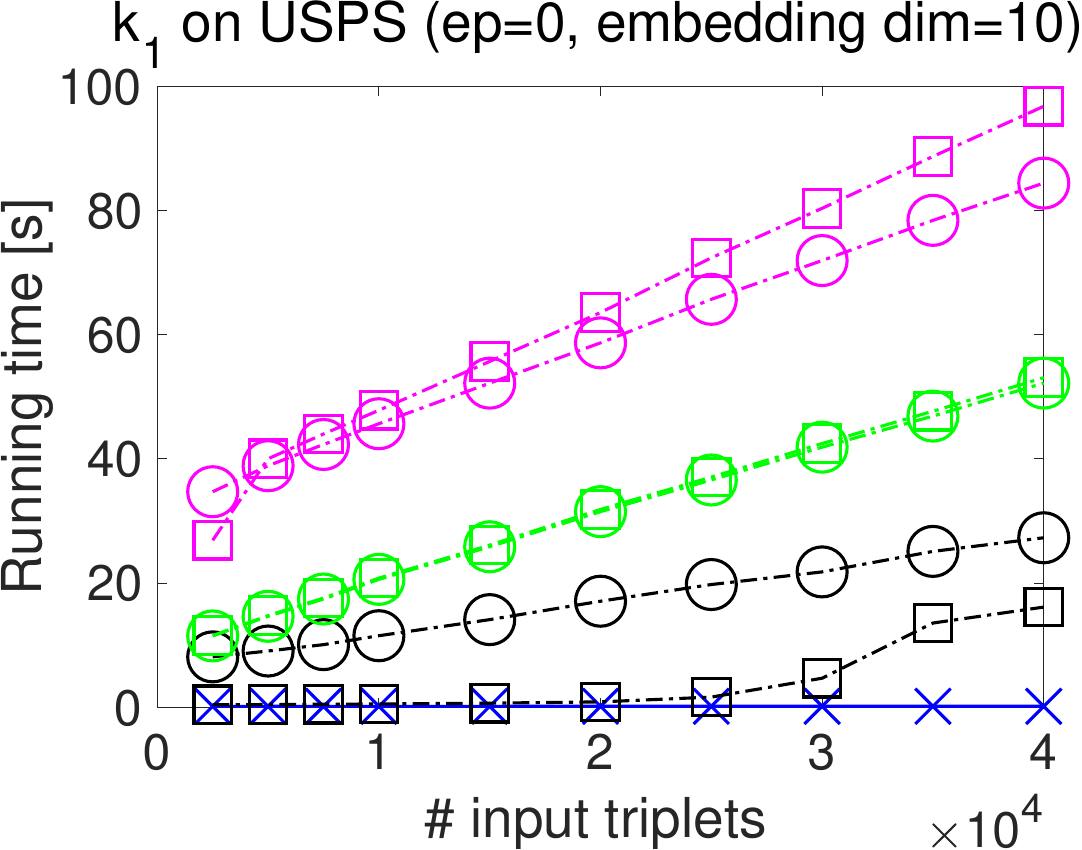}  \hspace{\spo}
\includegraphics[width=\siko]{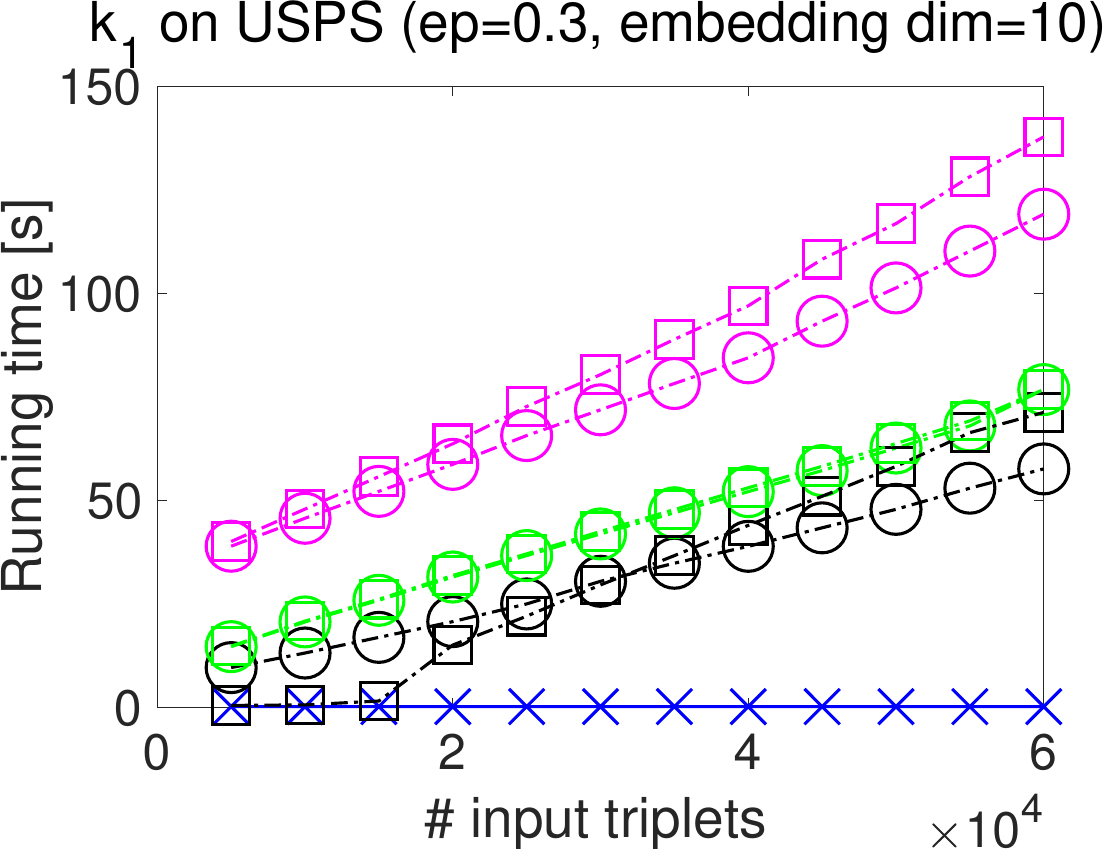}

\vspace{1mm}
\includegraphics[width=\siko]{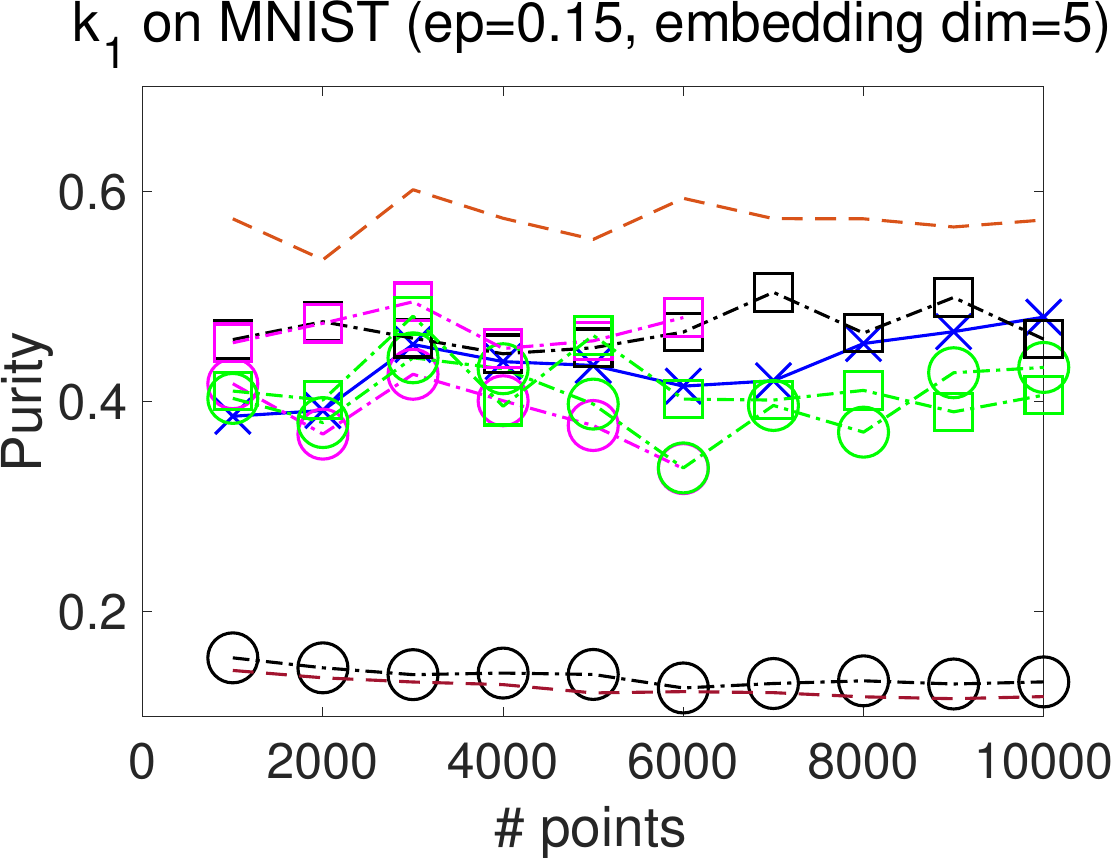} \hspace{\spo}
\includegraphics[width=\siko]{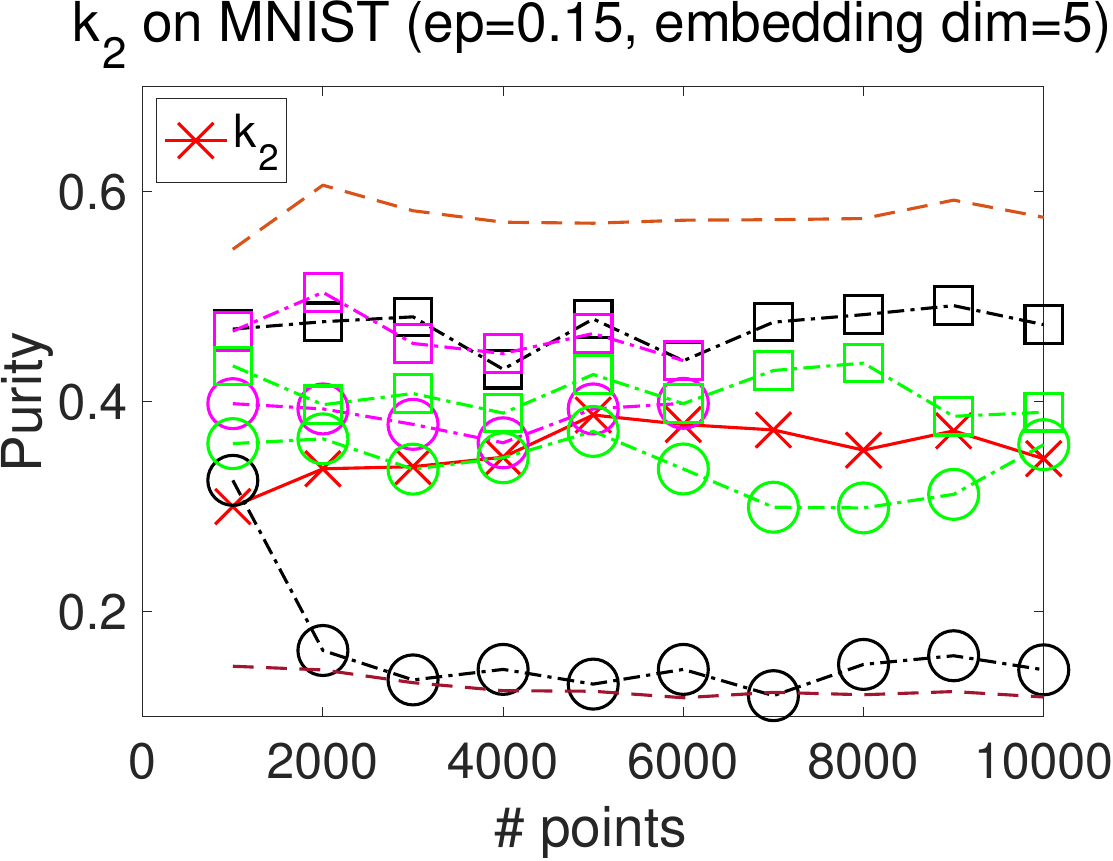} \hspace{\spo}
\includegraphics[width=\siko]{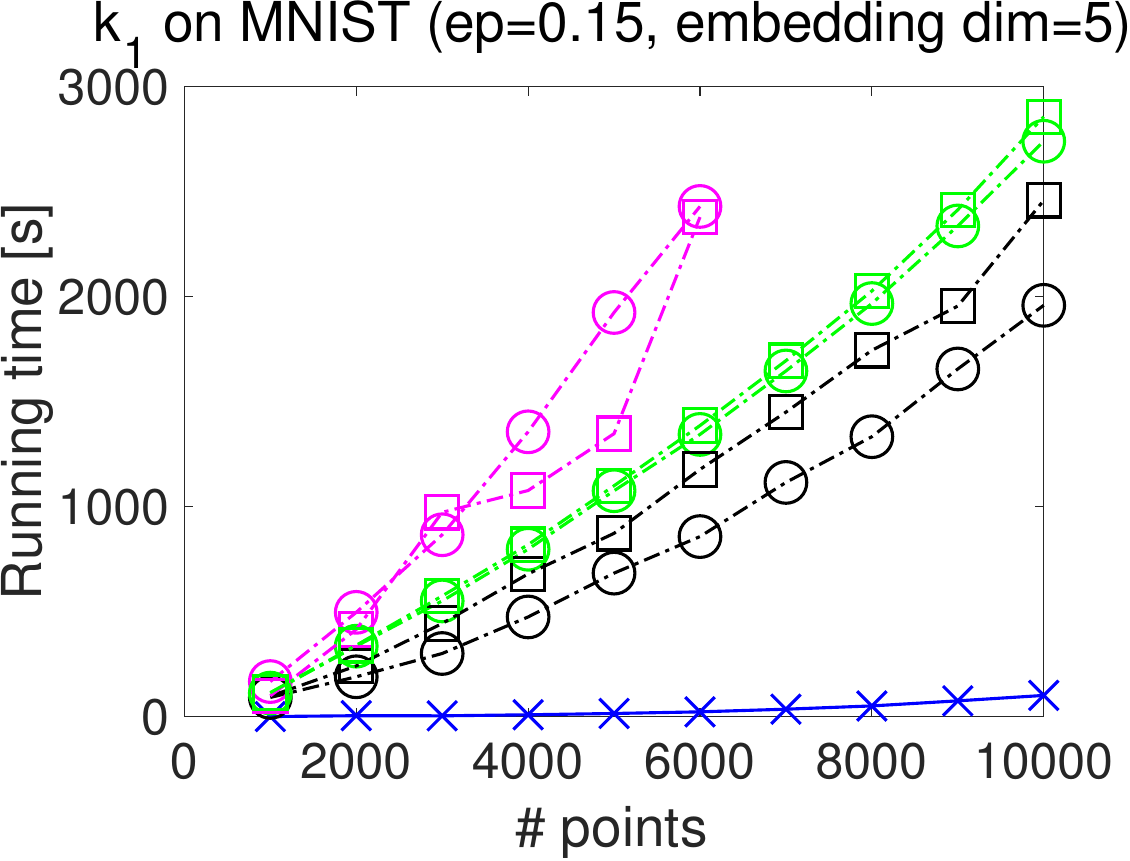}\hspace{\spo}
\includegraphics[width=\siko]{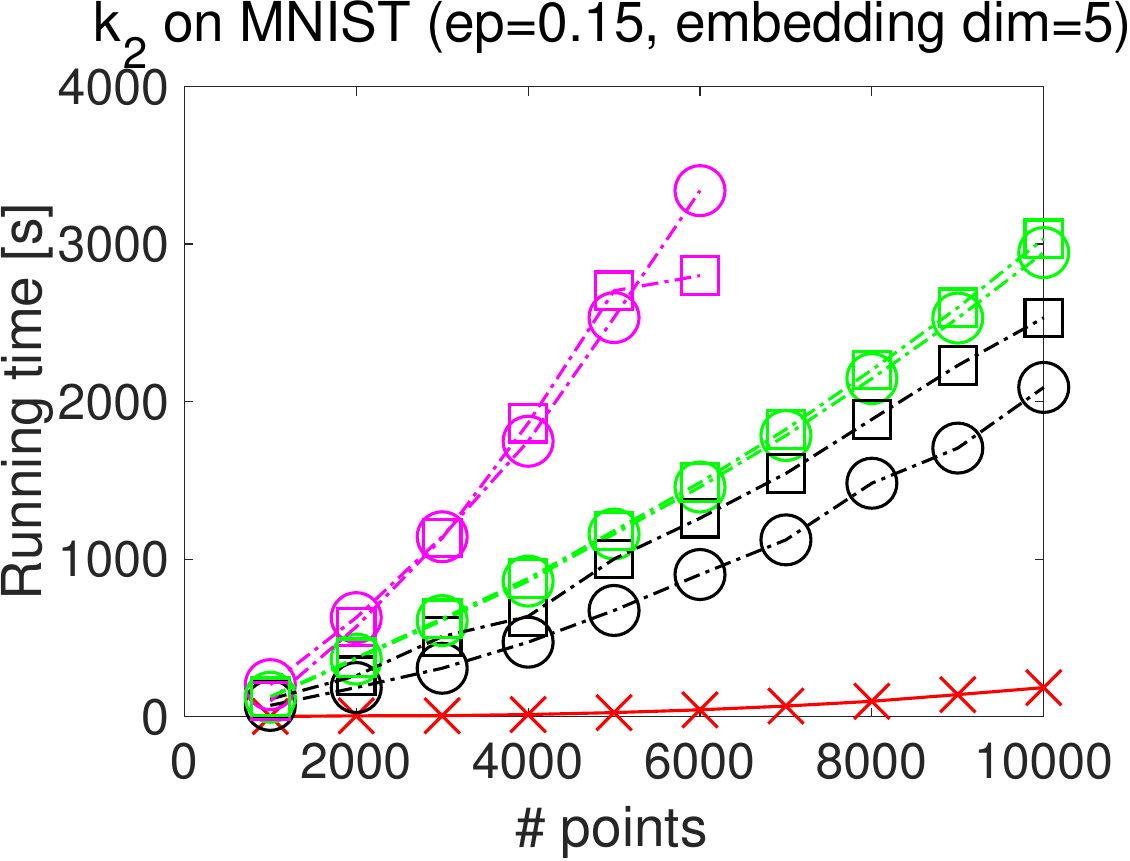}
\caption{\textbf{1st \& 2nd row (USPS digits for $\bm{k_1}$):} Clustering 1000 points from USPS digits 1, 2, and~3. 
Purity and running time  
as a function of the number of input 
triplets. 
\textbf{3rd row (MNIST digits):} Clustering subsets of MNIST digits. Purity and running time 
as a function of the number of points.}\label{clustering_art}
\end{figure*}

\paragraph{USPS digits}
We chose 1000 points uniformly at random from the subset of USPS digits 1, 2, and 3.
Using 15 landmark objects, we studied the performance of our approach and the ordinal embedding approach as a function of the number of input 
triplets. The first and the second row of Figure~\ref{clustering_art} show the results (average over 10 runs of an experiment)
 for   $k_1$. The  results for $k_2$ are shown in Figure~\ref{clustering_art2} in Section \ref{supp_figures} in the supplementary material. The first two plots of a row show the purity values of the various clusterings for $ep=0$ and $ep=0.3$, respectively. 
 The third and the fourth~plot show the corresponding time 
 (in sec) 
 that it took to compute 
 our kernel function 
 or an ordinal embedding. 
 We set the embedding dimension to 2 (1st row) or 10 (2nd row).  
 Based on the achieved purity values no method can be considered 
superior. Our kernel function $k_2$ performs slightly worse than~$k_1$ and the ordinal embedding algorithms. The GNMDS algorithm apparently cannot deal 
 with the landmark triplets at all and yields the same purity values as a random clustering when provided with the landmark triplets. 
 Our approach is highly superior regarding running time. 
 The running times of the ordinal embedding algorithms depend on the embedding dimension and $ep$ and in these experiments 
the dependence is monotonic.   
All computations were performed in Matlab R2016a on
a MacBook Pro with 2.9 GHz Intel Core i7 and 8 GB 1600 MHz DDR3. 
In order to make a fair comparison we did not use MEX files or sparse matrix operations in the implementation of our kernel~functions.

\paragraph{MNIST digits}
We studied the performance of the various methods as a function of the size $n$ of the data set 
with 
the number of input triplets 
growing 
linearly with $n$. For $i=1,\ldots,10$, we chose $n=i\cdot 10^3$ points uniformly at random from MNIST digits. We used 30 
 landmark objects and provided $150 n$ input similarity triplets. 
The third row of Figure~\ref{clustering_art} shows the purity 
values 
of the various methods for $k_1$ / $k_2$ (1st / 2nd plot) and the corresponding running times  (3rd / 4th  plot) 
when 
$ep=0.15$. The embedding dimension was set to 5. 
A spot check suggested  that setting it to 2 would have given  worse results, while setting it to 10 would have given 
similar
results, but would have led to a higher running time. 
We computed the t-STE embedding only for $n\leq 6000$ due to its high running time. It seems that GNMDS with random input triplets performs best, but for large values of $n$ our kernel function $k_1$ can compete with it.
For 10000 points, computing 
$k_1$ or $k_2$ took 100 or 180 seconds, while even the fastest embedding algorithm 
ran for 2000 seconds. 
For further comparison, Figure~\ref{clustering_art3} in Section \ref{supp_figures} in the supplementary material  shows  
a kernel PCA embedding based on $k_1$ (150$n$ landmark triplets) and a 2-dim GNMDS embedding (150$n$ random triplets) of $n=20000$ digits.
Here, computation of $k_1$ took 900 seconds, while GNMDS ran for more than 6000 seconds.

\section{Conclusion}
We proposed
two 
data-dependent 
kernel functions that can be evaluated when given only an arbitrary collection of similarity triplets for a data set~$\dataset$. Our kernel functions can be used to apply any kernel method to $\dataset$. Hence  they provide a generic alternative to the standard  ordinal embedding approach based on numerical optimization for machine learning with similarity triplets. 
In a number of experiments we demonstrated the meaningfulness of our kernel functions. A big advantage of our kernel functions compared to the ordinal embedding approach is that our kernel functions run significantly faster. A drawback is that, in general,  they seem to require a higher number of similarity triplets for capturing the structure of a data set. However,  in a landmark design our kernel functions can compete with the ordinal embedding approach in terms of the required number of  
triplets.

\subsubsection*{Acknowledgements} 
This work has been supported by the Institutional Strategy of the University of Tübingen (DFG, ZUK~63).

\bibliography{mybibfile.bib}

\begin{thebibliography}{31}
\providecommand{\natexlab}[1]{#1}
\providecommand{\url}[1]{\texttt{#1}}
\expandafter\ifx\csname urlstyle\endcsname\relax
  \providecommand{\doi}[1]{doi: #1}\else
  \providecommand{\doi}{doi: \begingroup \urlstyle{rm}\Url}\fi

\bibitem[Agarwal et~al.(2007)Agarwal, Wills, Cayton, Lanckriet, Kriegman, and
  Belongie]{AgarwalEtal07}
S.~Agarwal, J.~Wills, L.~Cayton, G.~Lanckriet, D.~Kriegman, and S.~Belongie.
\newblock Generalized non-metric multidimensional scaling.
\newblock In \emph{International Conference on Artificial Intelligence and
  Statistics (AISTATS)}, 2007.

\bibitem[Amid and Ukkonen(2015)]{Ukkonen_multiview}
E.~Amid and A.~Ukkonen.
\newblock Multiview triplet embedding: Learning attributes in multiple maps.
\newblock In \emph{International Conference on Machine Learning (ICML)}, 2015.

\bibitem[Amid et~al.(2016)Amid, Vlassis, and Warmuth]{amid2017}
E.~Amid, N.~Vlassis, and M.~Warmuth.
\newblock $t$-exponential triplet embedding.
\newblock arXiv:1611.09957v1 [cs.AI], 2016.

\bibitem[de~Silva and Tenenbaum(2004)]{landmark_mds}
V.~de~Silva and J.~Tenenbaum.
\newblock Sparse multidimensional scaling using landmark points.
\newblock Technical report, Stanford University, 2004.

\bibitem[Dhillon et~al.(2001)Dhillon, Guan, and Kulis]{kernel_k_means}
I.~Dhillon, Y.~Guan, and B.~Kulis.
\newblock Kernel k-means, spectral clustering and normalized cuts.
\newblock In \emph{International Conference on Knowledge Discovery and Data
  Mining (KDD)}, 2001.

\bibitem[Greene and Cunningham(2006)]{Greene_diagdom}
D.~Greene and P.~Cunningham.
\newblock Practical solutions to the problem of diagonal dominance in kernel
  document clustering.
\newblock In \emph{International Conference on Machine Learning (ICML)}, 2006.

\bibitem[Gustavson(1978)]{gustavson1978}
F.~G. Gustavson.
\newblock Two fast algorithms for sparse matrices: Multiplication and permuted
  transposition.
\newblock \emph{ACM Transactions on Mathematical Software}, 4\penalty0
  (3):\penalty0 250--269, 1978.

\bibitem[Haghiri et~al.(2017)Haghiri, von Luxburg, and
  Ghoshdastidar]{siavash_2017}
S.~Haghiri, U.~von Luxburg, and D.~Ghoshdastidar.
\newblock Comparison based nearest neighbor search.
\newblock In \emph{International Conference on Artificial Intelligence and
  Statistics (AISTATS)}, 2017.

\bibitem[Heikinheimo and Ukkonen(2013)]{crowdmedian}
H.~Heikinheimo and A.~Ukkonen.
\newblock The crowd-median algorithm.
\newblock In \emph{Conference on Human Computation and Crowdsourcing (HCOMP)},
  2013.
\newblock Data available on
  \href{http://www.anttiukkonen.com/}{http://www.anttiukkonen.com/}.

\bibitem[Heim et~al.(2015)Heim, Berger, Seversky, and Hauskrecht]{Heim2015}
E.~Heim, M.~Berger, L.~M. Seversky, and M.~Hauskrecht.
\newblock Efficient online relative comparison kernel learning.
\newblock In \emph{SIAM International Conference on Data Mining (SDM)}, 2015.

\bibitem[Higham(1990)]{fast_matrix_multiplication}
N.~Higham.
\newblock Exploiting fast matrix multiplication within the level 3 {BLAS}.
\newblock \emph{ACM Transactions on Mathematical Software}, 16\penalty0
  (4):\penalty0 352--368, 1990.

\bibitem[Jain et~al.(2016)Jain, Jamieson, and Nowak]{jamieson_finite}
L.~Jain, K.~Jamieson, and R.~Nowak.
\newblock Finite sample prediction and recovery bounds for ordinal embedding.
\newblock In \emph{Neural Information Processing Systems (NIPS)}, 2016.

\bibitem[Jamieson and Nowak(2011)]{JamNow11}
K.~Jamieson and R.~Nowak.
\newblock Low-dimensional embedding using adaptively selected ordinal data.
\newblock In \emph{Allerton Conference on Communication, Control, and
  Computing}, 2011.

\bibitem[Jiao and Vert(2015)]{jiao_kendall}
Y.~Jiao and J.-P. Vert.
\newblock The {Kendall} and {Mallows} kernels for permutations.
\newblock In \emph{International Conference on Machine Learning (ICML)}, 2015.

\bibitem[Kaplan et~al.(2006)Kaplan, Sharir, and Verbin]{kaplan2006}
H.~Kaplan, M.~Sharir, and E.~Verbin.
\newblock Colored intersection searching via sparse rectangular matrix
  multiplication.
\newblock In \emph{Symposium on Computational Geometry (SoCG)}, 2006.

\bibitem[Kendall(1938)]{kendalltau}
M.~Kendall.
\newblock A new measure of rank correlation.
\newblock \emph{Biometrika}, 30\penalty0 (1--2):\penalty0 81--93, 1938.

\bibitem[Kleindessner and von Luxburg(2014)]{kleindessner14}
M.~Kleindessner and U.~von Luxburg.
\newblock Uniqueness of ordinal embedding.
\newblock In \emph{Conference on Learning Theory (COLT)}, 2014.

\bibitem[Kleindessner and von Luxburg(2017)]{kleindessner16}
M.~Kleindessner and U.~von Luxburg.
\newblock Lens depth function and $k$-relative neighborhood graph: Versatile
  tools for ordinal data analysis.
\newblock \emph{JMLR}, 18\penalty0 (58):\penalty0 1--52, 2017.
\newblock Data available on
  \href{http://www.tml.cs.uni-tuebingen.de/team/luxburg/code\_and\_data/}{http://www.tml.cs.uni-tuebingen.de/team/luxburg/code\_and\_data/}.

\bibitem[Manning et~al.(2008)Manning, Raghavan, and
  Sch{\"u}tze]{manning_intro_inf_ret}
C.~D. Manning, P.~Raghavan, and H.~Sch{\"u}tze.
\newblock \emph{Introduction to Information Retrieval}.
\newblock Cambridge University Press, 2008.

\bibitem[McFee and Lanckriet(2011)]{McfLan11}
B.~McFee and G.~Lanckriet.
\newblock Learning multi-modal similarity.
\newblock \emph{JMLR}, 12:\penalty0 491--523, 2011.

\bibitem[Sch{\"o}lkopf et~al.(1999)Sch{\"o}lkopf, Smola, and
  M{\"u}ller]{kernel_pca}
B.~Sch{\"o}lkopf, A.~Smola, and K.-R. M{\"u}ller.
\newblock Kernel principal component analysis.
\newblock In B.~Sch{\"o}lkopf, C.~Burges, and A.~Smola, editors, \emph{Advances
  in Kernel Methods: Support Vector Learning}, pages 327--352. MIT Press, 1999.

\bibitem[Sch{\"o}lkopf et~al.(2002)Sch{\"o}lkopf, Weston, Eskin, Leslie, and
  Noble]{schoelk_diagdom}
B.~Sch{\"o}lkopf, J.~Weston, E.~Eskin, C.~Leslie, and W.~Noble.
\newblock A kernel approach for learning from almost orthogonal patterns.
\newblock In \emph{European Conference on Machine Learning (ECML)}, 2002.

\bibitem[Schultz and Joachims(2003)]{SchulzJoachims03}
M.~Schultz and T.~Joachims.
\newblock Learning a distance metric from relative comparisons.
\newblock In \emph{Neural Information Processing Systems (NIPS)}, 2003.

\bibitem[Stewart et~al.(2005)Stewart, Brown, and Chater]{stewart2005}
N.~Stewart, G.~D.~A. Brown, and N.~Chater.
\newblock Absolute identification by relative judgment.
\newblock \emph{Psychological Review}, 112\penalty0 (4):\penalty0 881--911,
  2005.

\bibitem[Tamuz et~al.(2011)Tamuz, Liu, Belongie, Shamir, and
  Kalai]{TamuzEtal2011}
O.~Tamuz, C.~Liu, S.~Belongie, O.~Shamir, and A.~Kalai.
\newblock Adaptively learning the crowd kernel.
\newblock In \emph{International Conference on Machine Learning (ICML)}, 2011.

\bibitem[Terada and von Luxburg(2014)]{terada14}
Y.~Terada and U.~von Luxburg.
\newblock Local ordinal embedding.
\newblock In \emph{International Conference on Machine Learning (ICML)}, 2014.

\bibitem[Ukkonen(2017)]{ukkonen_corrclust}
A.~Ukkonen.
\newblock Crowdsourced correlation clustering with relative distance
  comparisons.
\newblock In \emph{International Conference on Data Mining series (ICDM)},
  2017.

\bibitem[Ukkonen et~al.(2015)Ukkonen, Derakhshan, and
  Heikinheimo]{ukkonen_density}
A.~Ukkonen, B.~Derakhshan, and H.~Heikinheimo.
\newblock Crowdsourced nonparametric density estimation using relative
  distances.
\newblock In \emph{Conference on Human Computation and Crowdsourcing (HCOMP)},
  2015.

\bibitem[van~der Maaten and Weinberger(2012)]{stoch_trip_embed}
L.~van~der Maaten and K.~Weinberger.
\newblock Stochastic triplet embedding.
\newblock In \emph{IEEE International Workshop on Machine Learning for Signal
  Processing (MLSP)}, 2012.
\newblock Code available on
  \href{http://homepage.tudelft.nl/19j49/ste/}{http://homepage.tudelft.nl/19j49/ste/}.

\bibitem[Wilber et~al.(2014)Wilber, Kwak, and Belongie]{wilber2014}
M.~Wilber, I.~Kwak, and S.~Belongie.
\newblock Cost-effective hits for relative similarity comparisons.
\newblock In \emph{Conference on Human Computation and Crowdsourcing (HCOMP)},
  2014.
\newblock Data available on
  \href{http://vision.cornell.edu/se3/projects/cost-effective-hits/}{http://vision.cornell.edu/se3/projects/cost-effective-hits/}.

\bibitem[Wilber et~al.(2015)Wilber, Kwak, Kriegman, and Belongie]{Wilber2015}
M.~Wilber, I.~Kwak, D.~Kriegman, and S.~Belongie.
\newblock Learning concept embeddings with combined human-machine expertise.
\newblock In \emph{International Conference on Computer Vision (ICCV)}, 2015.

\end{thebibliography}
\bibliographystyle{plainnat}

\newpage

\normalsize
\appendix
\section{Supplementary material}

\subsection{Omitted figures}\label{supp_figures}

\vspace{5mm}

\begin{figure*}[h]
\centering
\begin{tabular}{>{\centering}p{\cellsi} >{\centering}p{\cellsi} 
>{\centering}p{\cellsi} >{\centering}p{\cellsi } >{\centering}p{2.8cm}}
\begin{minipage}[t][\cellhi][c]{\cellsi}
\begin{center}
400 points
\end{center}
\end{minipage} & 
\begin{minipage}[t][\cellhi][c]{\cellsi}
\begin{center}
Distance matrix
\end{center} 
\end{minipage} & 
\begin{minipage}[t][\cellhi][c]{\cellsi}
\begin{center}
$K_1$ 
\end{center}
\end{minipage} & 
\begin{minipage}[t][\cellhi][c]{\cellsi}
\begin{center}
$K_2$ 
\end{center}
\end{minipage} & 
\begin{minipage}[t][\cellhi][c]{2.8cm}
\begin{center}
Similarity scores
\end{center}
\end{minipage}  
\tabularnewline 
\rule{0pt}{20pt}
\hspace{-1.2mm}
\includegraphics[width=\picsiA]{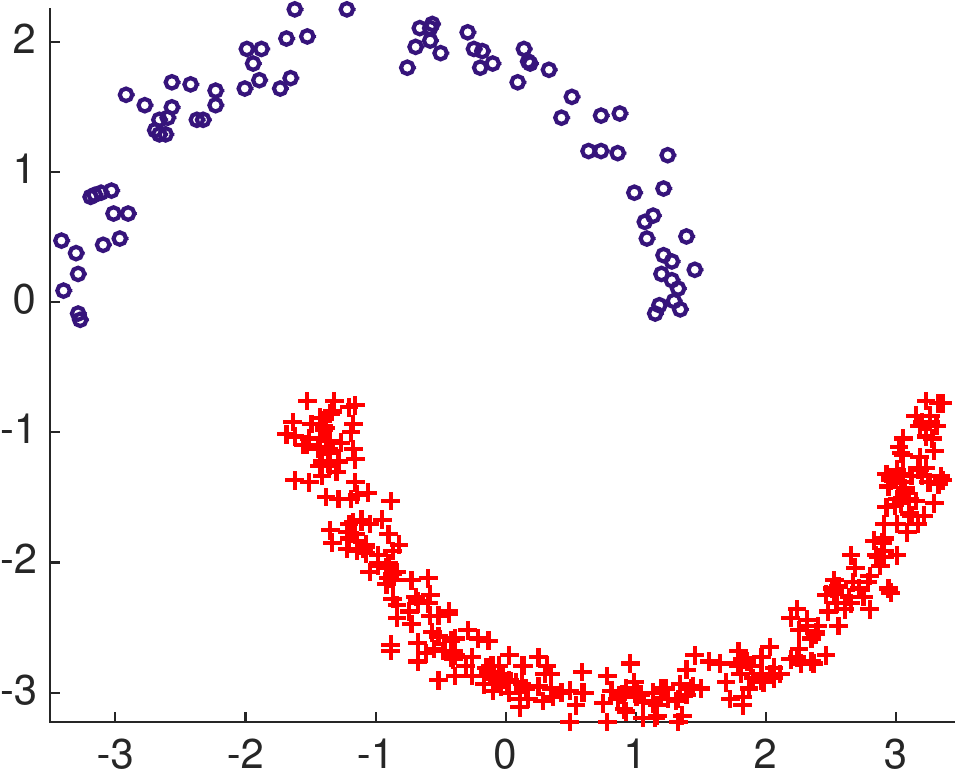} & 
\includegraphics[width=\picsiA]{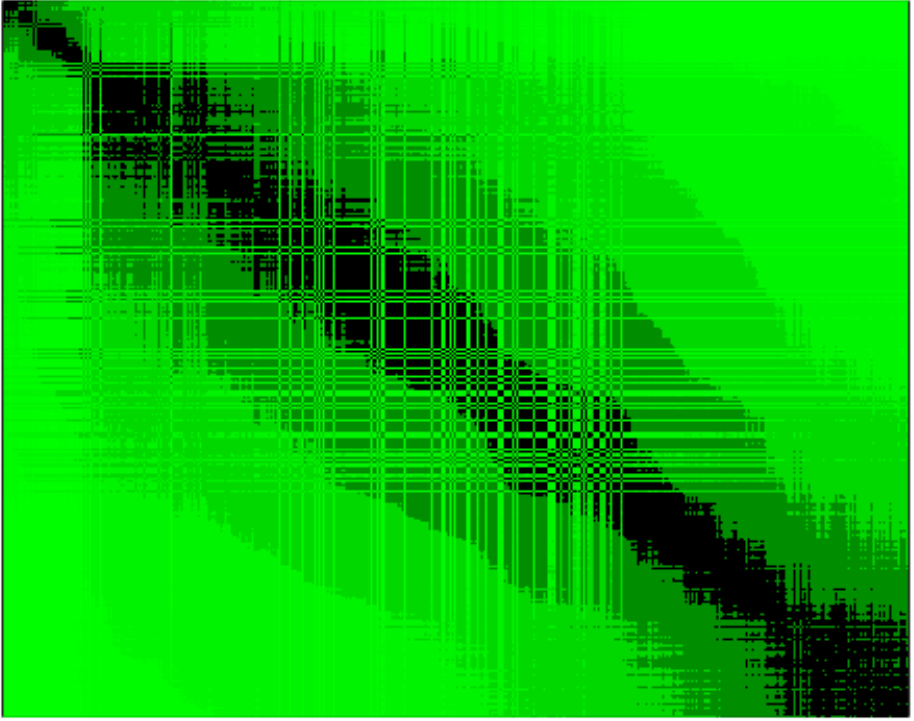} & 
\includegraphics[width=\picsiA]{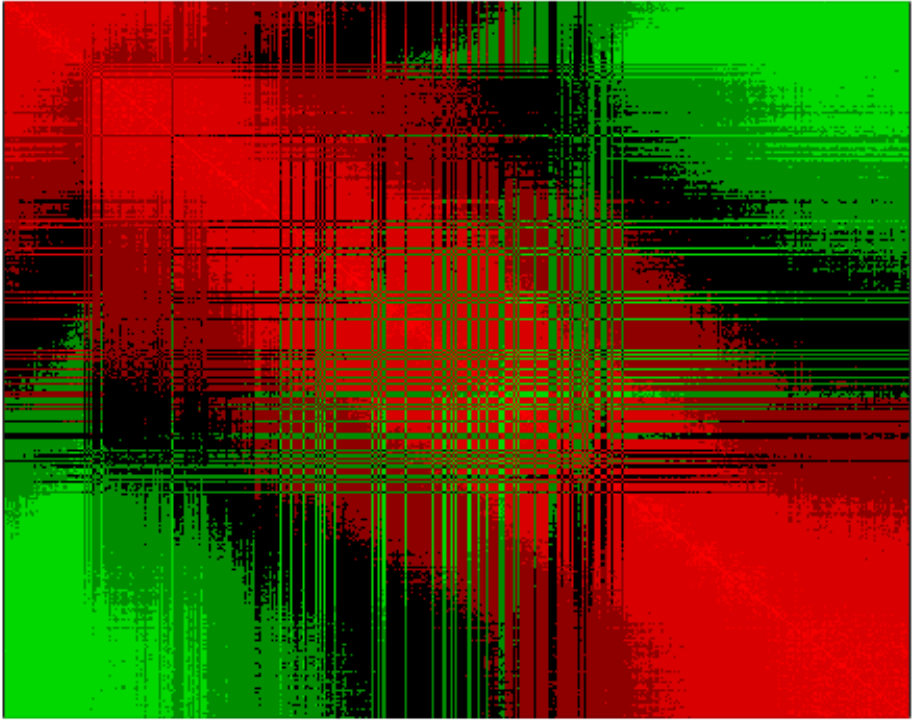} & 
\includegraphics[width=\picsiA]{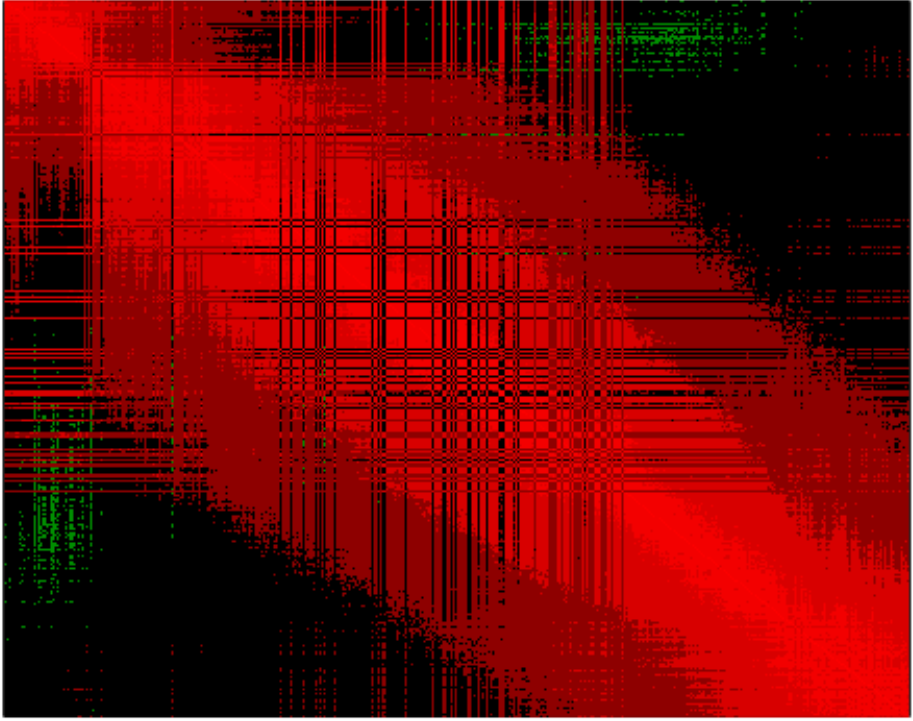} & 
\includegraphics[width=2.58cm]{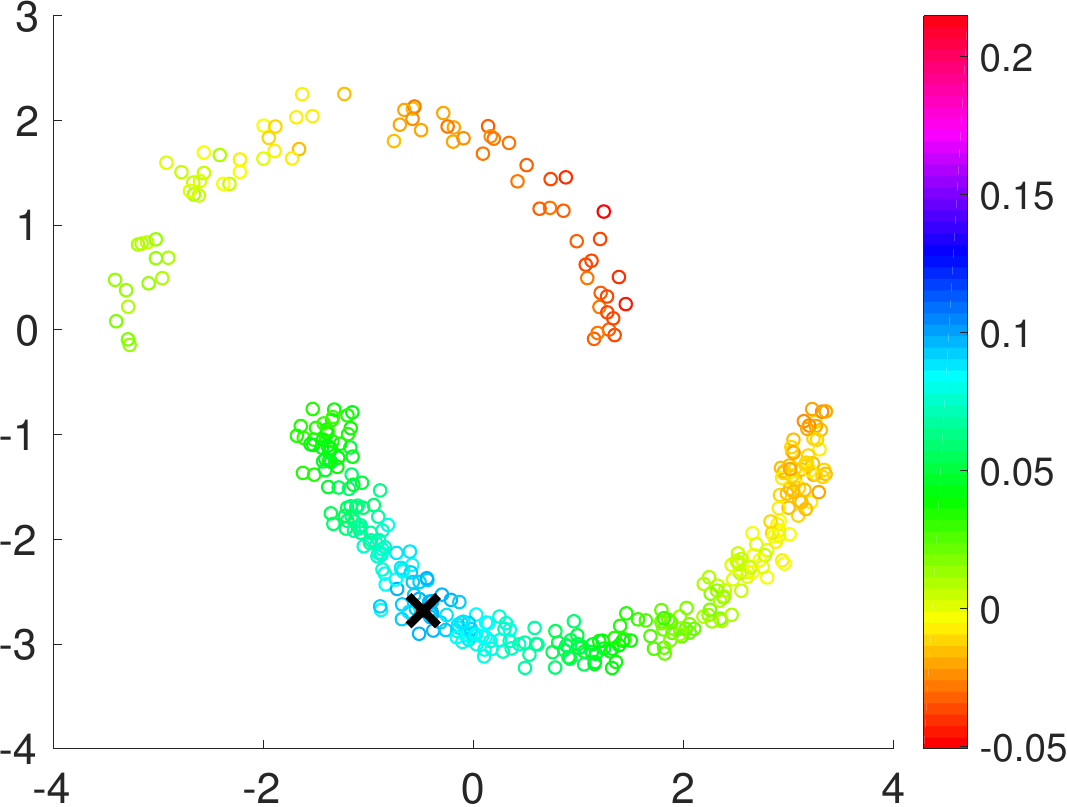}
\tabularnewline
\rule{0pt}{60pt}
\includegraphics[width=2cm]{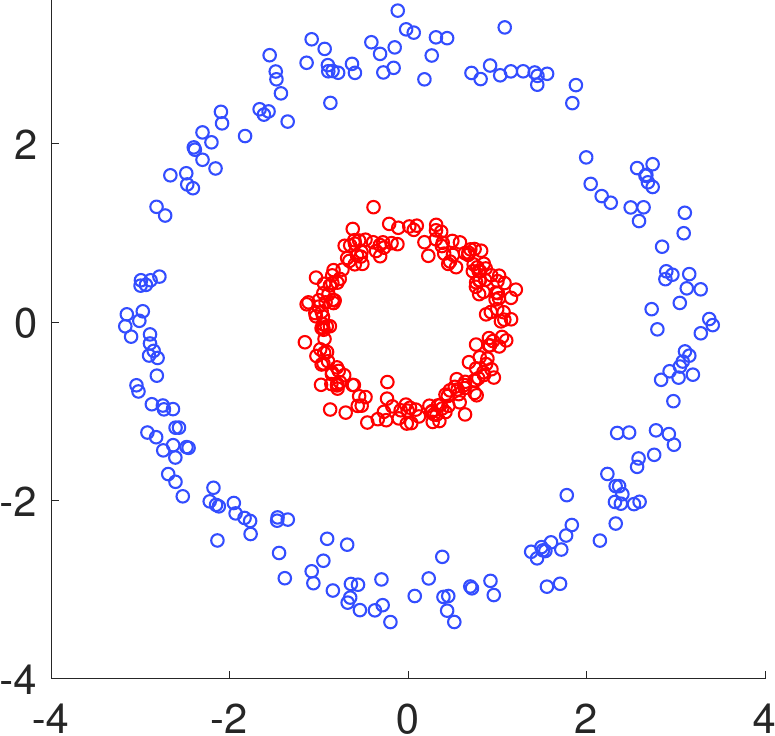} & 
\includegraphics[width=\picsiA]{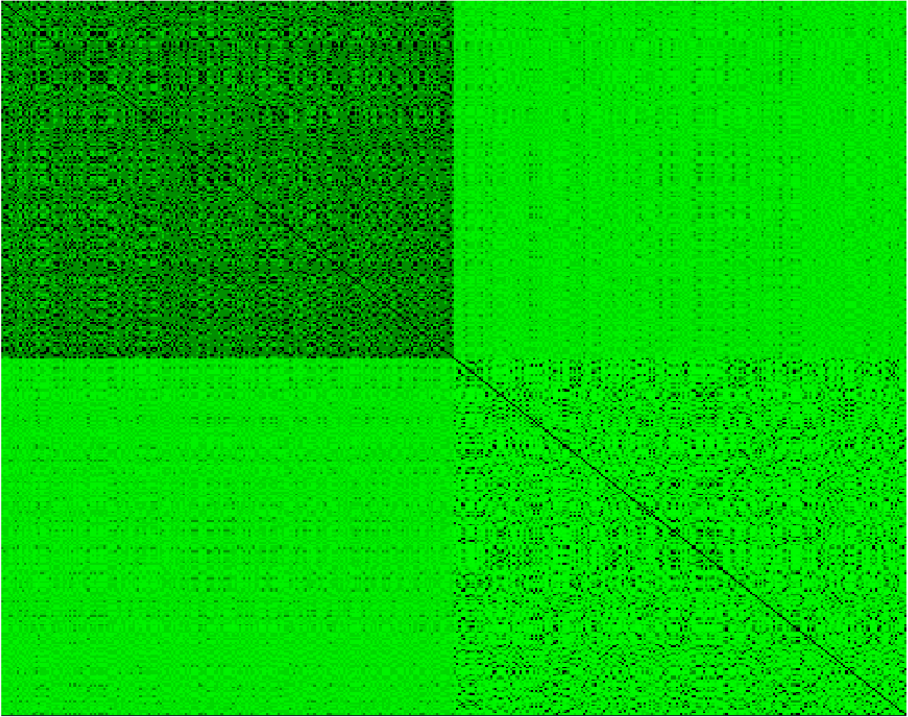} & 
\includegraphics[width=\picsiA]{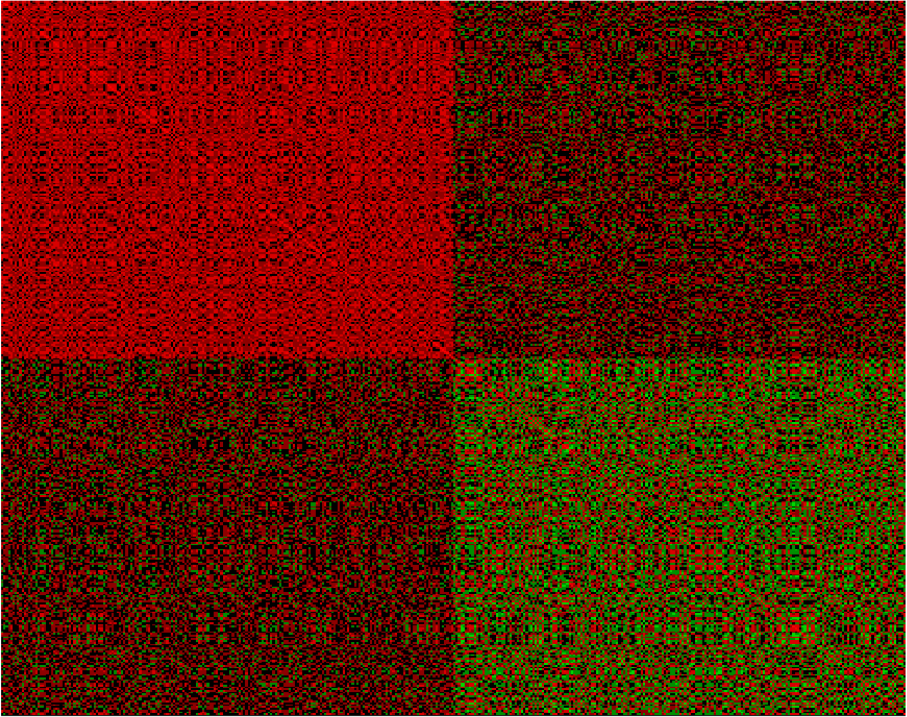} & 
\includegraphics[width=\picsiA]{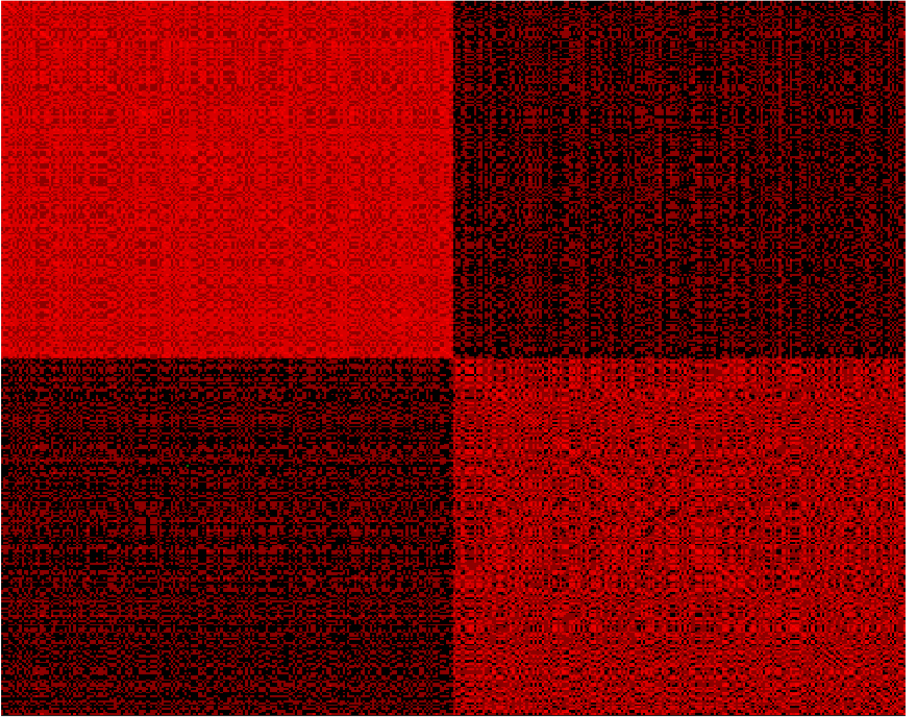} & 
\includegraphics[width=2.3cm]{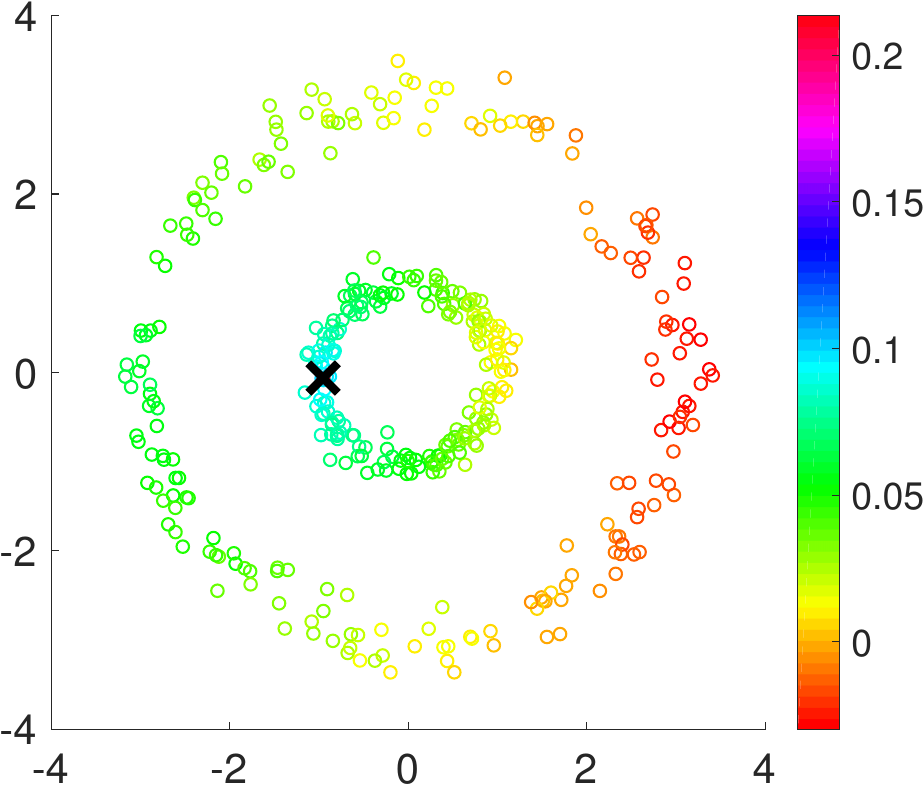}
\tabularnewline
\rule{0pt}{60pt}
\includegraphics[width=\picsiA]{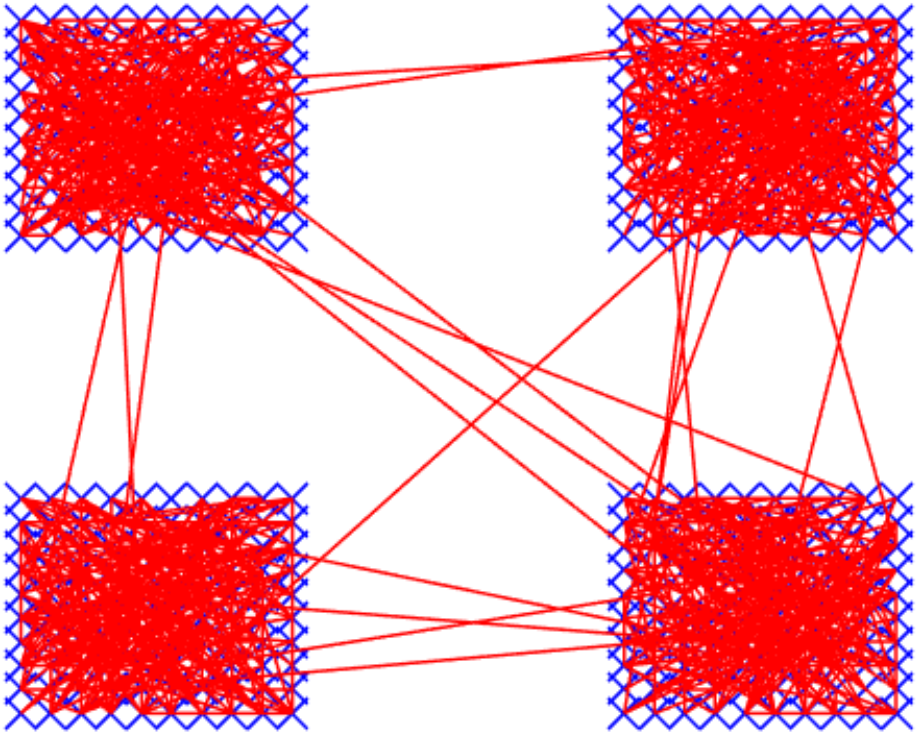} & 
\includegraphics[width=\picsiA]{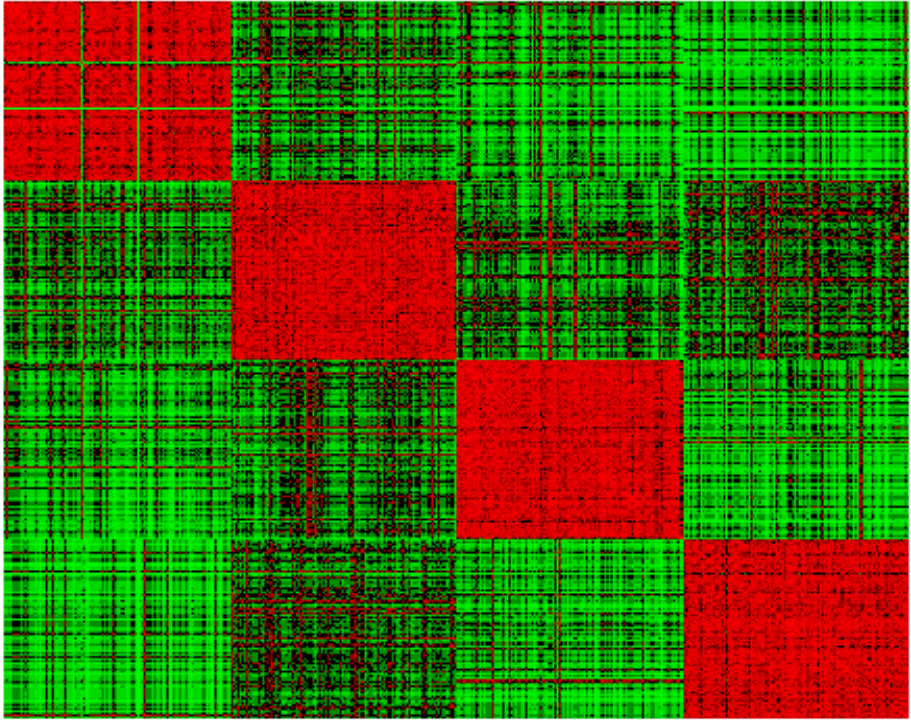} & 
\includegraphics[width=\picsiA]{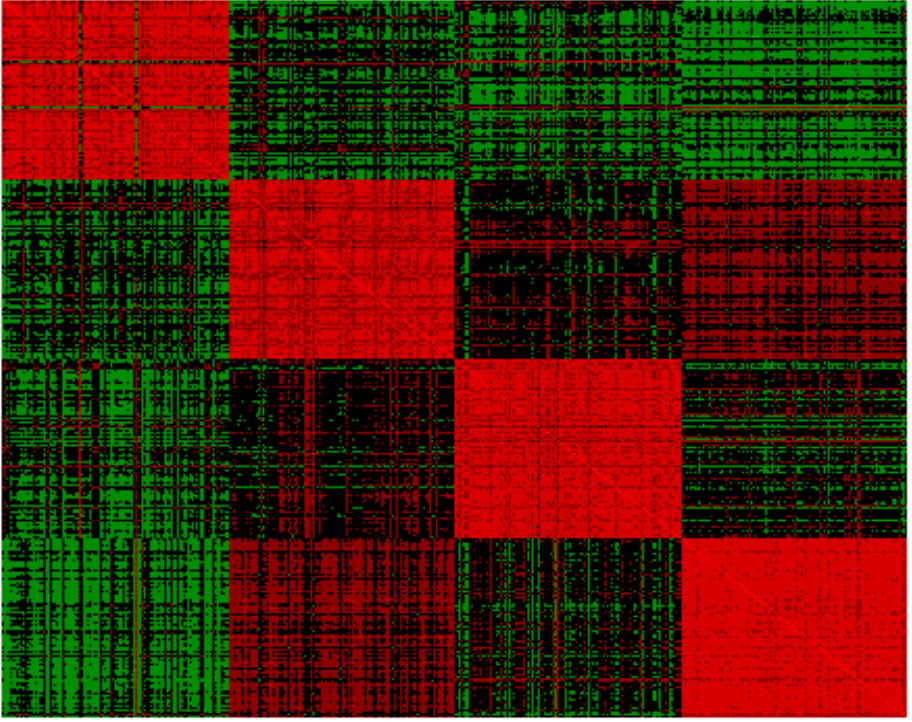} & 
\includegraphics[width=\picsiA]{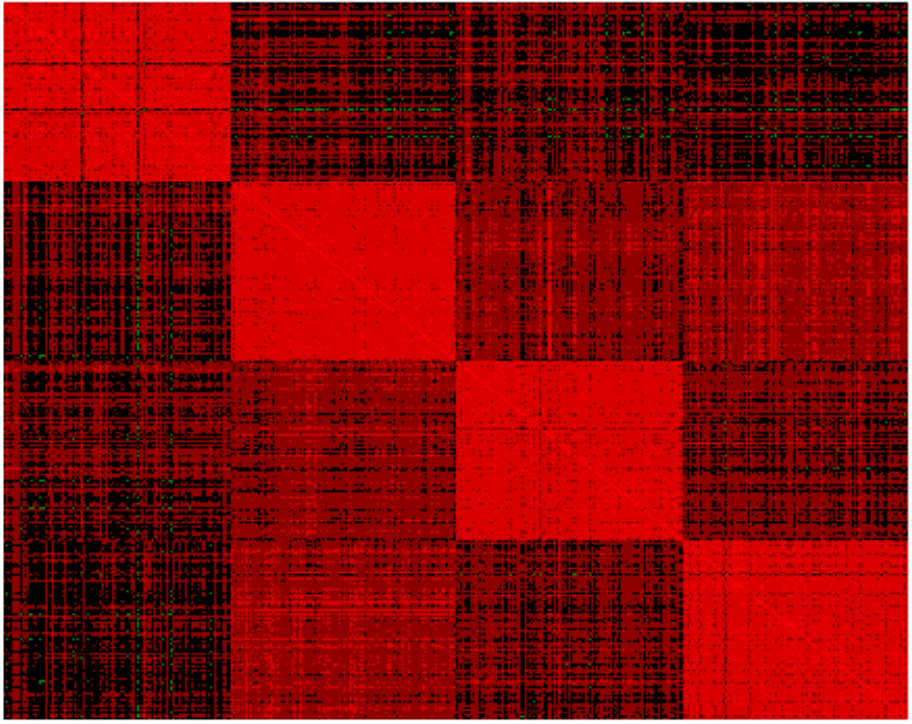} & 
\includegraphics[width=2.45cm]{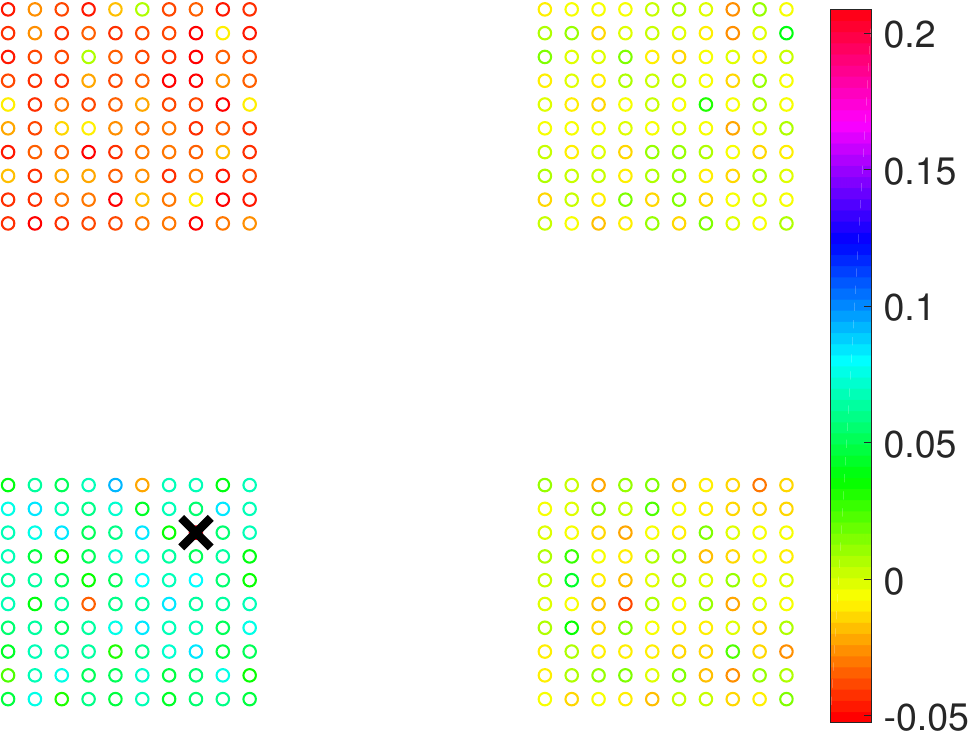}
\end{tabular}
\caption{Kernel matrices 
for 
three 
data sets, each  consisting of 400 points,  
based on 10\% of all similarity triplets (see Section \ref{sec_geom_intuition}). 
1st 
plot of a row:
Data points. 
2nd 
plot:
Distance matrix. 
3rd / 4th plot: Kernel 
matrix  
for $k_1$~/~$k_2$. 
6th 
plot:
Similarity scores 
between a fixed point 
and the other points~(for~$k_1$).}\label{fig_kernelmatrices2}
\end{figure*}

\vspace{1.2cm}

\begin{figure*}[h!]
\centering

\includegraphics[width=\siko]{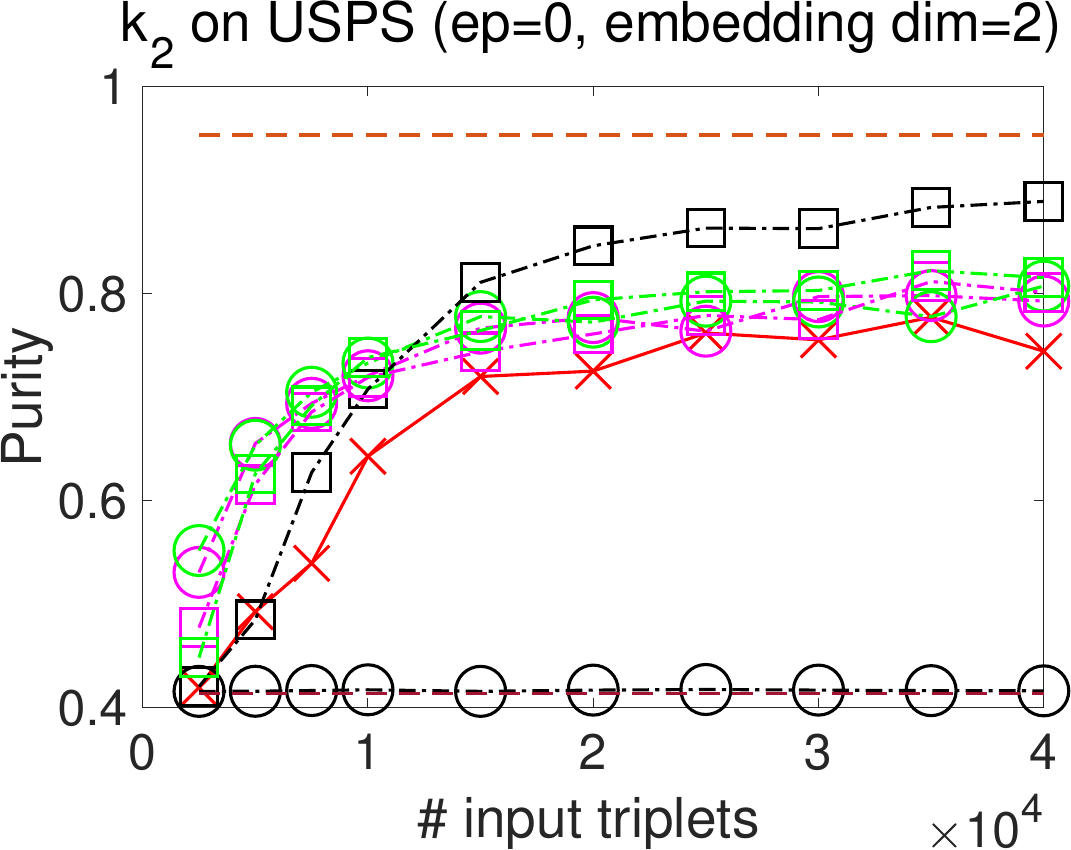} \hspace{\spo}
\includegraphics[width=\siko]{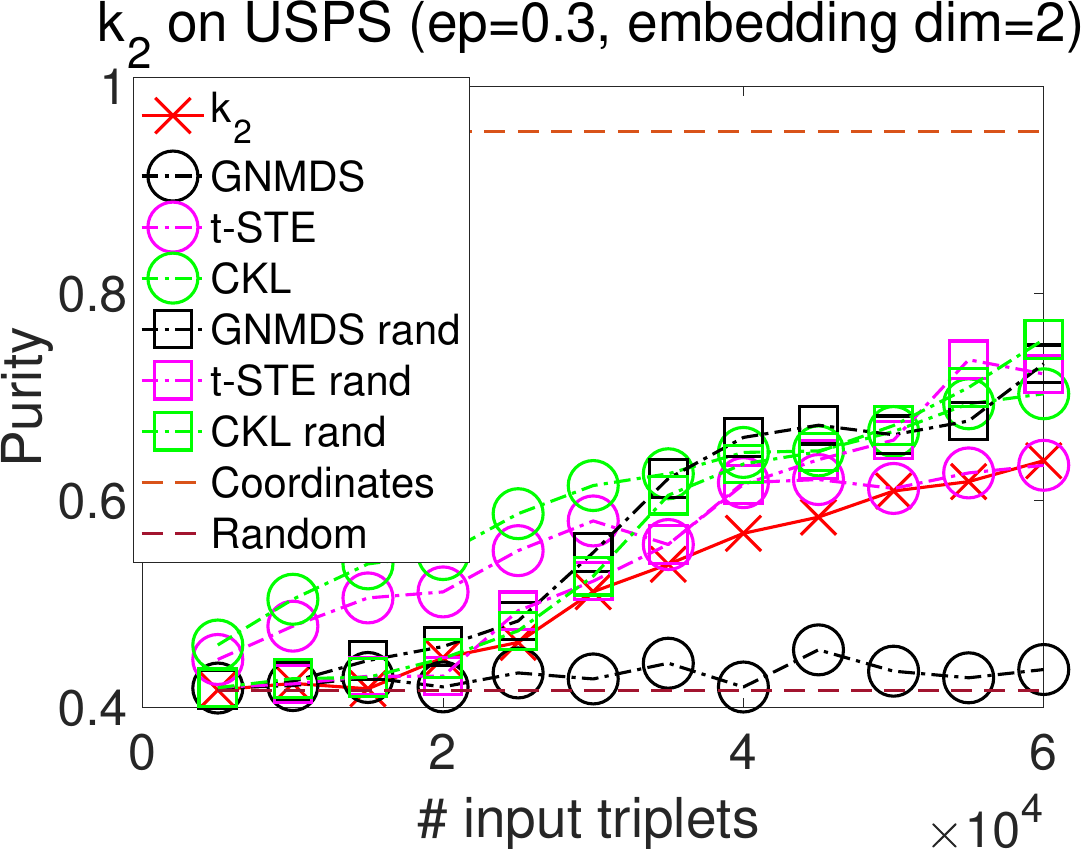} \hspace{\spo}
\includegraphics[width=\siko]{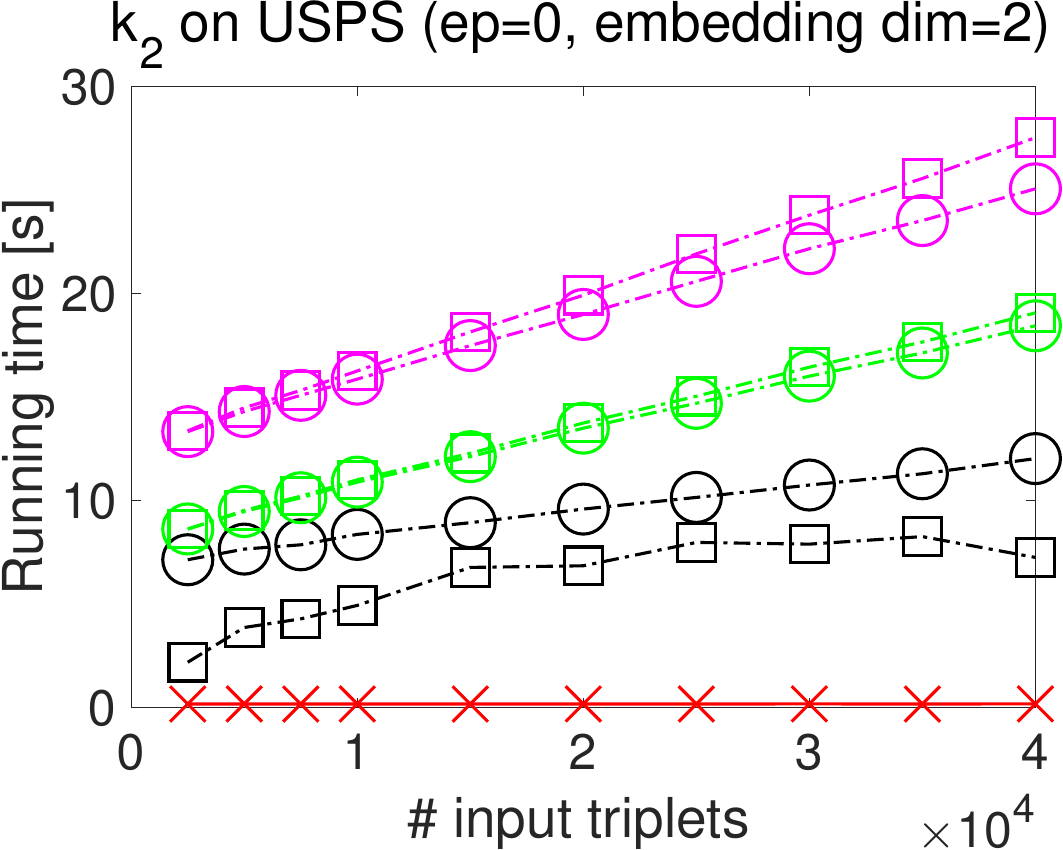}\hspace{\spo}
\includegraphics[width=\siko]{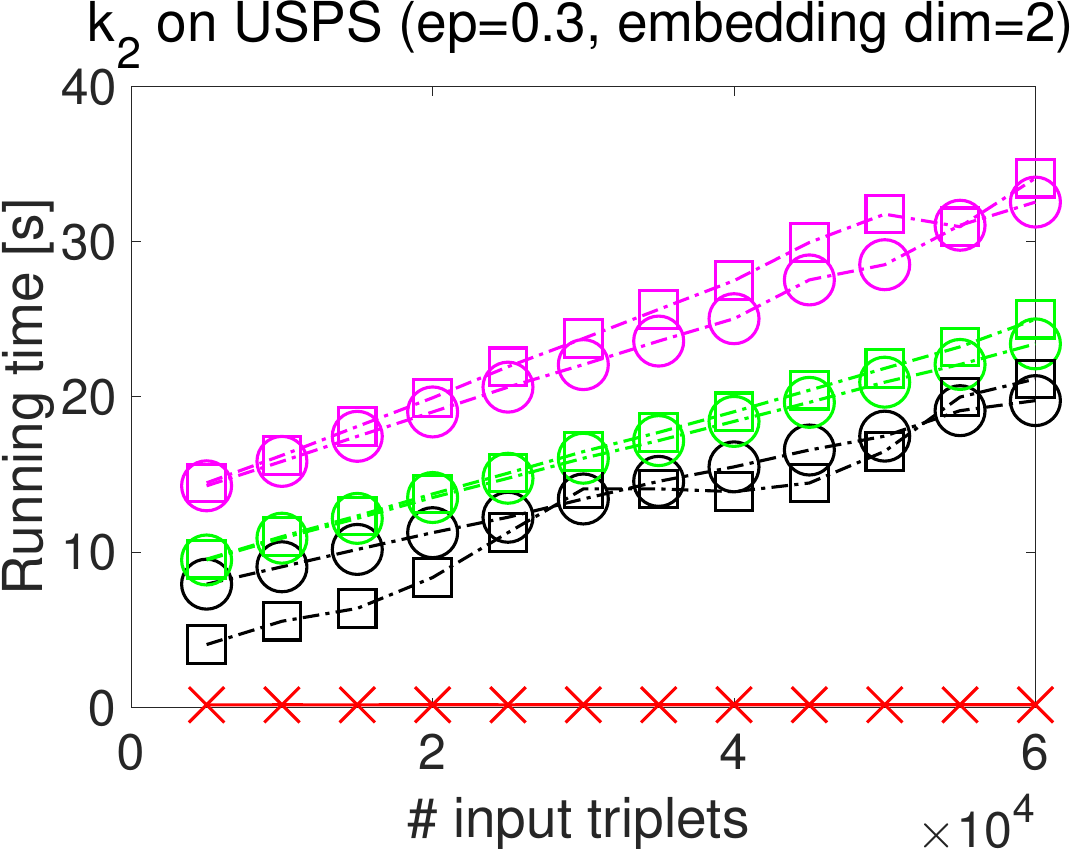}
\vspace{1mm}

\includegraphics[width=\siko]{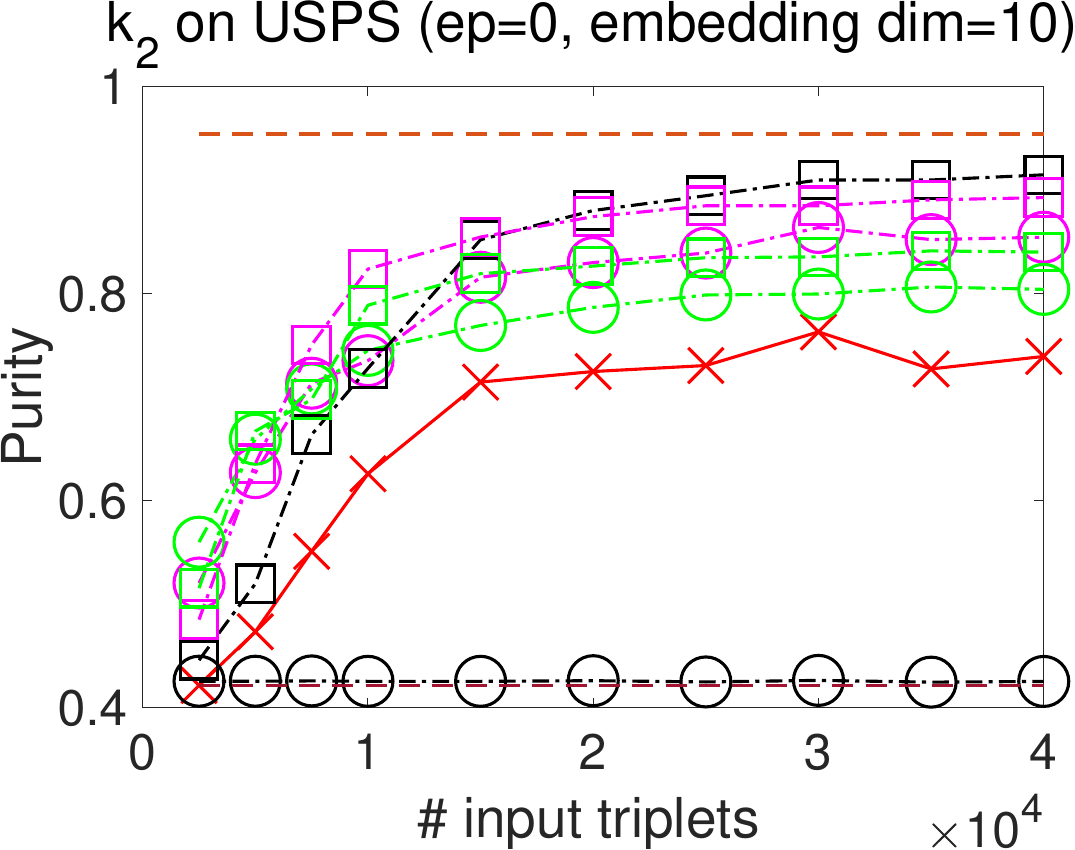} \hspace{\spo}
\includegraphics[width=\siko]{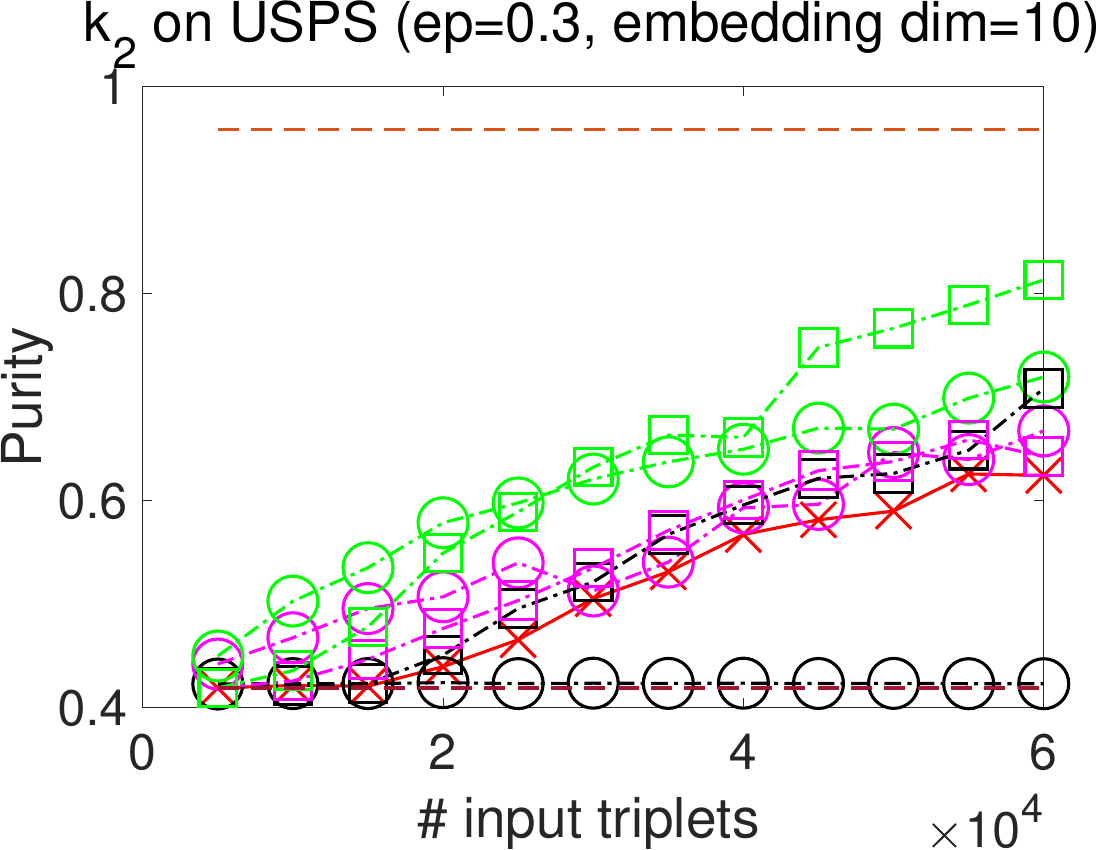} \hspace{\spo}
\includegraphics[width=\siko]{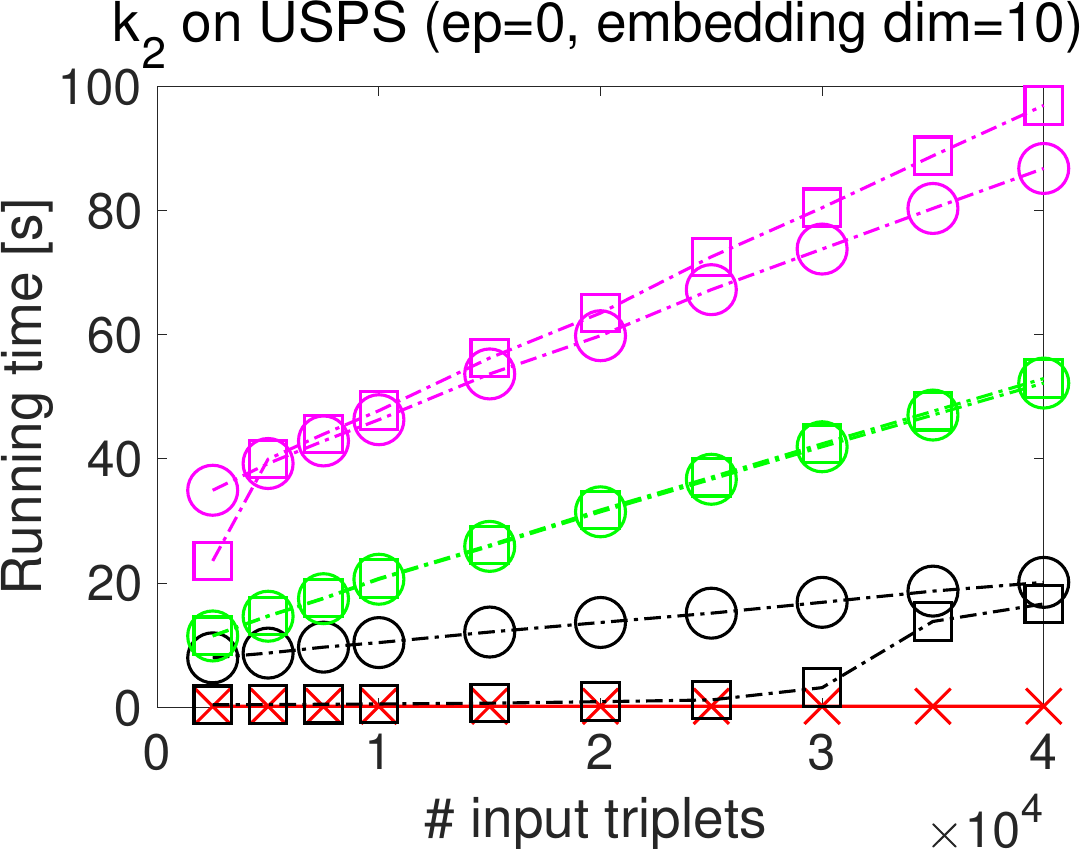}\hspace{\spo}
\includegraphics[width=\siko]{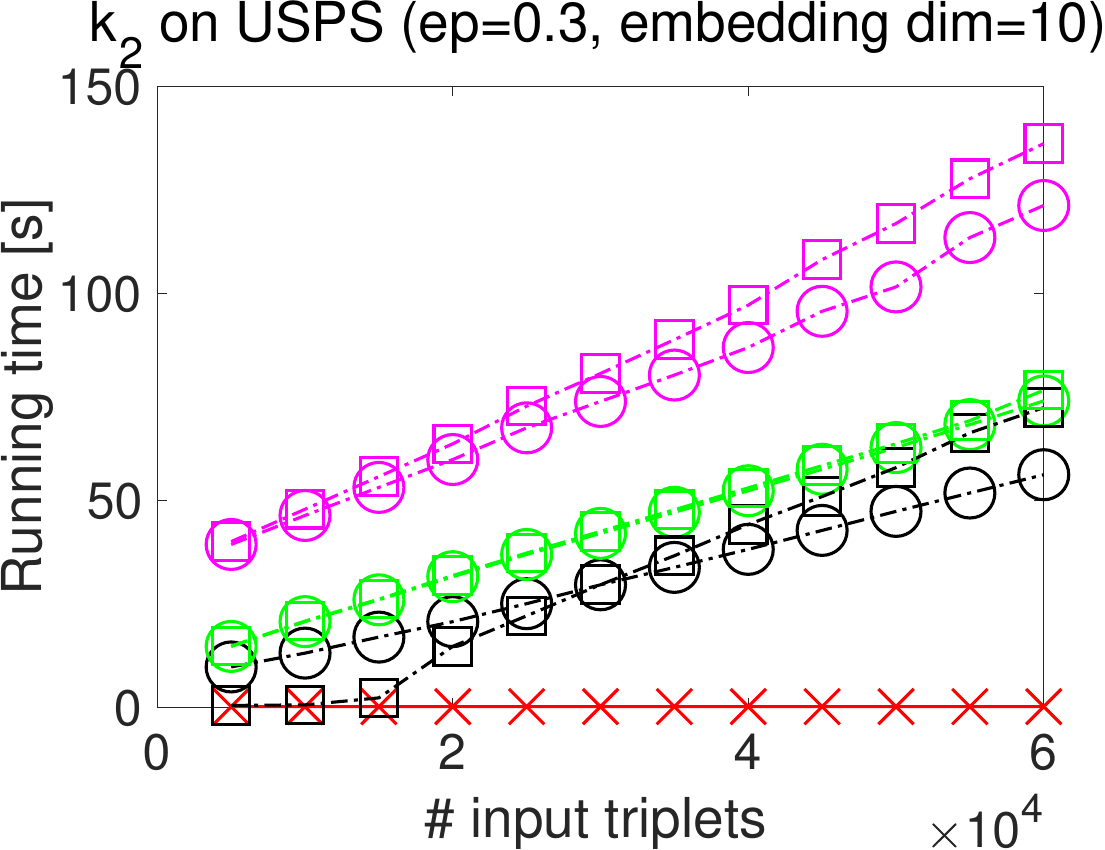}
\caption{USPS digits for $k_2$. Clustering 1000 points from USPS digits 1, 2, and~3 (see Section \ref{syn_exp}). 
Purity and running time  
as a function of the number of input 
triplets.}\label{clustering_art2}
\end{figure*}

\vspace{3cm}

\begin{figure*}[h]
\centering
\includegraphics[height=5cm]{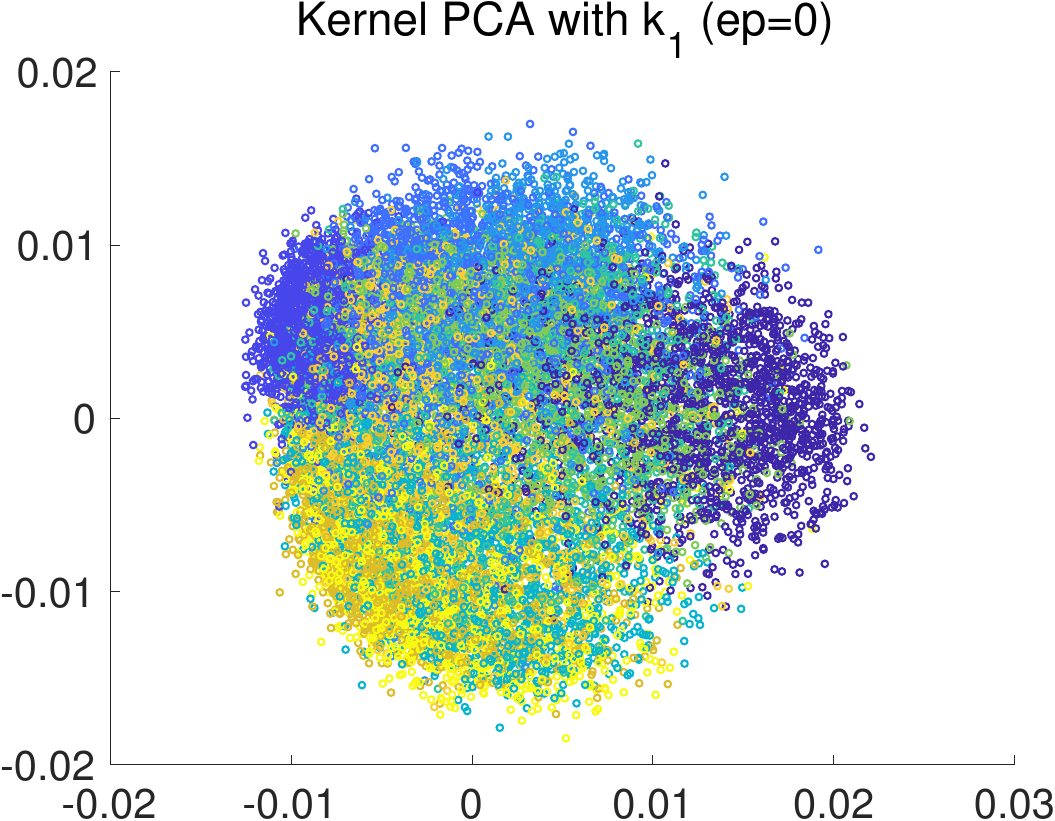}\hspace{\spo}
\hspace{1cm}
\begin{overpic}[height=5cm]{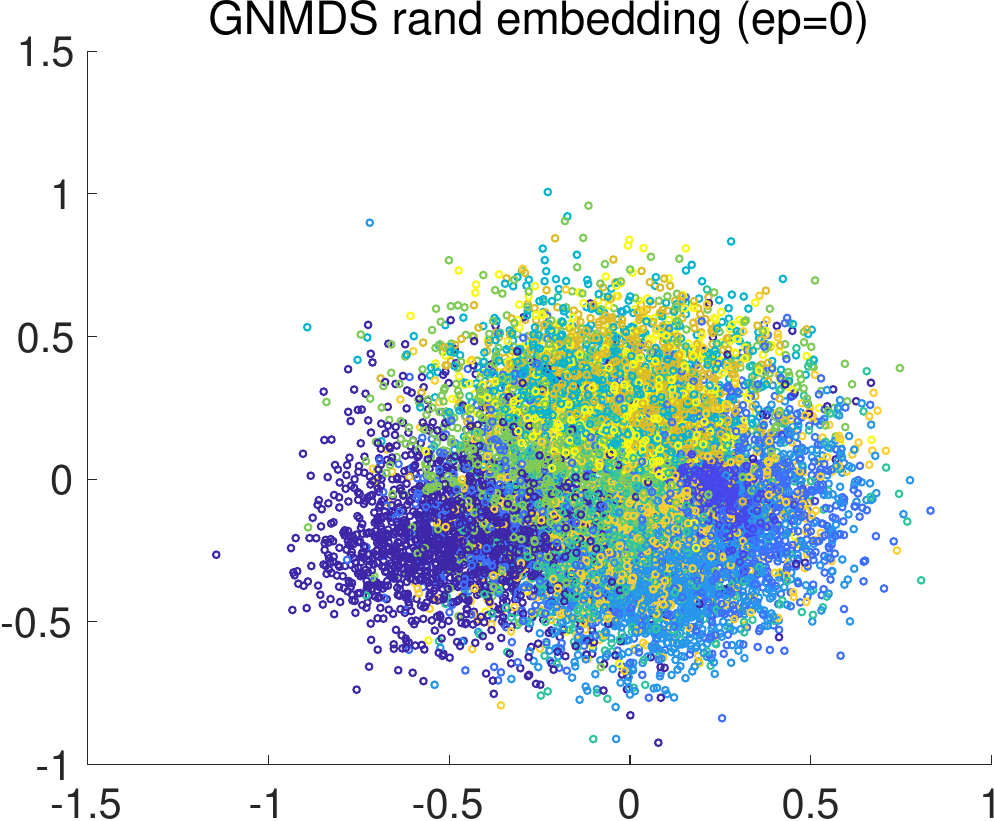}
\put(14,70){\includegraphics[width=5.2cm]{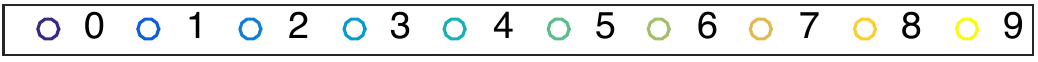}}
\end{overpic}
\caption{Embeddings of 20000 MNIST digits (see Section \ref{syn_exp}).}\label{clustering_art3}
\end{figure*}

\begin{figure*}[h!]
\vspace{1.7cm}
\centering
\includegraphics[width=0.72\textwidth]{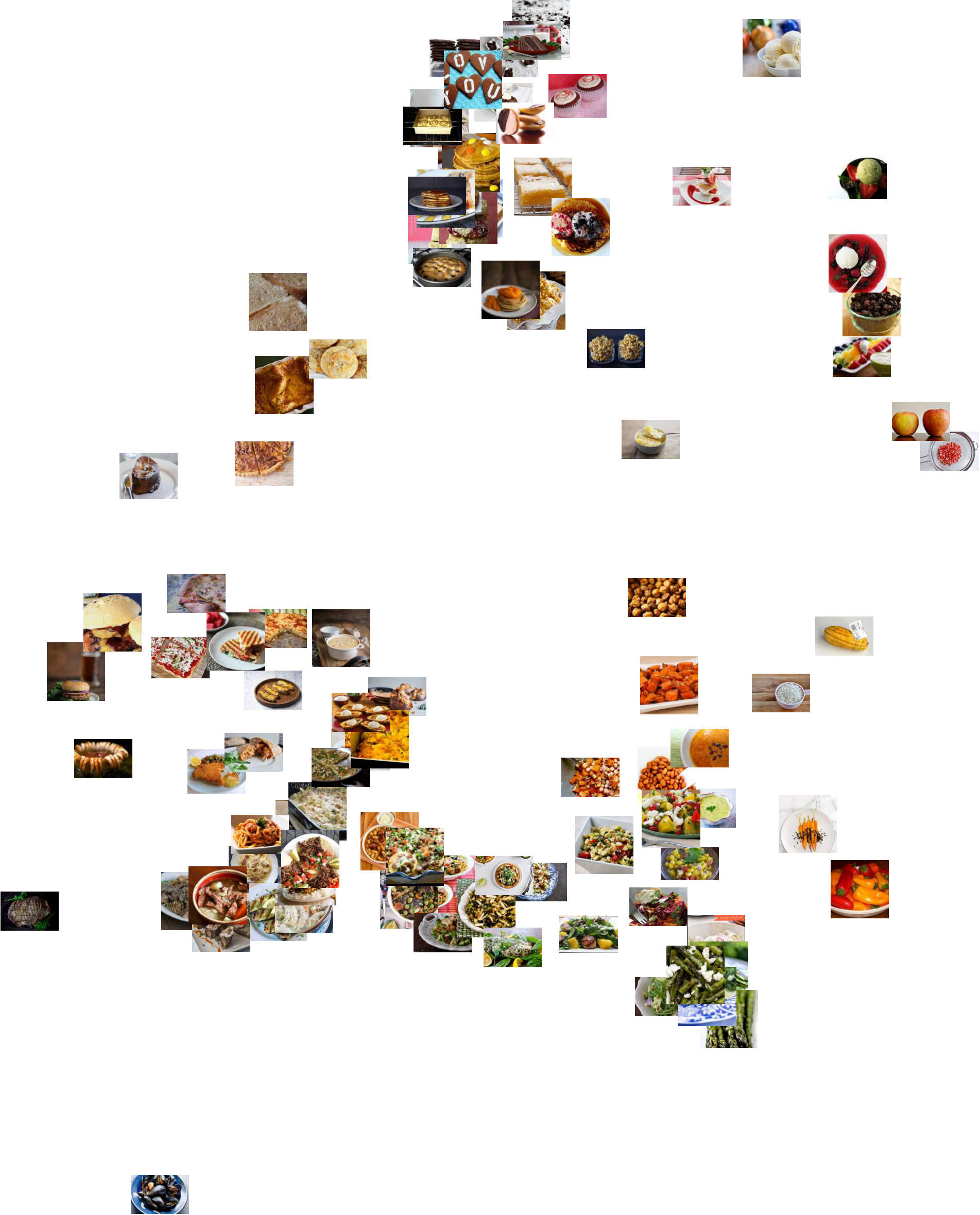}
\vspace{6mm}
\caption{An ordinal embedding of the food data set (see Section \ref{real_exp}). The embedding was computed with the GNMDS algorithm  \citep{AgarwalEtal07}.}\label{ordemb_food}
\end{figure*}

\newpage

\subsection{Experiments on the nature data set}\label{sec_supp_nature}

Similarly to the experiments on the food data set and the car data set presented in Section \ref{real_exp}, we applied 
kernel PCA 
and 
the kernelized version of complete-linkage clustering 
 based on our kernel functions to the nature data set. 
 The nature data set has been introduced in \citet{crowdmedian} and 
was also  
 used in \citet{ukkonen_corrclust}.
It consists of 120 images of landscapes, and \citet{crowdmedian} have collected statements of the kind ``Object $A$ is the outlier within the triple of objects $(A,B,C)$'' for it. Each such statement comprises two similarity triplets $d(B,C)<d(B,A)$ and $d(C,B)<d(C,A)$. 
Hence, the 3355  statements collected by \citeauthor{crowdmedian} provide 
6710 similarity triplets. These 6710 similarity triplets comprise  2994 unique triplets, and within the latter there are 636 pairs of contradicting triplets.

\begin{figure*}[t]
\centering
\includegraphics[width=0.96\textwidth]{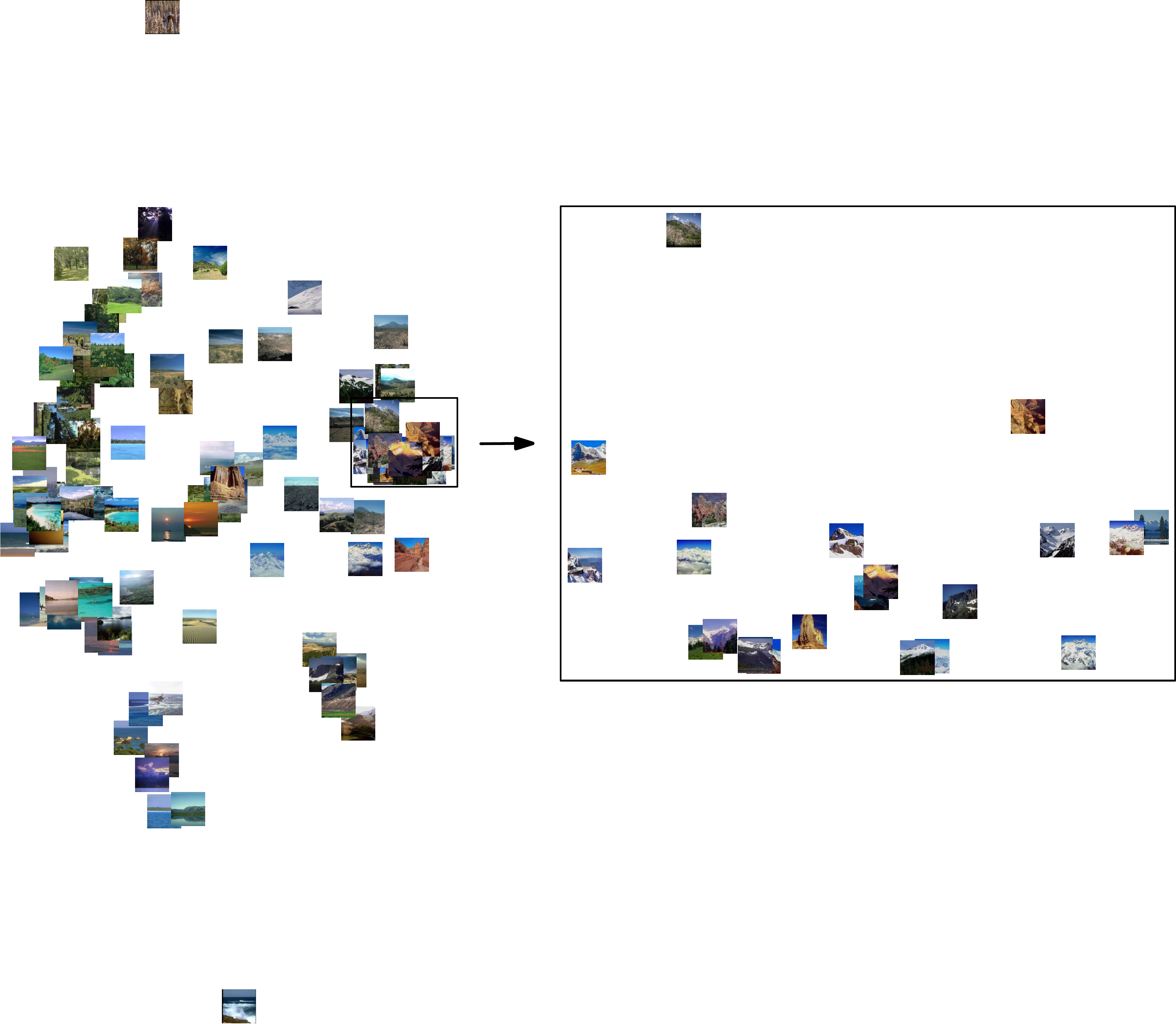}
\caption{Projection of the nature data set onto the first two kernel principal components based on~$k_1$.
}\label{kernelPCA_k1_nature}
\end{figure*}

\begin{figure*}[t]
\centering
\includegraphics[width=0.96\textwidth]{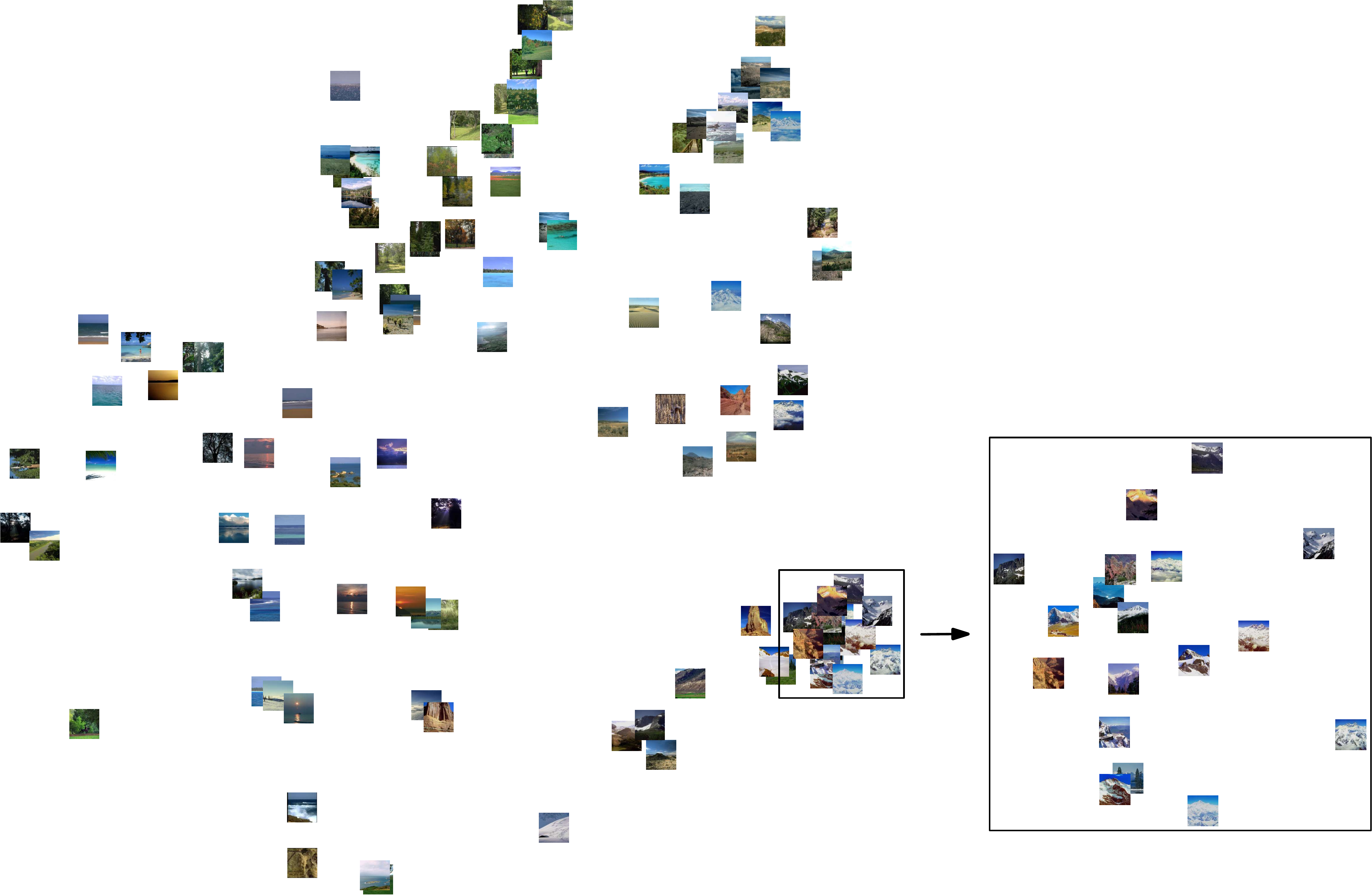}
\caption{Projection of the nature data set onto the first two kernel principal components based on~$k_2$.
}\label{kernelPCA_k2_nature}
\end{figure*}

\begin{figure*}[h!]
\centering
\includegraphics[width=0.96\textwidth]{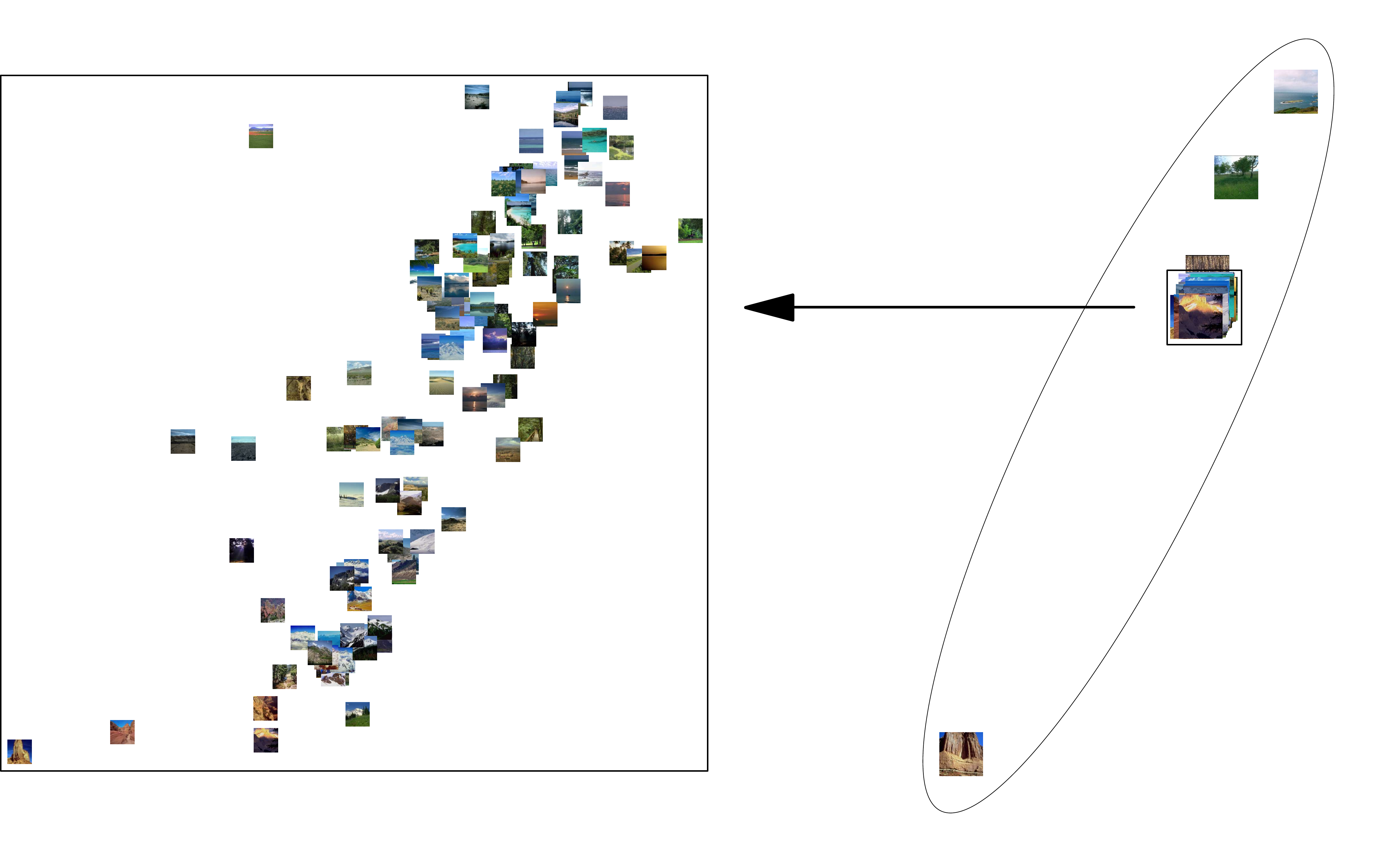}
\caption{A GNMDS embedding of the nature data set (surrounded by an ellipse) and a region that we zoomed in for more precise inspection.}\label{gnmds_emb_nature}
\end{figure*}

Figure \ref{kernelPCA_k1_nature} and Figure \ref{kernelPCA_k2_nature} show the projections of the nature data set onto the first two kernel principal components based on $k_1$ and $k_2$, respectively. In both figures, images  located close to each other are showing similar landscapes. For example, the regions that we zoomed in for more precise inspection only contain images showing mountains. Other regions only comprise images showing forests or coastlines. The two figures differ in their spatial extent: in the projection based on $k_2$, except for three dense regions of mountains and forests, respectively, the images are roughly uniformly spread. 
For comparison 
of our kernel PCA embeddings with an ordinal embedding, 
Figure \ref{gnmds_emb_nature} shows 
a GNMDS embedding \citep{AgarwalEtal07} of the nature data set.
In this ordinal embedding there are three outliers that  are located far apart from the bulk of the images. In the bulk of the images, which we zoomed in for more precise inspection, images  located close to each other are showing similar landscapes 
similarly to 
our kernel PCA embeddings.

\vspace{1cm}

\begin{figure*}[t]
\centering
\includegraphics[width=0.95\textwidth]{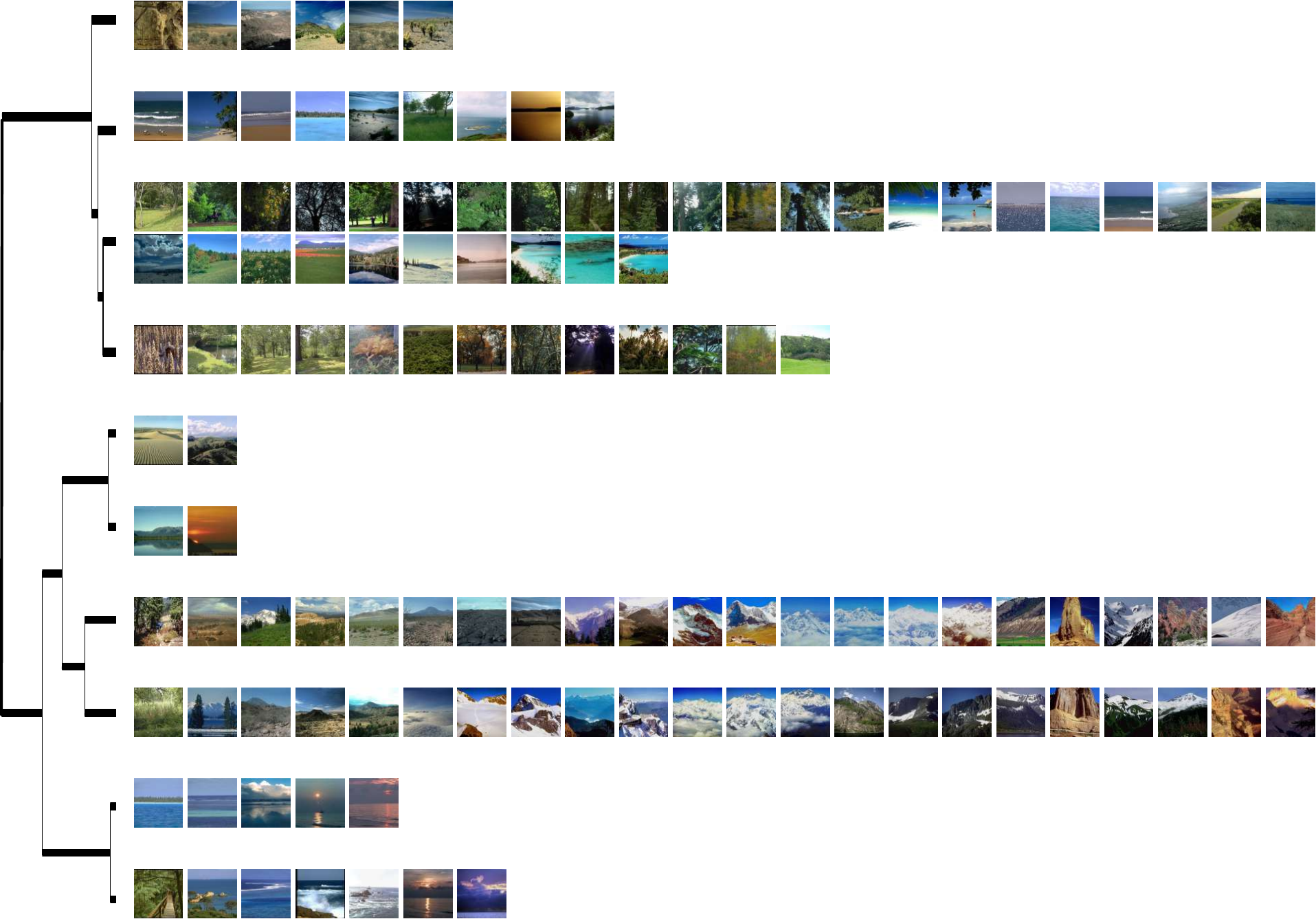}
\caption{Part of the dendrogram that we obtained by applying the  kernelized version of complete-linkage clustering based on $k_1$ to the nature data set.}\label{Linkage_k1_nature}
\end{figure*}

\begin{figure*}[h!]
\vspace{1cm}
\centering
\includegraphics[width=0.95\textwidth]{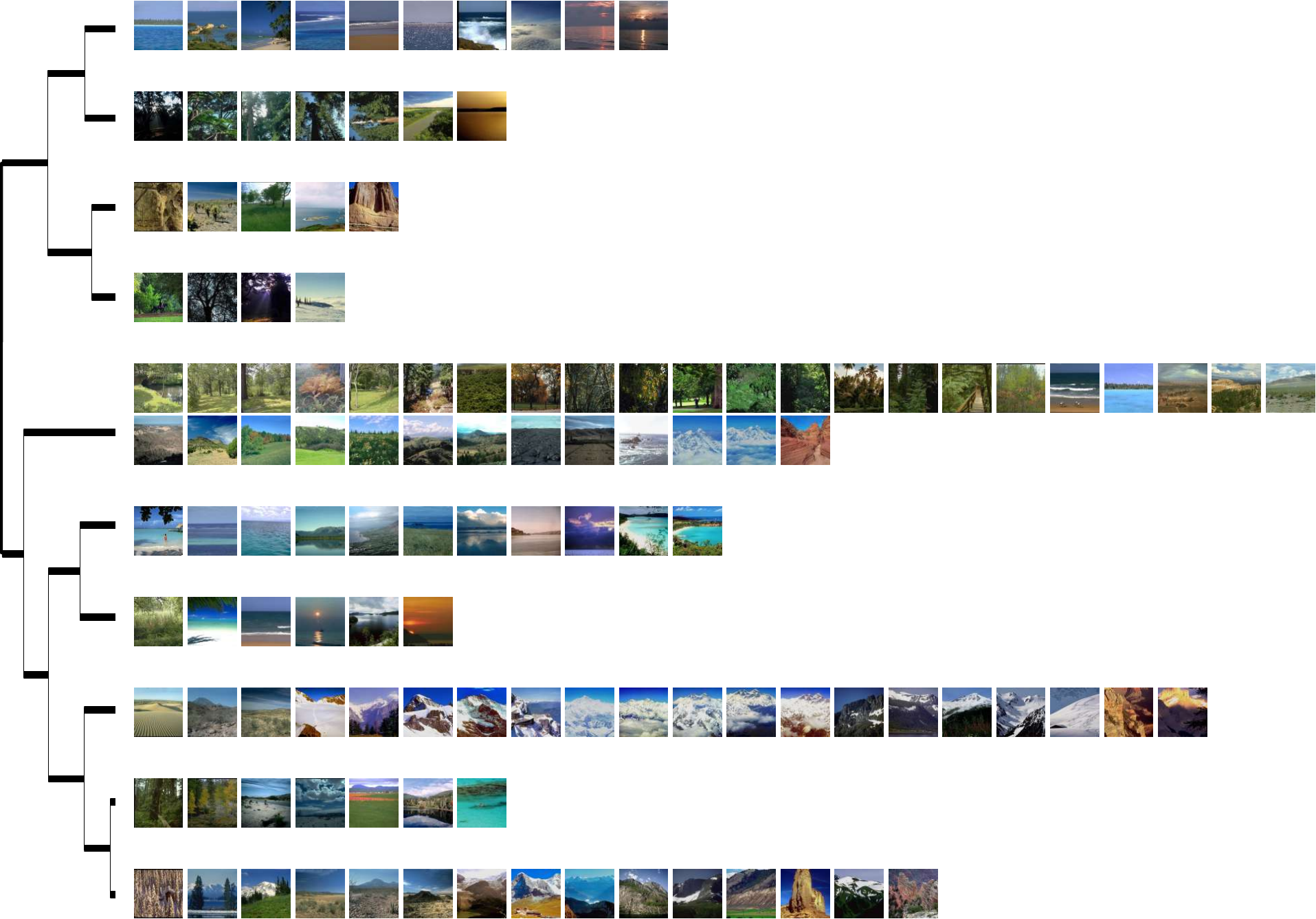}
\caption{Part of the dendrogram that we obtained by applying the  kernelized version of complete-linkage clustering based on $k_2$ to the nature data set.}\label{Linkage_k2_nature}
\end{figure*}

Figure \ref{Linkage_k1_nature} and Figure \ref{Linkage_k2_nature} show parts of the dendrograms that we obtained by applying the  kernelized version of complete-linkage clustering based on $k_1$ and $k_2$, respectively, to the nature data set. In both figures, most of the ten 
clusters  contain  homogeneous  images. For example, the first cluster in Figure \ref{Linkage_k1_nature} only contains images of desertlike landscapes whereas the fourth cluster only shows forests and the seventh and eighth cluster mainly consist of images of mountains. The sixth cluster in Figure~\ref{Linkage_k2_nature} only shows coastlines and so does the first cluster except for one image that shows clouds viewed from above (a \emph{sea} of clouds).

\vspace{18mm}

\vspace{20mm}

\end{document}